\documentclass{article}

\PassOptionsToPackage{sort}{natbib}
\usepackage[final]{neurips_2024}

\usepackage{amsmath}
\usepackage{amsfonts}
\usepackage{bm}
\usepackage{amssymb}
\usepackage{mathtools}
\usepackage{amsthm}
\usepackage{dsfont}
\usepackage{bbm} 
\usepackage{stmaryrd} 
\usepackage{enumitem}
\usepackage{amsthm}

\usepackage{mleftright} 
\newcommand{\norm}[1]{\left\lVert#1\right\rVert} 
\newcommand{\tr}[1]{\operatorname{Tr}\mleft(#1\mright)}
\newcommand{\softmax}[1]{\operatorname{\mathrm{softmax}}\mleft(#1\mright)}

\newcommand{\vertiii}[1]{{\left\vert\kern-0.25ex\left\vert\kern-0.25ex\left\vert #1 
    \right\vert\kern-0.25ex\right\vert\kern-0.25ex\right\vert}}

\newcommand{\score}[1]{\mcal{S}\mleft(#1\mright)}
\newcommand{\acc}[1]{\mrm{Acc}\mleft(#1\mright)}
\newcommand{\probmap}[1]{\sigma\mleft(#1\mright)}

\newcommand{\RR}{\mathbb{R}}
\newcommand{\PP}{\mathbb{P}}

\newcommand{\mbf}[1]{\mathbf{#1}}
\newcommand{\mcal}[1]{\mathcal{#1}}
\newcommand{\mrm}[1]{\mathrm{#1}}
\newcommand{\mbbm}[1]{\mathbbm{#1}}

\usepackage[table]{xcolor}  
\definecolor{rightblue}{RGB}{76,114,176} 
\definecolor{rightorange}{RGB}{221,132,82} 
\definecolor{aliceblue}{rgb}{0.94, 0.97, 1.0} 
\definecolor{darkcerulean}{rgb}{0.03, 0.27, 0.49} 
\definecolor{iris}{rgb}{0.35, 0.31, 0.81} 
\definecolor{carmine}{rgb}{0.59, 0.0, 0.09} 
\definecolor{green(munsell)}{rgb}{0.0, 0.66, 0.47} 
\definecolor{celadon}{rgb}{0.67, 0.88, 0.69} 
\definecolor{bluerow}{rgb}{0.0, 0.53, 0.74} 
\definecolor{blue-violet}{rgb}{0.54, 0.17, 0.89}
\definecolor{lavenderblue}{rgb}{0.8, 0.8, 1.0}
\def\propcolor{lavenderblue!25}
\usepackage{hyperref}
   \hypersetup{
     final=true,
     plainpages=false,
     pdfstartview=FitV,
     pdftoolbar=true,
     pdfmenubar=true,
     bookmarksopen=true,
     bookmarksnumbered=true,
     breaklinks=true,
     linktocpage=true,
     colorlinks=true,
     linkcolor=blue-violet, 
     urlcolor=iris, 
     citecolor=darkcerulean, 
     anchorcolor=black
   }

\usepackage{url}            
\usepackage{nicefrac}       
\usepackage{microtype}      
\usepackage{subfigure}

\usepackage{algorithm2e}
\RestyleAlgo{ruled}

\usepackage{booktabs}
\usepackage{multirow}
\usepackage{graphicx}
\usepackage{caption} 
\usepackage{mdframed}
\newmdtheoremenv[topline=false, bottomline=false, leftline=false, rightline=false, backgroundcolor=aliceblue,%
innertopmargin=\topskip, splittopskip=\topskip, skipbelow=\baselineskip, skipabove=\baselineskip]{boxthm}{Theorem}[section]
\newmdtheoremenv[topline=false, bottomline=false, leftline=false, rightline=false, backgroundcolor=aliceblue,%
innertopmargin=\topskip, splittopskip=\topskip, skipbelow=\baselineskip, skipabove=\baselineskip]{boxprop}[boxthm]{Proposition}
\newmdtheoremenv[topline=false, bottomline=false, leftline=false, rightline=false, backgroundcolor=aliceblue,%
innertopmargin=\topskip, splittopskip=\topskip, skipbelow=\baselineskip, skipabove=\baselineskip]{boxexample}[boxthm]{Example}
\newmdtheoremenv[topline=false, bottomline=false, leftline=false, rightline=false, backgroundcolor=aliceblue,%
innertopmargin=\topskip, splittopskip=\topskip, skipbelow=\baselineskip, skipabove=\baselineskip]{boxcor}[boxthm]{Corollary}
\newmdtheoremenv[topline=false, bottomline=false, leftline=false, rightline=false, backgroundcolor=aliceblue,%
innertopmargin=\topskip, splittopskip=\topskip, skipbelow=\baselineskip, skipabove=\baselineskip]{boxlem}[boxthm]{Lemma}
\newmdtheoremenv[topline=false, bottomline=false, leftline=false, rightline=false, backgroundcolor=aliceblue,%
innertopmargin=\topskip, splittopskip=\topskip, skipbelow=\baselineskip, skipabove=\baselineskip]{boxdef}[boxthm]{Definition}

\usepackage{tikz}
\usepackage{varwidth}

\usepackage{stmaryrd}

\usepackage{dirtytalk}
\MakeRobust{\say} 

\usepackage{makecell}

\usepackage{wrapfig}

\newcommand{\method}{\textsc{MaNo}}
\newcommand{\normalization}{\texttt{softrun}}
\newcommand{\threshold}{\Phi\mleft(\mcal{D}_\mrm{test}\mright)}

\usepackage{mathtools} 
\DeclarePairedDelimiterX{\infdivx}[2]{(}{)}{%
  #1\;\delimsize\|\;#2%
}
\newcommand{\kld}{\mrm{KL}\infdivx}

\newcommand{\II}[1]{\mathbb{I}(#1)}

\usepackage[many]{tcolorbox}

\def\size{0.3}
\usepackage[para]{footmisc}

\makeatletter
\long\def\@makefntext#1{\leavevmode
  \@makefnmark\nobreak
  #1%
}
\makeatother

\title{\method{}: Exploiting Matrix Norm for Unsupervised Accuracy Estimation Under Distribution Shifts}

\author{%
  \makebox[\size\textwidth]{Renchunzi Xie\textsuperscript{*}} \\
  \makebox[\size\textwidth]{Nanyang Technological University} \\
  \makebox[\size\textwidth]{Singapore}\\
  \makebox[\size\textwidth]{\href{mailto:renchunzi.xie@ntu.edu.sg}{\small \texttt{renchunzi.xie@ntu.edu.sg}}}
  \And
  \makebox[\size\textwidth]{Ambroise Odonnat\textsuperscript{*}} \\
  \makebox[\size\textwidth]{Huawei Noah's Ark Lab, Inria\textsuperscript{$\diamond$}} \\
  \makebox[\size\textwidth]{Paris, France}\\
  \makebox[\size\textwidth]{\href{mailto:ambroise.odonnat@gmail.com}{\small \texttt{ambroise.odonnat@gmail.com}}} \\
  \AND
  \makebox[\size\textwidth]{Vasilii Feofanov\textsuperscript{*}} \\
  \makebox[\size\textwidth]{Huawei Noah's Ark Lab} \\
  \makebox[\size\textwidth]{Paris, France}\\
  \makebox[\size\textwidth]{\href{mailto:vasilii.feofanov@gmail.com}{\small \texttt{vasilii.feofanov@gmail.com}}}
  \And 
  \makebox[\size\textwidth]{Weijian Deng} \\
  \makebox[\size\textwidth]{Australian National University} \\
  \makebox[\size\textwidth]{Canberra, Australia}\\
  \makebox[\size\textwidth]{\href{mailto:weijian.deng@anu.edu.au}{\small \texttt{weijian.deng@anu.edu.au}}}
  \And
  \makebox[\size\textwidth]{Jianfeng Zhang} \\
  \makebox[\size\textwidth]{Huawei Noah’s Ark Lab} \\
  \makebox[\size\textwidth]{Shenzhen, China}\\
  \makebox[\size\textwidth]{\href{mailto:zhangjianfeng3@huawei.com}{\small \texttt{zhangjianfeng3@huawei.com}}}
  \AND
  \makebox[\size\textwidth]{Bo An} \\
  \makebox[\size\textwidth]{Skywork AI} \\
  \makebox[\size\textwidth]{Nanyang Technological University} \\
  \makebox[\size\textwidth]{Singapore}\\
  \makebox[\size\textwidth]{\href{mailto:boan@ntu.edu.sg}{\small \texttt{boan@ntu.edu.sg}}}
}

\begin{document}

\def\thefootnote{*}\footnotetext{Equal contribution. Correspondence to: Bo An - \href{mailto:boan@ntu.edu.sg}{\texttt{boan@ntu.edu.sg}}.} 
\def\thefootnote{$\diamond$}\footnotetext{Univ. Rennes 2, CNRS, IRISA}

\renewcommand{\thefootnote}{\arabic{footnote}}
\setcounter{footnote}{0}

\addtocontents{toc}{\protect\setcounter{tocdepth}{0}}

\maketitle


\begin{abstract}
Leveraging the model’s outputs, specifically the logits, is a common approach to estimating the test accuracy of a pre-trained neural network on out-of-distribution (OOD) samples without requiring access to the corresponding ground-truth labels.
Despite their ease of implementation and computational efficiency, current logit-based methods are vulnerable to overconfidence issues, leading to prediction bias, especially under the natural shift. In this work, we first study the relationship between logits and generalization performance from the view of low-density separation assumption. Our findings motivate our proposed method \method{} that \textbf{(1)}~applies a data-dependent normalization on the logits to reduce prediction bias, and \textbf{(2)} takes the $L_p$ norm of the matrix of normalized logits as the estimation score. Our theoretical analysis highlights the connection between the provided score and the model's uncertainty. 
We conduct an extensive empirical study on common unsupervised accuracy estimation benchmarks and demonstrate that \method{} achieves state-of-the-art performance across various architectures in the presence of synthetic, natural, or subpopulation shifts. The code is available at \url{https://github.com/Renchunzi-Xie/MaNo}.
\end{abstract}

\section{Introduction}
\label{sec:intro}
The deployment of machine learning models in real-world scenarios is frequently challenged by distribution shifts between the training and test data. These shifts can substantially deteriorate the model's performance during testing~\citep{quinonero2008dataset,geirhos2018imagenet,koh2021wilds} and introduce significant risks related to AI safety~\citep{hendrycks2022x,deng2021labels}. To alleviate this issue, it is common to monitor model performance by periodically collecting the ground truth labels for a subset of the current test dataset~\citep{lu2023characterizing}. However, this approach is often resource-intensive and time-consuming, which motivates the importance of estimating the model's performance on out-of-distribution (OOD) data in an unsupervised manner, also known as \textit{Unsupervised Accuracy Estimation} \citep{donmez2010unsupervised}. 

Due to privacy constraints and computational efficiency, one of the most popular ways to estimate accuracy without labels is to rely on the model's outputs, called logits, as a source of confidence in the model's predictions~\citep{hendrycks2016baseline, garg2022leveraging, deng2023confidence, guillory2021predicting}. For instance, \textit{ConfScore}~\citep{hendrycks2016baseline} leverages the average maximum \texttt{softmax} probability as the test accuracy estimator, while \citet{deng2023confidence} has recently proposed to estimate the accuracy via the nuclear norm of the \texttt{softmax} probability matrix. These approaches, however, tend to underperform on the natural shift applications while the intuition behind the use of logits remains unclear. This motivates us to ask:
\begin{center}
\tcbox[on line, boxsep=0mm, left=1.5mm,right=1.5mm, boxrule=0.7pt, colback=\propcolor]{
\textbf{Question 1:} \textit{What explains the correlation between logits and generalization performance?}
}
\end{center}
In Section~\ref{sec:proposed_method}, we show that logits are connected to the model's margins, \textit{\textit{i.e.}}, the distances between the learned embeddings, and the decision boundaries. Inspired by the low-density separation (LDS) assumption~\citep{chapelle05lds,feofanov23qlds} that optimal decision boundaries should lie in low-density regions, we propose \method{}, an estimation score that aggregates the margins at a dataset level by taking the $L_p$-norm of the normalized model's prediction matrix to evaluate the density around decision boundaries. Nevertheless, logit-based approaches are known to suffer from overconfidence~\citep{wei2022mitigating, odonnat2024leveraging}, resulting in high prediction bias, especially under poorly-calibrated scenarios. This leads us to another critical question:
\begin{center}
\tcbox[on line, boxsep=0mm, left=1.5mm,right=1.5mm, boxrule=0.7pt, colback=\propcolor]{
\textbf{Question 2:} \textit{How to alleviate the overconfidence issues of logits-based methods?}  
}
\end{center}
In Section~\ref{sec:normalization}, we reveal that this question is connected to the normalization of logits and show that the widely-used \texttt{softmax} normalization accumulates errors in the presence of prediction bias, which can lead to overconfidence and significantly degrade the performance of existing accuracy estimation methods in poorly-calibrated scenarios. To mitigate this issue, we propose a novel normalization strategy called \normalization{} that takes into account the empirical distribution of logits and aims to find a trade-off between information completeness of ground-truth logits and error accumulation.
 
\paragraph{Summary of our contributions.} \textbf{(1)} We show that logits are informative of generalization performance through the lens of the low-density separation assumption by reflecting the distances to decision boundaries. \textbf{(2)} We identify the failure of the commonly-used \texttt{softmax} normalization that accumulates errors under poorly calibrated because of its overconfidence, leading to biased estimation. \textbf{(3)} We propose \method{}, a training-free estimation method that quantifies the global distances to decision boundaries by taking the $L_p$ norm of the logits matrix. \method{} relies on \normalization{}, a novel normalization technique that makes a trade-off between information completeness and error accumulation and is robust to different calibration scenarios. In addition, we demonstrate its connection to the model's uncertainty. \textbf{(4)} We demonstrate the superiority of \method{} compared to $11$ competitors with a large-scale empirical evaluation including $12$ benchmarks across diverse distribution shifts. Results show that \method{} consistently improves over the state-of-the-art baselines, including on the challenging natural shift.

\section{Problem Statement}
\label{sec:background}
\paragraph{Setting.} Consider a classification task with input space $\mcal{X}\subset\RR^D$ and label space $\mcal{Y} = \{1, \ldots, K\}$. Let $p_S$ and $p_T$ be the source and target distributions on $\mcal{X} \times \mcal{Y}$, respectively, with $p_S \neq p_T$. We parameterize a neural network $f \colon \mcal{X} \to \RR^K$ as $f = f_{\mbf{W}} \circ f_{\bm{\varphi}}$, where $f_{\bm{\varphi}} \colon \mcal{X} \to \RR^q$ is a feature extractor and $f_{\mbf{W}} \colon \RR^q \to \RR^K$ is a linear classifier with parameters $\mbf{W} = \mleft(\bm{\omega}_k\mright)_{k=1}^K \in \RR^{q \times K}$. Further, we denote an input by $\mbf{x}$, its corresponding label by $y$, its representation by $\mbf{z}=f_{\bm{\varphi}}(\mbf{x})$ and logits by $\mbf{q} = f(\mbf{x}) = (\bm{\omega}_k^\top\mbf{z})_k \in \RR^K$. The accuracy of $f$ on $\mcal{D}$ is defined as $\acc{f, \mcal{D}} \coloneqq \frac{1}{\lvert \mcal{D} \rvert} \sum_{(\mbf{x}, y) \in \mcal{D}} \mbbm{1}_{\hat{y} = y}$ with predicted labels $\hat{y}$. The probability simplex is denoted by $\Delta_K = \{\mbf{p} \in [0,1]^K | \mathbbm{1}_K^\top\mbf{p}=1\}$.

\paragraph{Unsupervised accuracy estimation.}  
Given a model $f$ pre-trained on a training set $\mcal{D}_\mrm{train}$ with samples drawn i.i.d. from $p_S$, the goal of unsupervised accuracy estimation is to \underline{assess} its generalization \underline{performance} on a given \underline{unlabeled test set} $\mcal{D}_\mrm{test} = \{\mbf{x}_i\}_{i=1}^N$ with $N$ samples drawn i.i.d. from $p_T$. More specifically, we assume (i) a source-free regime (no direct access to $\mcal{D}_\mrm{train}$), (ii) no access to test labels, and (iii) a distribution shift, \textit{\textit{i.e.}} $p_S \neq p_T$, . In this challenging setup, which often occurs in real-world scenarios when ground-truth labels are inaccessible at a test time, we aim to design an estimation score $\score{f, \mcal{D}_\mrm{test}}$ that exhibits a linear correlation with the true OOD accuracy $\mrm{Acc}\mleft( f, \mcal{D}_\mrm{test}\mright)$. Following the standard closed-set setting, both $p_T$ and $p_S$ involve the same $K$ classes. For an extended discussion of related work on unsupervised accuracy estimation, we refer the reader to Appendix~\ref{app:related_work}.

\section{What Explains the Correlation between Logits and Test Accuracy?}
\label{sec:proposed_method}
Although existing literature has shown the feasibility of unsupervised accuracy prediction under distribution shift by utilizing the model's logits \citep{garg2022leveraging, deng2023confidence, guillory2021predicting}, the reason behind this empirical success remains unclear. In this section, we seek to understand when and why logits can be informative for analyzing generalization performance. Based on the derived understanding, we propose our approach, \method{}, for estimating generalization performance.

\subsection{Motivation}
\label{sec:explain}
\paragraph{Logits reflect the distances to decision boundaries.} 
We analyze logits from a linear classification perspective in the embedding space, where the decision boundary of class $k$ is the hyperplane $\{\bm{z}'\in\RR^q | \bm{\omega}_k^\top \bm{z}'=0\}$. In Appendix~\ref{app:lem_distance_boundary}, we remind that the distance from a point $\mbf{z}$ to hyperplane $\bm{\omega}_k$ is given by $\mrm{d}(\bm{\omega}_k, \mbf{z}) = \lvert \bm{\omega}_k^\top \bm{z}\rvert/\norm{\bm{\omega}_k}$. As the pre-trained model is fixed and $\bm{\omega}_k$ can be normalized, we derive that the logits in absolute values are proportional to the distance from the learned embeddings to the decision boundaries, \textit{i.e.}, $|\mbf{q}_k|=\lvert\bm{\omega}_k^\top\mbf{z}\rvert \propto d(\bm{\omega}_k, \mbf{z}),\forall k$. This indicates that the magnitude of logits reflects how close the corresponding embedding is from each decision boundary.  

\begin{wrapfigure}{r}{0.6\textwidth}
    \centering
    \subfigure[High-density region]{
\includegraphics[width=0.28\textwidth]{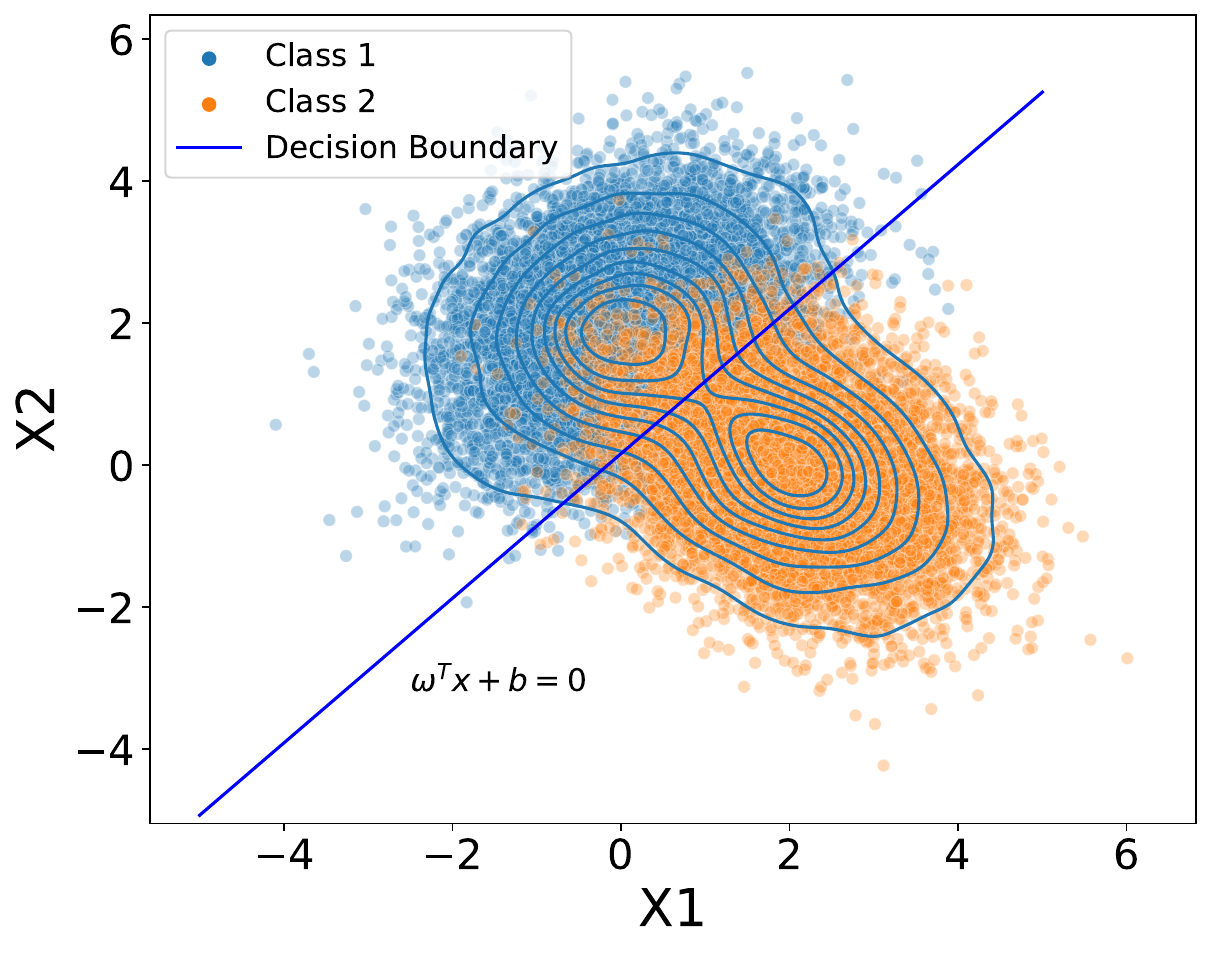}
        \label{fig:high_density}}
    \subfigure[Low-density region.]{
\includegraphics[width=0.28\textwidth]{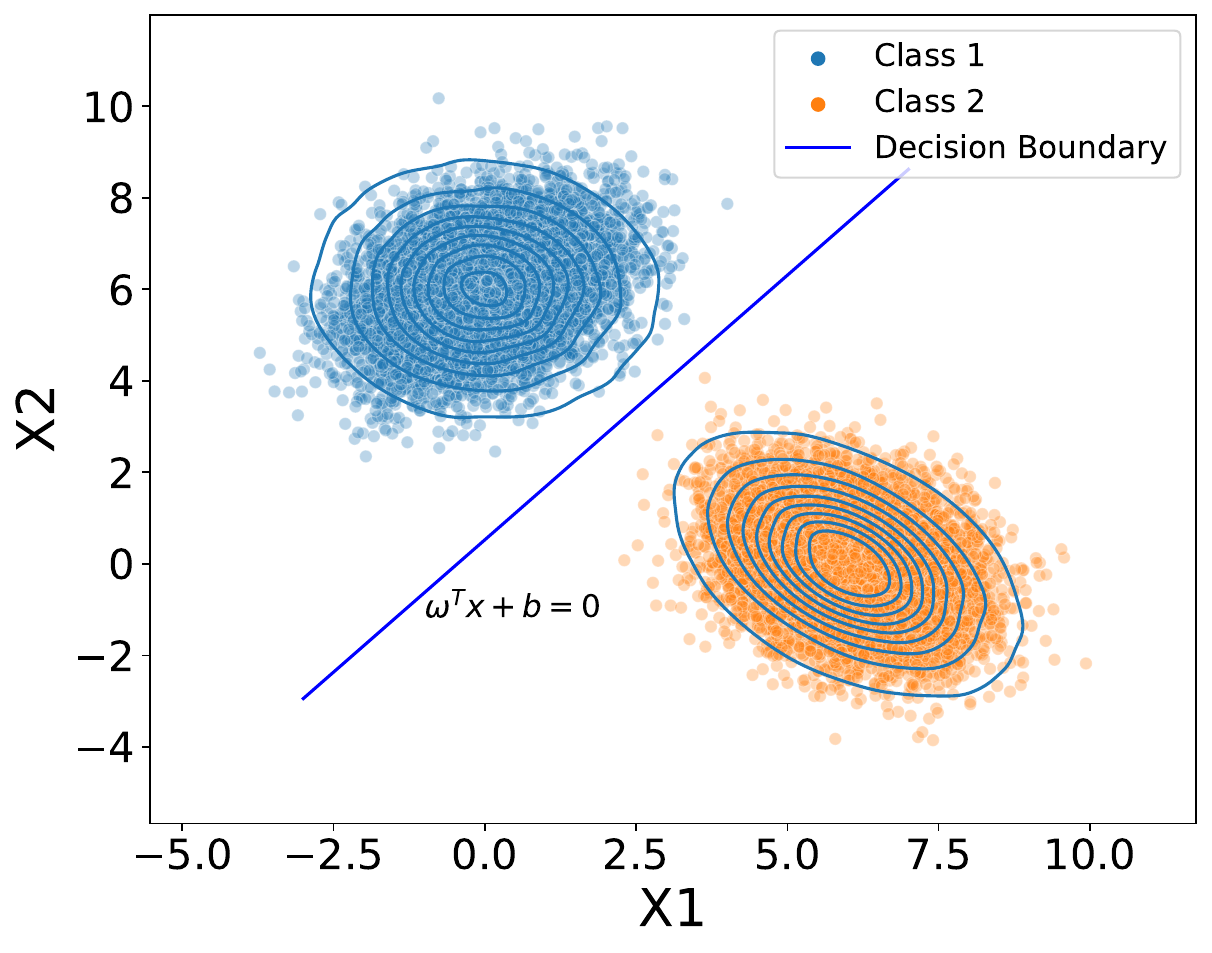}
        \label{fig:low_density}}
    \caption{\textbf{Illustration of the LDS assumption.} When the boundary passes through dense regions \textbf{(a)}, margins have little predictive power and cannot be used without labels. On the contrary, margins are informative in sparse regions \textbf{(b)}.}
    \label{fig:lds_assumption}
\end{wrapfigure}

\paragraph{Low-density separation assumption.}
When dealing with unlabeled data, it is required to make assumptions on the relationship between the distance to decision boundaries and generalization performance. The low-density separation assumption (LDS, \citeauthor{chapelle05lds}, \citeyear{chapelle05lds}) states that optimal decision boundaries should lie in low-density regions (Figure~\ref{fig:lds_assumption}) so that unlabeled margin $\lvert\bm{\omega}_k^\top\mbf{z}\rvert$ reflects reliable confidence in predicting $\mbf{x}$ to the class $k$. The assumption is often empirically supported as the misclassified samples tend to be significantly closer to the decision boundary than the correctly classified ones \citep{mickisch2020understanding}. 
This might indicate that \textbf{the absolute values of the logits are positively correlated to its generalization performance}.  

\paragraph{Assumptions on the prediction bias.} It is important to note that the LDS assumption has been initially proposed for semi-supervised learning where labeled and unlabeled data are assumed to come from the same distribution, which is not the case in our setting.  
This leads to logits writing $f(\mbf{x}) = \mbf{q}^* + \bm{\varepsilon}$ in the general case\footnote{We write this decomposition without loss of generality as no restrictions are imposed on $\bm{\varepsilon}$.}, \textit{i.e.}, subject to a potentially non-negligible prediction bias $\bm{\varepsilon} = (\varepsilon_k)_k$ with respect to the ground-truth logits $\mbf{q}^* \in \RR^K$. The following proposition shows the impact of the prediction bias on the divergence between the true class posterior probabilities, assumed modeled as $\mbf{p} = \softmax{\mbf{q}^*}\in \Delta_K$, and the estimated ones $\mbf{s} = \softmax{f(\mbf{x})}\in \Delta_K$.

\begin{boxprop}
\label{prop:kl_bounds}
Let $\bm{\varepsilon}_{+}\!=\!(\max_{l}\{\varepsilon_l\} - \varepsilon_k)_k$. Then, the KL divergence between $\mbf{p}$ and $\mbf{s}$ verifies
\begin{equation*}
    0 \leq \mrm{KL}\mleft(\mbf{p} || \mbf{s}\mright) \leq \bm{\varepsilon}_{+}^T\mbf{p}.
\end{equation*}
\end{boxprop}
The proposition indicates that a large approximation error of the posterior may be caused by prediction bias that has a large norm and\,/\,or bad alignment with the true probabilities. Thus, the logit-based methods assume that the magnitude of the bias is reasonably bounded while the direction of bias does not drastically harm the ranking of classes by probabilities. We elaborate on this discussion and present the proof of Proposition \ref{prop:kl_bounds} in Appendix \ref{app:impact-pred-err}.

\subsection{\method{}: Predicting Generalization Performance With Matrix Norm of Logits}
\label{sec:method}
We have shown a connection between the feature-to-boundary distances and generalization performance as well as the impact of the prediction bias. Based on the derived intuition, we introduce \method{} that leverages the model margins at the dataset level performing two steps: normalization and aggregation. The pseudo-code of \method{} is provided in Appendix~\ref{app:pseudo_code}.

\paragraph{Step 1: Normalization.} 
Given that logits can exhibit significant variations in their scale depending on the input $\mbf{x}$, it is crucial to normalize the logits within a standardized range to prevent outliers from exerting disproportionate influence on the estimation. A natural range stems from the fact that most deep classifiers have outputs in $\Delta_K$, which amounts to applying a normalization function $\sigma \colon \RR^K \to \Delta_K$ on top of the pre-trained neural network~\citep{mensch19geomsoftmax}, where $\Delta_K$ refers to probability simplex. This ensures having logits entries in~$[0, 1]$. For each test sample $\mbf{x}_i$, we first extract its learned feature representation $\mbf{z}_i = f_{\bm{\varphi}}(\mbf{x}_i)$. Then, logits corresponding to this representation are computed as $\mbf{q}_i = f_{\bm{W}}(\mbf{z}_i) \in \RR^K$. The normalization procedure results in a prediction matrix 
$\mbf{Q} \in \RR^{N \times K}$ with each row $ \mbf{Q}_i$ containing the normalized logits of an input sample:
\begin{equation}
\label{eq:row_matrix}
    \mbf{Q}_i = \sigma(\mbf{q}_i) \in \Delta_K,
\end{equation}
where $\sigma$ denotes the normalization function for the logits values.
It is worth noting that not all normalization methods are appropriate candidates. The selection of a suitable normalization function $\sigma$ based on different calibration scenarios will be discussed in detail in Section~\ref{sec:normalization}.

\paragraph{Step 2: Aggregation.} Once the logits are scaled, we aggregate the dataset-level information on feature-to-boundary distances by taking the entry-wise $L_p$ norm of the prediction matrix $\mbf{Q}$, which can be expressed as:
\begin{equation}
\label{eq:estimation_score}
     \score{f, \mathcal{D}_{\text{test}}} = \frac{1}{\sqrt[\leftroot{-2}\uproot{2}p]{NK}}\lVert\mbf{Q}\rVert_p = \mleft(\frac{1}{NK}\sum_{i=1}^N\sum_{k=1}^K \lvert\sigma(\bm{q}_i)_k\rvert^p\mright)^{\frac{1}{p}},
\end{equation}
As we have $||\mathbf{Q}||_p \leq \sqrt[\leftroot{-2}\uproot{2}_p]{NK} \max(\mathbf{Q}_{ij})=\sqrt[\leftroot{-2}\uproot{2}p]{NK}$ ($\mathbf{Q}_{ij} \in [0, 1]$), the scaling by $\sqrt[\leftroot{-2}\uproot{2}p]{NK}$ leads to $\score{f, \mathcal{D}_{\text{test}}} \in [0, 1]$, providing a standardized metric regardless of variations in the size of the test dataset $N$ and the number of classes $K$. As $p$ increases, \method{} puts greater emphasis on high-margin terms, focusing on confident classification hyperplanes. In the extreme case where $p \rightarrow \infty$, we have $|\mbf{Q}||_p \rightarrow \max(\mbf{Q}_{ij})$. In practice, we choose $p=4$ in all experiments and provide an ablation study on $p$ in Appendix~\ref{app:choice_lp}. As the $L_p$ norm is straightforward to compute, our approach is scalable and efficient compared to the current state-of-the-art method Nuclear~\citep{deng2023confidence} that requires performing a singular value decomposition. 

\subsection{Theoretical Analysis of \method{}}
\label{sec:theoretical_insights}
In this section, we provide the theoretical support for the positive correlation between \method{} and test accuracy. More specifically, we reveal that our proposed score is connected with the uncertainty of the neural network's predictions in Theorem~\ref{thm:connection_uncertainty}. Before presenting this result, we recall below the definition of Tsallis $\alpha$-entropies introduced in~\citet{tsallis1988entropy}.
\begin{boxdef}[Tsallis $\alpha$-entropies~\citep{tsallis1988entropy}]
\label{def:tsallis_entropy}
    Let $\alpha > 1$ and $k > 0$. The Tsallis $\alpha$-entropy is defined as:
    \begin{equation*}
        \mathbf{H}_{\alpha}^\mathrm{T}(\mathbf{p}) = k(\alpha-1)^{-1}(1 - \lVert\mbf{p}\rVert_{\alpha}^{\alpha}).
    \end{equation*}
\end{boxdef}
In this work, we choose $k=\frac{1}{\alpha}$ following~\citet{blondel19Fyclassifiers}. The Tsallis entropies generalize the Shannon entropy (limit case $\alpha \to 1$) and have been used in various applications ~\citep{blondel19Fyclassifiers, blondel21FYlearning, muzellec17tsallis}. More details can be found in Appendix~\ref{app:tsallis_entropy}. The following theorem, whose proof is deferred to Appendix~\ref{app:thm_connection_uncertainty}, states that the estimation score obtained with \method{} is a function of the average Tsallis entropy of the normalized neural network's logits.
\begin{boxthm}[Connection to uncertainty]
\label{thm:connection_uncertainty}
Let $p > 1$, $a=\frac{p(p-1)}{K}$ and $b = \frac{1}{K}$. Given a test set $\mcal{D}_\mrm{test} = \{\mbf{x}_i\}_{i=1}^N$, corresponding logits $\mbf{q}_i = f(\mbf{x}_i)$, a normalization function $\sigma \colon \RR^K \to \Delta_K$ and $p > 1$, the estimation score $\score{f, \mcal{D}_\mrm{test}}$ provided by \method{} (Algorithm~\ref{alg:our_method}) verifies 
\begin{equation}
\label{eq:connection_uncertainty}
    \score{f, \mcal{D}_\mrm{test}}^p = - a\mleft(\frac{1}{N}\sum_{i=1}^N \mathbf{H}_p^\mathrm{T}\mleft(\probmap{\mbf{q}_i}\mright)\mright) + b.
\end{equation}
\end{boxthm}
As $a > 0$, Theorem~\ref{thm:connection_uncertainty} implies that the estimation score provided by \method{} is negatively correlated with the average Tsallis-entropy on the test set. In particular, the less certain the model is on test data, the lower the test accuracy is and the higher the entropy term is in Eq.~\eqref{eq:connection_uncertainty}, resulting in a lower score $\score{f, \mcal{D}_\mrm{test}}$. As the converse sense holds, \textsc{MaNo} provides a score positively correlated to the test accuracy. This follows the findings of~\citet{guillory2021predicting, wang2021tent} and empirically confirmed in Section~\ref{sec:experiments} for various architectures, datasets, and types of shift. 

\section{How to Alleviate Overconfidence Issues of Logit-Based Methods?} \label{sec:normalization}
The most common normalization technique of existing logit-based approaches is the \texttt{softmax} normalization. In this section, we show that the widely used \texttt{softmax} is sensitive to prediction bias, which hinders the quality of the estimation in poorly calibrated scenarios. To alleviate this issue, we propose a novel normalization strategy, \normalization{}, which balances the information completeness and overconfidence accumulation based on calibration.
\begin{figure}[!t] 
    \centering
    \subfigure[Calibration curves.]{
\includegraphics[width=0.31\textwidth]{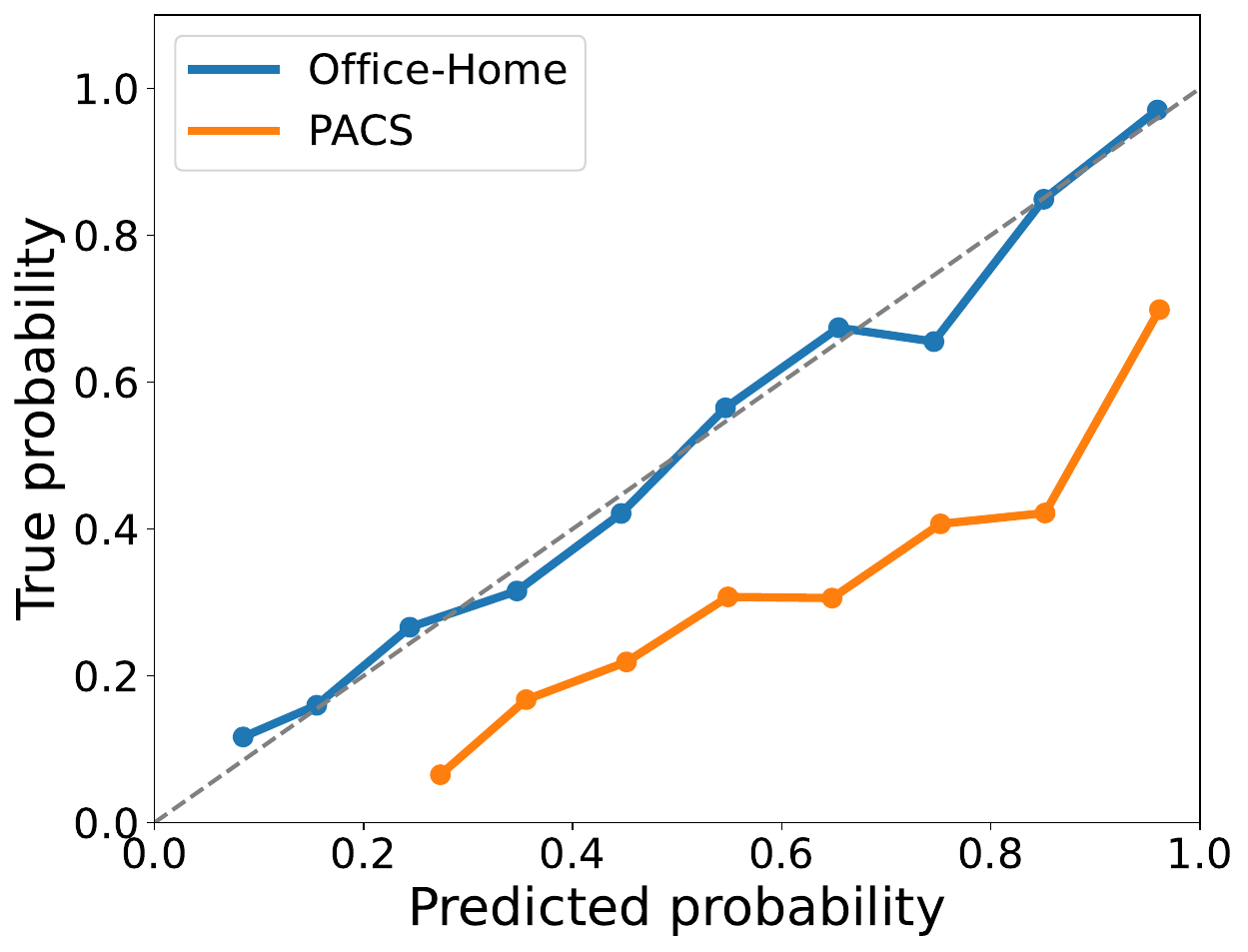}
        \label{fig:reliability}}
    \subfigure[Type of normalization.]{
\includegraphics[width=0.3\textwidth]{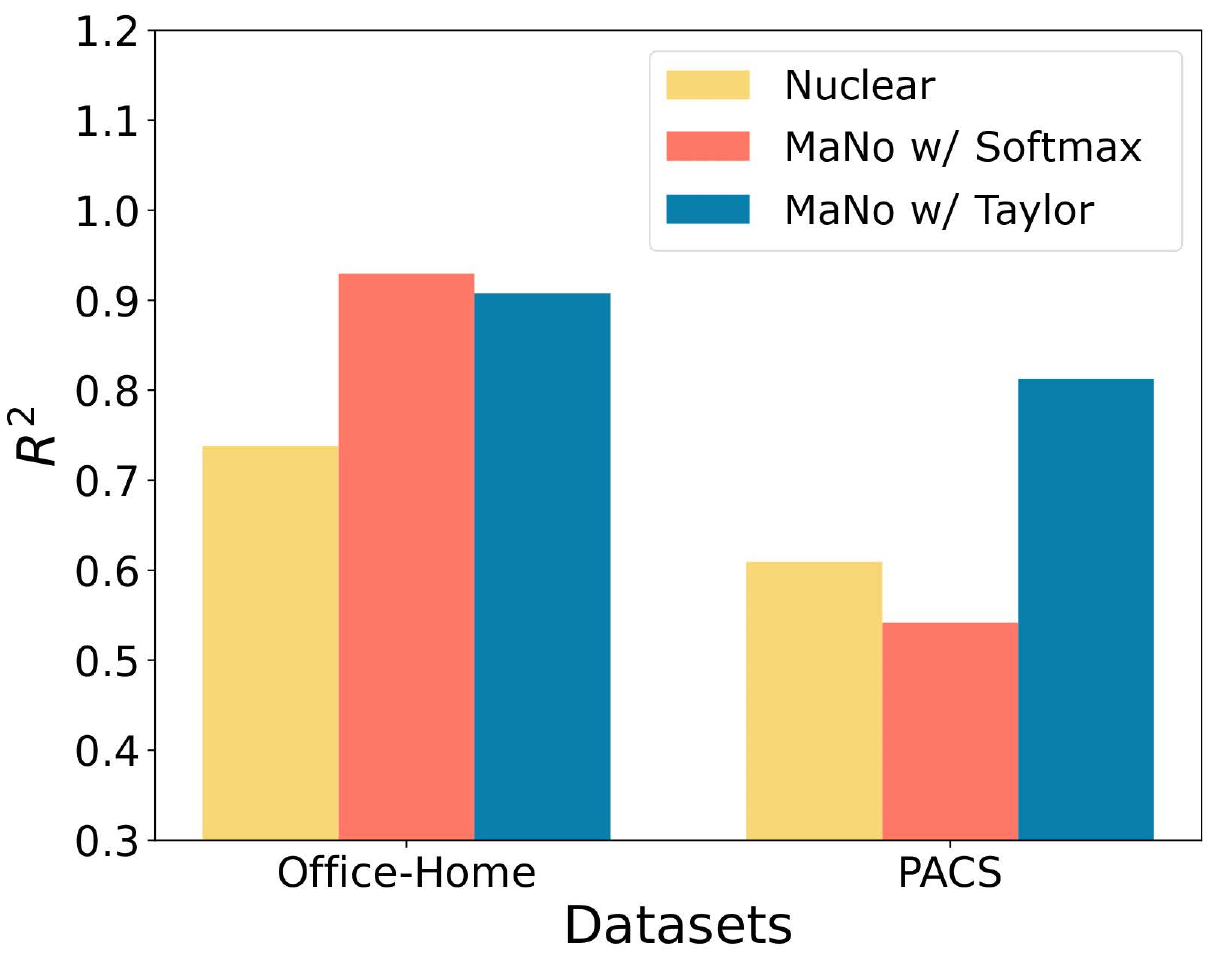}
        \label{fig:norm_selection}}
    \subfigure[Order $n$ in Eq.~\eqref{eq:taylor}.]{
    \includegraphics[width=0.31\textwidth]{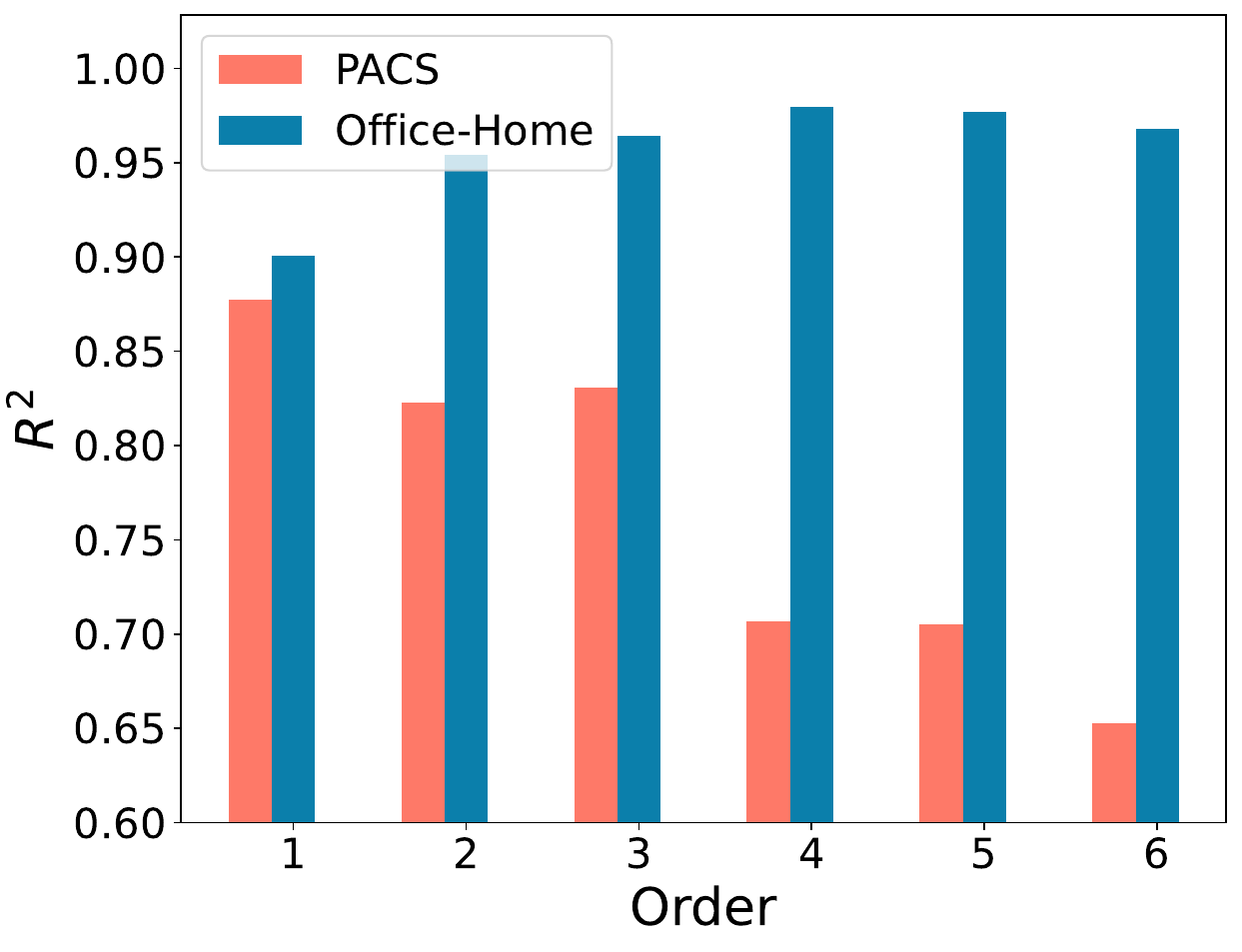}
        \label{fig:order}}
    \caption{\textbf{Empirical evidence with Resnet18.} \textbf{(a)} The model is well-calibrated on Office-Home and miscalibrated on PACS. (b) \normalization{} is superior to the state-of-the-art Nuclear~\citep{deng2023confidence} in all scenarios while the \texttt{softmax} heavily fails on PACS. \textbf{(c)} Increasing the approximation order $n$ in Eq.~\eqref{eq:taylor} is detrimental on PACS and beneficial on Office-Home. The optimal trade-off in all calibration scenarios is taking $ n \in \{2, 3\}$.}
     \vspace{-10pt}
\label{fig:normalization_calibration}
\end{figure}

\subsection{The Failure of Softmax Normalization Under Poorly-Calibrated Scenarios}
\label{sec:softmax-failure}
It is widely known that the \texttt{softmax} normalization can suffer from overconfidence issues~\citep{wei2022logitnorm, odonnat2024leveraging} and saturation of its outputs~\citep{chen2017NoisySI}, with one entry close to one while the others are close to zero. 
 
\paragraph{Analysis.} To alleviate those issues, we first notice that the \texttt{softmax} can be decomposed as $\softmax{\mbf{q}} = \exp(\mbf{q})/\sum_{k=1}^K \exp(\mbf{q}_k) = (\phi \circ \exp)(\mbf{q})$, where $\phi \colon \RR^K_+ \to \Delta_K$ writes $\phi(\mbf{u})= \mbf{u} / \sum_{k=1}^K\mbf{u}_k = \mbf{u} / \lVert\mbf{u}\rVert_1$. While $\phi$ has appealing property for normalization (see Proposition~\ref{prop:property_phi}), the exponential can accumulate prediction errors, leading to the \texttt{softmax} overconfidence and a biased accuracy estimation. In particular, assume that the $k$-th entry of the output of the neural network on a test sample $\mbf{x}_i$ writes $\mbf{q}^*_{i, k} + \varepsilon_k$, where $\mbf{q}^*_{i, k}$ are the ground-truth logits and $\varepsilon_k$ is the prediction error. Then, the $n^{\mrm{th}}$-order Taylor polynomial of the exponential writes
\begin{equation}
\label{eq:taylor}
    \exp(\mbf{q}^*_{i, k}+\varepsilon_k) \approx 1+(\mbf{q}^*_{i, k}+\varepsilon_k)+\frac{(\mbf{q}^*_{i, k}+\varepsilon_k)^2}{2!} + ... +\frac{(\mbf{q}^*_{i, k}+\varepsilon_k)^n}{n!}.
\end{equation}
Consequently, logit-based accuracy estimation methods using \texttt{softmax} are sensitive to prediction bias, leading to low-quality estimations in poorly calibrated scenarios. 
\paragraph{Empirical evidence.} We illustrate this phenomenon in Figure~\ref{fig:reliability} on two datasets, where a pre-trained ResNet18 exhibits more pronounced calibration issues on PACS~\citep{li2017deeper} compared to Office Home~\citep{venkateswara2017deep}. Figure~\ref{fig:norm_selection} shows that using \method{} with \texttt{softmax} normalization is appropriate on Office-Home where the model is well calibrated but not in a miscalibrated scenario on PACS. Conversely, using \method{} with $2^{\mrm{nd}}$-order Taylor approximation is appropriate under miscalibration on PACS but not on Office-Home. In both cases, we see that \method{} can surpass the state-of-the-art method Nuclear~\citep{deng2023confidence} provided it uses the appropriate normalization $\sigma$. Figure~\ref{fig:order} illustrates the impact of truncating Eq.~\eqref{eq:taylor} up to the $n$-th order. We conclude that a trade-off is needed between \emph{information completeness on true logits and error accumulation} depending on the type of calibration scenario. Specifically, when the model is poorly calibrated on a given dataset (\textit{i.e.}, $\varepsilon_k$ large in absolute value), the normalization should focus on avoiding error accumulation, and when the model is well calibrated (\textit{i.e.}, $\varepsilon_k$ small in absolute value), the normalization should focus on information completeness. 

\subsection{Softrun: The Proposed Normalization Strategy}
The above analysis shows that different calibration scenarios emphasize different information during normalization. Therefore, we propose a normalization strategy called \normalization{} that normalizes the model outputs based on the calibration scenario. Given logits $\mbf{q}_i \in \RR^K$ and reusing the function $\phi$ previously introduced, it takes the general form:
\begin{equation}
\label{eq:normalization}
    \sigma(\mbf{q}_i) = (\phi \circ v)(\mbf{q}_i) = \frac{v(\mbf{q}_i)}{\sum_{k=1}^K v(\mbf{q}_i)_k} \in \Delta_K.
\end{equation}
where $v \colon \RR^K \to \RR^K_+$ is designed to avoid error accumulation under poorly-calibrated scenarios by truncating the exponential (Taylor $n=2$ in Eq.~\eqref{eq:taylor}) and using complete logits information under well-calibrated scenarios (\texttt{softmax}). As in practice, the calibration of the model on test data is unknown, \normalization{} employs a simple yet effective strategy reminiscent of pseudo-labeling~\citep{lee2013pl, sohn2020fixmatch}. More specifically, given a test dataset $\mcal{D}_\mrm{test} = \{\mbf{x}_i\}_{i=1}^N$ and corresponding logits $\mbf{q}_i = f(\mbf{x}_i)$, a criterion $\threshold{}$ is computed at the dataset level and the normalized logits are defined as\footnote{In practice if Taylor is applied, we replace $v(\mbf{q}_i)$ by $v(\mbf{q}_i) - \min v(\mbf{q}_i)$ to make sure the final outputs have nonnegative entries. This is especially needed for approximation orders $n \geq 3$.}
\begin{equation}
\label{eq:criterion}
    v(\mbf{q}_i) = \begin{cases}
1+\mbf{q}_i+\frac{\mbf{q}_i^2}{2}, &\text{if } \threshold{} \leq \eta \qquad \text{(Taylor)}\\
\exp(\mbf{q}_i), &\text{if } \threshold{} > \eta \qquad \text{(\texttt{softmax})}
    \end{cases}.
\end{equation}
We define $\threshold{} = -\frac{1}{NK} \sum_{j=1}^N\sum_{k=1}^K \log(\frac{\exp(\mbf{q}_j)_k}{\sum_{j=1}^K\exp(\mbf{q}_i)_j)})$, which is equal, up to a constant, to the average KL divergence between the uniform distribution and the predicted \texttt{softmax} probabilities. It follows from~\citet{tian2021exploring} that showed that this KL divergence was small when the uncertainty of the model was high and large for confident models. Hence, when the model is uncertain, \textit{i.e.}, $\threshold{} \leq \eta$, we truncate the exponential to reduce error accumulation, and when the model is certain, \textit{i.e.}, $\threshold{} > \eta$, complete information is used with the exact exponential (and we recover the \texttt{softmax}). Thus, \normalization{} is designed to treat the problem with an additional level of complexity often overlooked by previous methods. While this comes at the cost of introducing the hyperparameter $\eta$, we fix $\eta=5$ across all our experiments. This, along with the design of $\normalization{}$, is justified both theoretically in Appendix~\ref{app:discuss_softrun} and experimentally in  Appendix~\ref{app:eta}.

\section{Experiments}
\label{sec:experiments}
In this section, we conduct experiments with \method{} that uses \normalization{} to properly normalize logits.
\subsection{Experimental Setup}
\paragraph{Pre-training datasets.} For pre-training the neural network, we use a diverse set of datasets including CIFAR-10, CIFAR-100 \citep{krizhevsky2009learning}, TinyImageNet \citep{le2015tiny}, ImageNet \citep{deng2009imagenet}, PACS \citep{li2017deeper}, Office-Home \citep{venkateswara2017deep}, DomainNet \citep{peng2019moment} and RR1-WILDS \citep{taylor2019rxrx1}, and BREEDS \citep{santurkar2020breeds} which leverages class hierarchy of ImageNet \citep{deng2009imagenet} to create $4$ datasets including Living-17, Nonliving-26, Entity-13 and Entity-30. 

\paragraph{Test datasets.} In our comprehensive evaluation, we consider $12$ datasets with $3$ types of distribution shifts: the synthetic, the natural, and the subpopulation shifts. To verify the effectiveness of our method under the synthetic shift, we use CIFAR-10C, CIFAR-100C, and ImageNet-C \citep{hendrycks2019benchmarking} that span $19$ types of corruption across $5$ severity levels, as well as TinyImageNet-C \citep{hendrycks2019benchmarking} with $15$ types of corruption and $5$ severity levels. For the natural shift, we use the domains excluded from training from PACS, Office-Home, DomainNet, and RR1-WILDS as the OOD datasets. For the novel subpopulation shift, we consider the BREEDS benchmark with Living-17, Nonliving-26, Entity-13, and Entity-30 which were constructed from ImageNet-C.

\paragraph{Training details.} To show the versatility of our method across different architectures, we perform experiments with ResNet18, ResNet50 \citep{he2016deep}, and WRN-50-2 \citep{zagoruyko2016wide} models. We train them for $20$ epochs for CIFAR-10 \citep{krizhevsky2009learning} and $50$ epochs for the other datasets. In all cases, we use SGD with a learning rate of $10^{-3}$, cosine learning rate decay \citep{loshchilov2016sgdr}, a momentum of $0.9$, and a batch size of $128$. 

\paragraph{Evaluation metrics.} We use the coefficient of determination $R^2 \in [0, 1]$~\citep{nagelkerke1991note} and the Spearman’s rank correlation coefficient $\rho \in [-1, 1]$~\citep{kendall1948rank} to evaluate performance. The former measures the linearity and goodness of fit and $1$ indicates a perfect fit. $\rho$ measures monotonicity and values close to $\{-1, 1\}$ indicate strong correlation while $0$ indicates no correlation.
 \begin{table*}[!t]
    \centering
    \caption{Method comparison on four benchmarks using ResNet18, ResNet50, and WRN-50-2 under the \textbf{synthetic shift}, where $R^2$ refers to coefficients of determination, and $\rho$ refers to the absolute value of Spearman correlation coefficients (higher is better). The best results for each metric are in \textbf{bold}. Overall, \method{} achieves the highest $R^2$ and $\rho$ values across different datasets and architectures, indicating its superior performance.
    }
    \renewcommand\arraystretch{1.2}
    \resizebox{0.99\textwidth}{!}{
    \setlength{\tabcolsep}{1.2mm}{
    \begin{tabular}{cccccccccccccccccccccccccc}
        \toprule
        \multicolumn{26}{c}{Synthetic Shift} \\
        \toprule
        \multirow{2}{*}{Dataset} &\multirow{2}{*}{Model} &\multicolumn{2}{c}{Rotation} &\multicolumn{2}{c}{ConfScore} &\multicolumn{2}{c}{Entropy} &\multicolumn{2}{c}{AgreeScore} &\multicolumn{2}{c}{ATC} &\multicolumn{2}{c}{Fr\'{e}chet} &\multicolumn{2}{c}{Dispersion}&\multicolumn{2}{c}{ProjNorm} &\multicolumn{2}{c}{MDE} &\multicolumn{2}{c}{COT} &\multicolumn{2}{c}{Nuclear} &\multicolumn{2}{c}{\method{}}\\
        \cline{3-26}
        & &$R^2$ &$\rho$ &$R^2$ &$\rho$&$R^2$ &$\rho$&$R^2$ &$\rho$&$R^2$ &$\rho$&$R^2$ &$\rho$&$R^2$ &$\rho$ &$R^2$ &$\rho$&$R^2$ &$\rho$&$R^2$ &$\rho$&$R^2$ &$\rho$&$R^2$ &$\rho$\\
        \midrule
         \multirow{4}{*}{CIFAR 10} & ResNet18 &0.822 &0.951 &0.869 &0.985 &0.899 &0.987 &0.663 &0.929 &0.884 &0.985 &0.950 &0.971 &0.968 &0.990 &0.936 &0.982 &0.957 &0.987 &0.989 &0.995 &\textbf{0.995} &\textbf{0.997} &\textbf{0.995} &\textbf{0.997}\\
          & ResNet50 &0.835 &0.961 &0.935 &0.993 &0.945 &0.994 &0.835 &0.985 &0.946 &0.994 &0.858 &0.964 &0.987 &0.990 &0.944 &0.989 &0.978 &0.963 &0.984 &0.996 &0.994 &0.996 &\textbf{0.996} &\textbf{0.997}\\
          & WRN-50-2 &0.862 &0.976 &0.943 &0.994 &0.942 &0.994 &0.856 &0.986 &0.947 &0.994 &0.814 &0.973 &0.962 &0.988 &0.961 &0.989 &0.930 &0.809 &0.988 &0.994 &0.994 &0.995 &\textbf{0.996} &\textbf{0.992}\\
          \cline{2-26}
          & \textcolor[rgb]{0.0, 0.53, 0.74}{Average} &\textcolor[rgb]{0.0, 0.53, 0.74}{0.840} &\textcolor[rgb]{0.0, 0.53, 0.74}{0.963} &\textcolor[rgb]{0.0, 0.53, 0.74}{0.916} &\textcolor[rgb]{0.0, 0.53, 0.74}{0.991} &\textcolor[rgb]{0.0, 0.53, 0.74}{0.930} &\textcolor[rgb]{0.0, 0.53, 0.74}{0.992} &\textcolor[rgb]{0.0, 0.53, 0.74}{0.785} &\textcolor[rgb]{0.0, 0.53, 0.74}{0.967} &\textcolor[rgb]{0.0, 0.53, 0.74}{0.926} &\textcolor[rgb]{0.0, 0.53, 0.74}{0.991} &\textcolor[rgb]{0.0, 0.53, 0.74}{0.874} &\textcolor[rgb]{0.0, 0.53, 0.74}{0.970} &\textcolor[rgb]{0.0, 0.53, 0.74}{\textbf{0.972}} &\textcolor[rgb]{0.0, 0.53, 0.74}{0.990} &\textcolor[rgb]{0.0, 0.53, 0.74}{0.947} &\textcolor[rgb]{0.0, 0.53, 0.74}{0.987} &\textcolor[rgb]{0.0, 0.53, 0.74}{0.955}&\textcolor[rgb]{0.0, 0.53, 0.74}{0.920}&\textcolor[rgb]{0.0, 0.53, 0.74}{0.987}&\textcolor[rgb]{0.0, 0.53, 0.74}{0.995} &\textcolor[rgb]{0.0, 0.53, 0.74}{0.995} &\textcolor[rgb]{0.0, 0.53, 0.74}{\textbf{0.996}} &\textcolor[rgb]{0.0, 0.53, 0.74}{\textbf{0.996}} &\textcolor[rgb]{0.0, 0.53, 0.74}{0.995} \\
          \midrule
    \multirow{4}{*}{CIFAR 100} & ResNet18 &0.860 &0.936 &0.916 &0.985 &0.891 &0.979 &0.902 &0.973 &0.938 &0.986 &0.888 &0.968 &0.952 &0.988 &0.979 &0.980 &0.975 &0.994 &0.991 &0.995 &0.989 &0.995 &\textbf{0.996} &\textbf{0.996}\\
          & ResNet50 &0.908 &0.962 &0.919 &0.984 &0.884 &0.977 &0.922 &0.982 &0.921 &0.984 &0.837 &0.972 &0.951 &0.985 &0.988 &0.991 &0.988 &0.995 &0.985 &0.996 &0.979 &0.994 &\textbf{0.995} &\textbf{0.997}\\
          & WRN-50-2 &0.924 &0.970 &0.971 &0.984 &0.968 &0.981 &0.955 &0.977 &0.978 &0.993 &0.865 &0.987 &0.980 &0.991 &0.990 &0.991 &0.995 &0.994 &0.987 &0.997 &0.962 &0.988 &\textbf{0.996} &\textbf{0.998}\\
          \cline{2-26}
          & \textcolor[rgb]{0.0, 0.53, 0.74}{Average} &\textcolor[rgb]{0.0, 0.53, 0.74}{0.898} &\textcolor[rgb]{0.0, 0.53, 0.74}{0.956} &\textcolor[rgb]{0.0, 0.53, 0.74}{0.936} &\textcolor[rgb]{0.0, 0.53, 0.74}{0.987} &\textcolor[rgb]{0.0, 0.53, 0.74}{0.915} &\textcolor[rgb]{0.0, 0.53, 0.74}{0.983} &\textcolor[rgb]{0.0, 0.53, 0.74}{0.927} &\textcolor[rgb]{0.0, 0.53, 0.74}{0.982} &\textcolor[rgb]{0.0, 0.53, 0.74}{0.946} &\textcolor[rgb]{0.0, 0.53, 0.74}{0.988} &\textcolor[rgb]{0.0, 0.53, 0.74}{0.864} &\textcolor[rgb]{0.0, 0.53, 0.74}{0.976} &\textcolor[rgb]{0.0, 0.53, 0.74}{0.962} &\textcolor[rgb]{0.0, 0.53, 0.74}{0.988} &\textcolor[rgb]{0.0, 0.53, 0.74}{0.985} &\textcolor[rgb]{0.0, 0.53, 0.74}{0.987}  &\textcolor[rgb]{0.0, 0.53, 0.74}{0.986}&\textcolor[rgb]{0.0, 0.53, 0.74}{0.994}&\textcolor[rgb]{0.0, 0.53, 0.74}{0.988}&\textcolor[rgb]{0.0, 0.53, 0.74}{0.996} &\textcolor[rgb]{0.0, 0.53, 0.74}{0.977} &\textcolor[rgb]{0.0, 0.53, 0.74}{0.993} &\textcolor[rgb]{0.0, 0.53, 0.74}{\textbf{0.996}} &\textcolor[rgb]{0.0, 0.53, 0.74}{\textbf{0.997}}\\
          \midrule
   \multirow{4}{*}{TinyImageNet} & ResNet18 &0.786 &0.946 &0.670 &0.869 &0.592 &0.842 &0.561 &0.853 &0.751 &0.945 &0.826 &0.970 &0.966 &0.986 &0.970 &0.981 &0.941 &0.993 &\textbf{0.985} &0.994 &0.983 &0.994 &0.981 &\textbf{0.996}\\
          & ResNet50 &0.786 &0.947 &0.670 &0.869 &0.651 &0.892 &0.560 &0.853 &0.751 &0.945 &0.826 &0.971 &0.977 &0.986 &0.979 &0.987 &0.941 &0.993 &\textbf{0.980} &0.994 &0.965 &0.994 &\textbf{0.980} &\textbf{0.996}\\
          & WRN-50-2 &0.878 &0.967 &0.757 &0.951 &0.704 &0.935 &0.654 &0.904 &0.635 &0.897 &0.884 &0.984 &0.968 &0.986 &0.965 &0.983 &0.961 &0.996 &\textbf{0.985} &0.997 &0.962 &0.988 &0.979 &\textbf{0.997}\\
          \cline{2-26}
          & \textcolor[rgb]{0.0, 0.53, 0.74}{Average} &\textcolor[rgb]{0.0, 0.53, 0.74}{0.805} &\textcolor[rgb]{0.0, 0.53, 0.74}{0.959} &\textcolor[rgb]{0.0, 0.53, 0.74}{0.727} &\textcolor[rgb]{0.0, 0.53, 0.74}{0.920} &\textcolor[rgb]{0.0, 0.53, 0.74}{0.650} &\textcolor[rgb]{0.0, 0.53, 0.74}{0.890} &\textcolor[rgb]{0.0, 0.53, 0.74}{0.599} &\textcolor[rgb]{0.0, 0.53, 0.74}{0.878} &\textcolor[rgb]{0.0, 0.53, 0.74}{0.693} &\textcolor[rgb]{0.0, 0.53, 0.74}{0.921} &\textcolor[rgb]{0.0, 0.53, 0.74}{0.847} &\textcolor[rgb]{0.0, 0.53, 0.74}{0.976} &\textcolor[rgb]{0.0, 0.53, 0.74}{0.970} &\textcolor[rgb]{0.0, 0.53, 0.74}{0.987} &\textcolor[rgb]{0.0, 0.53, 0.74}{0.972} &\textcolor[rgb]{0.0, 0.53, 0.74}{0.984}  &\textcolor[rgb]{0.0, 0.53, 0.74}{0.950}&\textcolor[rgb]{0.0, 0.53, 0.74}{0.995}&\textcolor[rgb]{0.0, 0.53, 0.74}{\textbf{0.984}}&\textcolor[rgb]{0.0, 0.53, 0.74}{0.995} &\textcolor[rgb]{0.0, 0.53, 0.74}{0.968} &\textcolor[rgb]{0.0, 0.53, 0.74}{0.993} &\textcolor[rgb]{0.0, 0.53, 0.74}{0.980} &\textcolor[rgb]{0.0, 0.53, 0.74}{\textbf{0.996}}\\
          \midrule
    \multirow{4}{*}{ImageNet} & ResNet18 &- &- &0.979 &0.991 &0.963 &0.991 &- &- &0.974 &0.983 &0.802 &0.974  &0.940 &0.971 &0.975 &0.993 &0.924 &0.994 &\textbf{0.996} &\textbf{0.998} &0.992 &0.997 &0.992 &0.997\\
          & ResNet50 &- &- &0.980 &0.994 &0.967 &0.992 &- &- &0.970 &0.983 &0.855 &0.974 &0.938 &0.968 &0.986 &0.993 &0.886 &0.994 &\textbf{0.993} &0.996 &0.985 &0.997 &0.991 &\textbf{0.998}\\
          & WRN-50-2 &- &- &0.983 &0.991 &0.963 &0.991 &- &- &0.983 &0.993 &0.909 &0.988 &0.939 &0.976 &0.978 &0.993 &0.880 &0.997 &0.989 &0.994 &0.987 &0.998 &\textbf{0.996} &\textbf{0.998}\\
          \cline{2-26}
          & \textcolor[rgb]{0.0, 0.53, 0.74}{Average} &\textcolor[rgb]{0.0, 0.53, 0.74}{-} &\textcolor[rgb]{0.0, 0.53, 0.74}{-} &\textcolor[rgb]{0.0, 0.53, 0.74}{0.981} &\textcolor[rgb]{0.0, 0.53, 0.74}{0.993} &\textcolor[rgb]{0.0, 0.53, 0.74}{0.969} &\textcolor[rgb]{0.0, 0.53, 0.74}{0.992} &\textcolor[rgb]{0.0, 0.53, 0.74}{-} &\textcolor[rgb]{0.0, 0.53, 0.74}{-} &\textcolor[rgb]{0.0, 0.53, 0.74}{0.976} &\textcolor[rgb]{0.0, 0.53, 0.74}{0.987} &\textcolor[rgb]{0.0, 0.53, 0.74}{0.855} &\textcolor[rgb]{0.0, 0.53, 0.74}{0.979} &\textcolor[rgb]{0.0, 0.53, 0.74}{0.939} &\textcolor[rgb]{0.0, 0.53, 0.74}{0.972} &\textcolor[rgb]{0.0, 0.53, 0.74}{0.980} &\textcolor[rgb]{0.0, 0.53, 0.74}{0.993}   &\textcolor[rgb]{0.0, 0.53, 0.74}{0.897}&\textcolor[rgb]{0.0, 0.53, 0.74}{0.995}&\textcolor[rgb]{0.0, 0.53, 0.74}{0.993}&\textcolor[rgb]{0.0, 0.53, 0.74}{0.996} &\textcolor[rgb]{0.0, 0.53, 0.74}{0.988} &\textcolor[rgb]{0.0, 0.53, 0.74}{0.998} &\textcolor[rgb]{0.0, 0.53, 0.74}{\textbf{0.993}} &\textcolor[rgb]{0.0, 0.53, 0.74}{\textbf{0.998}}\\   
         \bottomrule
     \end{tabular}}}
     \vspace{-8pt}
    \label{tab:main 1}
\end{table*}
\paragraph{Baselines.} We consider $11$ baselines commonly evaluated in the unsupervised accuracy estimation studies, including \textit{Rotation Prediction} (Rotation) \citep{deng2021does}, \textit{Averaged Confidence} (ConfScore) \citep{hendrycks2016baseline}, \textit{Entropy} \citep{guillory2021predicting}, \textit{Agreement Score} (AgreeScore) \citep{jiang2021assessing}, \textit{Averaged Threshold Confidence} (ATC) \citep{garg2022leveraging}, \textit{AutoEval} (Fr\'{e}chet) \citep{deng2021labels}, \textit{ProjNorm} \citep{yu2022predicting}, \textit{Dispersion Score} (Dispersion) \citep{xie2023importance}, MDE \citep{peng2024energy}, COT \citep{lu2024characterizing}, and \textit{Nuclear Norm} (Nuclear) \citep{deng2023confidence}. 

\subsection{Main Takeaways} 
\begin{figure}[!h]
    \includegraphics[width=\linewidth]{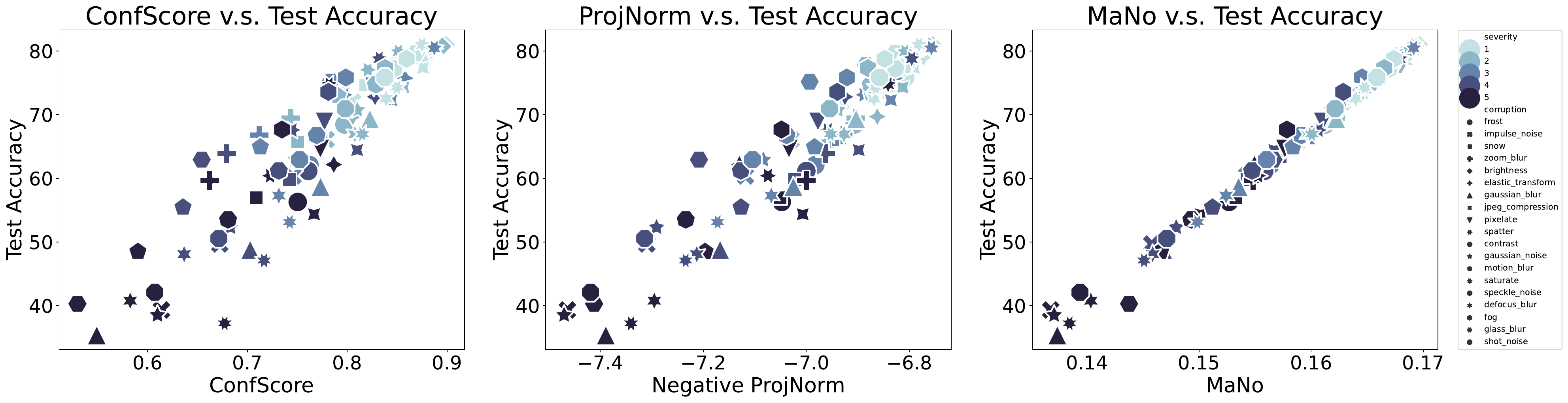}
    \vspace{-10pt}
    \caption{OOD error prediction versus ground-truth error on Entity-13 with ResNet18. This scatter plot compares \method{} with Dispersion Score and ProjNorm. Each point represents one dataset under a specific type and severity of corruption. Different shapes indicate different types of corruption, while darker colors indicate higher severity levels. This indicates the qualitative superiority of \method{}.}
    \label{fig:scatters}
    \vspace{-8pt}
\end{figure}

\begin{figure}[!h]
\centering
\begin{minipage}{.44\linewidth}
  \centering
  \includegraphics[height=0.7\linewidth]{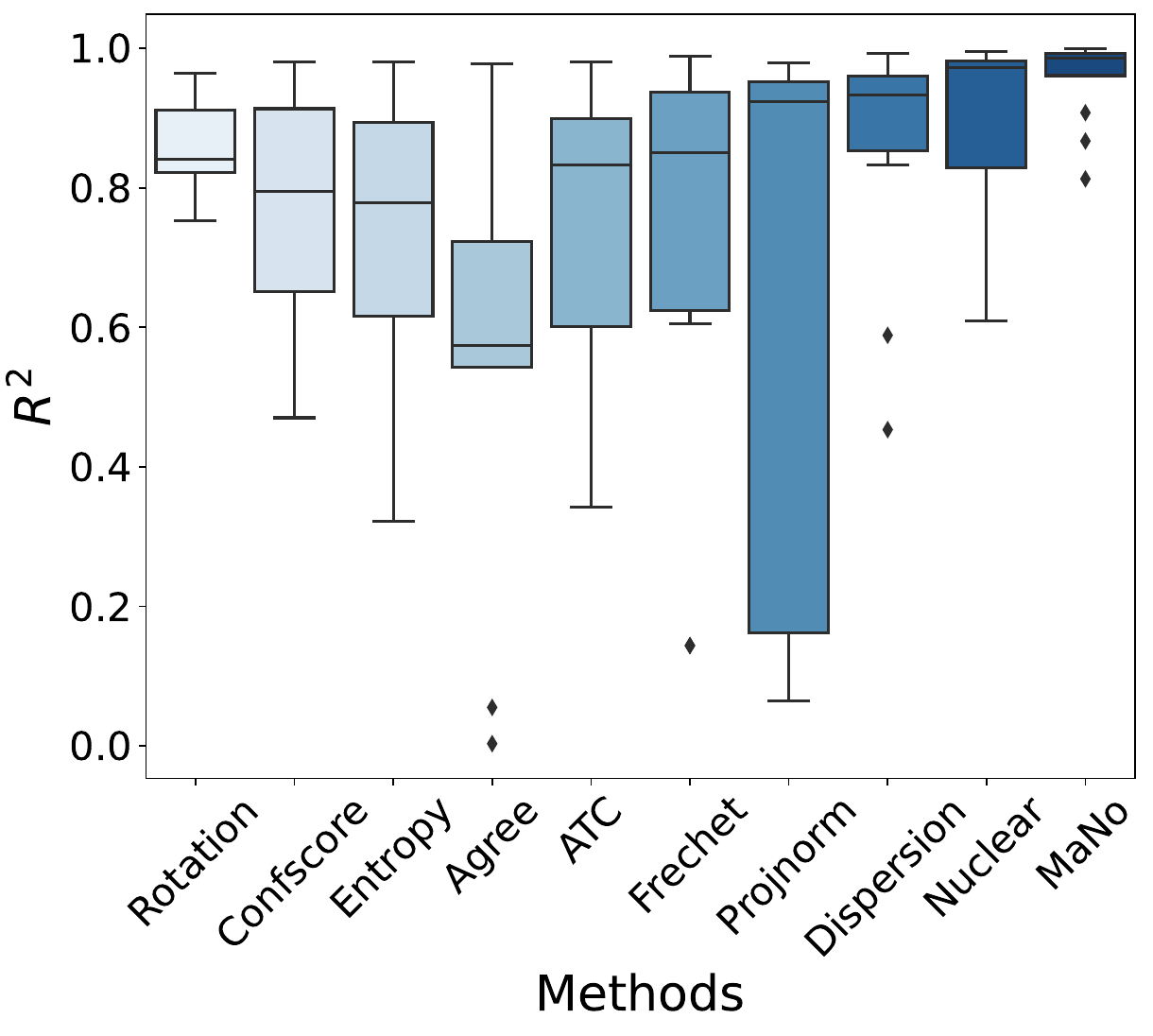}
  \captionof{figure}{$R^2$ distribution with ResNet18 on all distribution shifts. Overall, \method{} leads to the best and most robust estimations.}
  \label{fig:boxplot}
\end{minipage}%
\qquad
\begin{minipage}[h]{.44\linewidth}
  \centering
  \includegraphics[height=0.7\linewidth]{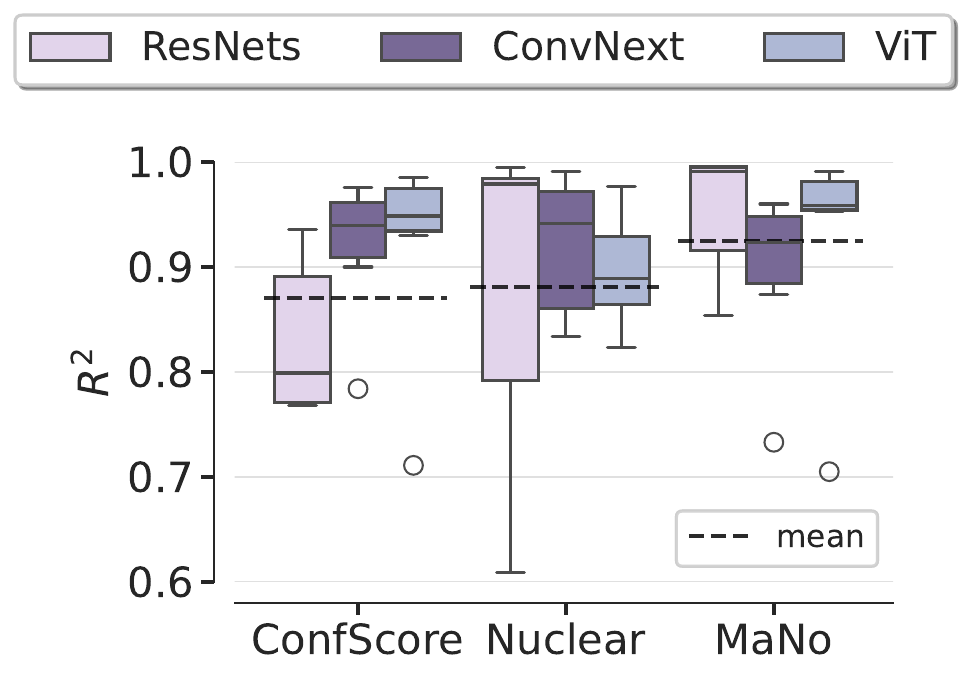}
  \captionof{figure}{$R^2$ distribution using ResNets (average), ConvNext, or ViT on all distribution shifts. Again, \method{} is the best method.}
  \label{fig:all_arch}
\end{minipage}
\vspace{-8pt}
\end{figure}

\paragraph{\method{} improves over state-of-the-art.} Tables~\ref{tab:main 1} and \ref{tab:main 3} present the numerical results of unsupervised accuracy estimation across $8$ datasets using $3$ different network architectures, evaluated under synthetic and subpopulation shifts. These shifts are quantified by $R^2$ and $\rho$. Empirical results demonstrate that these distribution shifts do not significantly impact calibration (\textit{i.e.}, $\threshold{} > \eta$). We observe that \method{} consistently outperforms other baselines, achieving state-of-the-art performance. For instance, \method{} achieves $R^2 > 0.960$ and $\rho > 0.990$ under subpopulation shift, whereas the performance of other baselines does not reach such consistently high levels.
\begin{table*}[!t]
    \centering
    \caption{ Method comparison on four benchmarks using ResNet18, ResNet50, and WRN-50-2 under \textbf{subpopulation shift} with $R^2$ and $\rho$ metrics (the higher the better).
    The best results for each metric are in \textbf{bold}. Overall, \method{} surpasses all its competitors.}
    \renewcommand\arraystretch{1.2}
    \resizebox{0.99\textwidth}{!}{
    \setlength{\tabcolsep}{1.2mm}{
    \begin{tabular}{cccccccccccccccccccccccccc}
        \toprule
        \multicolumn{26}{c}{Subpopulation Shift}\\
        \toprule
        \multirow{2}{*}{Dataset} &\multirow{2}{*}{Model} &\multicolumn{2}{c}{Rotation} &\multicolumn{2}{c}{ConfScore} &\multicolumn{2}{c}{Entropy} &\multicolumn{2}{c}{AgreeScore} &\multicolumn{2}{c}{ATC} &\multicolumn{2}{c}{Fr\'{e}chet} &\multicolumn{2}{c}{Dispersion}&\multicolumn{2}{c}{ProjNorm} &\multicolumn{2}{c}{MDE} &\multicolumn{2}{c}{COT} &\multicolumn{2}{c}{Nuclear} &\multicolumn{2}{c}{\method{}}\\
        \cline{3-26}
        & &$R^2$ &$\rho$ &$R^2$ &$\rho$&$R^2$ &$\rho$&$R^2$ &$\rho$&$R^2$ &$\rho$&$R^2$ &$\rho$&$R^2$ &$\rho$ &$R^2$ &$\rho$&$R^2$ &$\rho$&$R^2$ &$\rho$&$R^2$ &$\rho$&$R^2$ &$\rho$\\
        \midrule
           \multirow{4}{*}{Entity-13} & ResNet18 &0.927 &0.961 &0.795 &0.940 &0.794 &0.935 &0.543 &0.919 &0.823 &0.945 &0.950 &0.981 &0.937 &0.968 &0.952 &0.981 &0.927 &0.995 &0.960 &0.985 &0.978 &0.991 &\textbf{0.992} &\textbf{0.996}\\
          & ResNet50 &0.932 &0.976 &0.728 &0.941 &0.698 &0.928 &0.901 &0.964 &0.783 &0.950 &0.903 &0.959 &0.764 &0.892 &0.944 &0.974 &0.912 &0.993 &0.935 &0.971 &0.989 &0.996 &\textbf{0.993} &\textbf{0.998}\\
          & WRN-50-2 &0.939 &0.983 &0.930 &0.977 &0.919 &0.973 &0.871 &0.935 &0.936 &0.980 &0.906 &0.958 &0.815 &0.905 &0.950 &0.977 &0.925 &0.995 &0.944 &0.979 &0.989 &0.995 &\textbf{0.992} &\textbf{0.996}\\
          \cline{2-26}
          & \textcolor[rgb]{0.0, 0.53, 0.74}{Average} & \textcolor[rgb]{0.0, 0.53, 0.74}{0.933} & \textcolor[rgb]{0.0, 0.53, 0.74}{0.973} & \textcolor[rgb]{0.0, 0.53, 0.74}{0.817} & \textcolor[rgb]{0.0, 0.53, 0.74}{0.953} & \textcolor[rgb]{0.0, 0.53, 0.74}{0.804} & \textcolor[rgb]{0.0, 0.53, 0.74}{0.945} & \textcolor[rgb]{0.0, 0.53, 0.74}{0.772} & \textcolor[rgb]{0.0, 0.53, 0.74}{0.939} & \textcolor[rgb]{0.0, 0.53, 0.74}{0.847} & \textcolor[rgb]{0.0, 0.53, 0.74}{0.958} & \textcolor[rgb]{0.0, 0.53, 0.74}{0.920} & \textcolor[rgb]{0.0, 0.53, 0.74}{0.966} & \textcolor[rgb]{0.0, 0.53, 0.74}{0.948} & \textcolor[rgb]{0.0, 0.53, 0.74}{0.977} & \textcolor[rgb]{0.0, 0.53, 0.74}{0.839} & \textcolor[rgb]{0.0, 0.53, 0.74}{0.922} & \textcolor[rgb]{0.0, 0.53, 0.74}{0.921} & \textcolor[rgb]{0.0, 0.53, 0.74}{0.995} & \textcolor[rgb]{0.0, 0.53, 0.74}{0.947}  & \textcolor[rgb]{0.0, 0.53, 0.74}{0.979}  & \textcolor[rgb]{0.0, 0.53, 0.74}{0.985} & \textcolor[rgb]{0.0, 0.53, 0.74}{0.994} &\textcolor[rgb]{0.0, 0.53, 0.74}{\textbf{0.993}} &\textcolor[rgb]{0.0, 0.53, 0.74}{\textbf{0.996}}\\
          \midrule
           \multirow{4}{*}{Entity-30} & ResNet18 &0.964 &0.979 &0.570 &0.836 &0.553 &0.832 &0.542 &0.935 &0.611 &0.845 &0.849 &0.978 &0.929 &0.968 &0.952 &0.987 &0.931 &0.994 &0.971 &0.993 &0.980 &0.993 &\textbf{0.991} &\textbf{0.996}\\
          & ResNet50 &0.961 &0.980 &0.878 &0.969 &0.838 &0.956 &0.914 &0.975 &0.924 &0.973 &0.835 &0.956 &0.783 &0.914 &0.937 &0.986 &0.918 &0.995 &0.958 &0.982 &0.978 &0.994 &\textbf{0.988} &\textbf{0.997}\\
          & WRN-50-2 &0.940 &0.978 &0.897 &0.974 &0.878 &0.970 &0.826 &0.955 &0.936 &0.984 &0.927 &0.973 &0.927 &0.973 &0.959 &0.986 &0.925 &0.995 &0.944 &0.979 &0.985 &0.996 &\textbf{0.988} &\textbf{0.997}\\
          \cline{2-26}
          & \textcolor[rgb]{0.0, 0.53, 0.74}{Average} & \textcolor[rgb]{0.0, 0.53, 0.74}{0.955} & \textcolor[rgb]{0.0, 0.53, 0.74}{0.978} & \textcolor[rgb]{0.0, 0.53, 0.74}{0.781} & \textcolor[rgb]{0.0, 0.53, 0.74}{0.926} & \textcolor[rgb]{0.0, 0.53, 0.74}{0.756} & \textcolor[rgb]{0.0, 0.53, 0.74}{0.919} & \textcolor[rgb]{0.0, 0.53, 0.74}{0.728} & \textcolor[rgb]{0.0, 0.53, 0.74}{0.956} & \textcolor[rgb]{0.0, 0.53, 0.74}{0.823} & \textcolor[rgb]{0.0, 0.53, 0.74}{0.934} & \textcolor[rgb]{0.0, 0.53, 0.74}{0.871} & \textcolor[rgb]{0.0, 0.53, 0.74}{0.969}& \textcolor[rgb]{0.0, 0.53, 0.74}{0.880} & \textcolor[rgb]{0.0, 0.53, 0.74}{0.952}  & \textcolor[rgb]{0.0, 0.53, 0.74}{0.949} & \textcolor[rgb]{0.0, 0.53, 0.74}{0.987}  & \textcolor[rgb]{0.0, 0.53, 0.74}{0.925}  & \textcolor[rgb]{0.0, 0.53, 0.74}{0.995}  & \textcolor[rgb]{0.0, 0.53, 0.74}{0.970}  & \textcolor[rgb]{0.0, 0.53, 0.74}{0.988} & \textcolor[rgb]{0.0, 0.53, 0.74}{0.981} & \textcolor[rgb]{0.0, 0.53, 0.74}{0.994} &\textcolor[rgb]{0.0, 0.53, 0.74}{\textbf{0.989}} &\textcolor[rgb]{0.0, 0.53, 0.74}{\textbf{0.996}}\\
          \midrule
           \multirow{4}{*}{Living-17}  & ResNet18 &0.876 &0.973 &0.913 &0.973 &0.898 &0.970 &0.586 &0.736 &0.940 &0.973 &0.768 &0.950 &0.900 &0.958 &0.923 &0.970 &0.927 &0.985 &0.972 &0.984 &0.975 &0.987 &\textbf{0.980} &\textbf{0.991}\\
          & ResNet50 &0.906 &0.956 &0.880 &0.967 &0.853 &0.961 &0.633 &0.802 &0.938 &0.976 &0.771 &0.926 &0.851 &0.929 &0.903 &0.924 &0.914 &0.985 &0.953 &0.973 &0.967 &0.976 &\textbf{0.975} &\textbf{0.997}\\
          & WRN-50-2 &0.909 &0.957 &0.928 &0.980 &0.921 &0.977 &0.652 &0.793 &0.966 &0.984 &0.931 &0.967 &0.931 &0.966 &0.915 &0.970 &0.914 &0.983 &0.965 &0.990 &0.951 &0.978 &\textbf{0.961} &\textbf{0.996}\\
          \cline{2-26}
          & \textcolor[rgb]{0.0, 0.53, 0.74}{Average} & \textcolor[rgb]{0.0, 0.53, 0.74}{0.933} & \textcolor[rgb]{0.0, 0.53, 0.74}{0.974} & \textcolor[rgb]{0.0, 0.53, 0.74}{0.907} & \textcolor[rgb]{0.0, 0.53, 0.74}{0.973} & \textcolor[rgb]{0.0, 0.53, 0.74}{0.814} & \textcolor[rgb]{0.0, 0.53, 0.74}{0.969} & \textcolor[rgb]{0.0, 0.53, 0.74}{0.623} & \textcolor[rgb]{0.0, 0.53, 0.74}{0.777} & \textcolor[rgb]{0.0, 0.53, 0.74}{\textbf{0.948}} & \textcolor[rgb]{0.0, 0.53, 0.74}{\textbf{0.978}} & \textcolor[rgb]{0.0, 0.53, 0.74}{0.817} & \textcolor[rgb]{0.0, 0.53, 0.74}{0.949} & \textcolor[rgb]{0.0, 0.53, 0.74}{0.894} & \textcolor[rgb]{0.0, 0.53, 0.74}{0.951} & \textcolor[rgb]{0.0, 0.53, 0.74}{0.913} & \textcolor[rgb]{0.0, 0.53, 0.74}{0.969} & \textcolor[rgb]{0.0, 0.53, 0.74}{0.918} & \textcolor[rgb]{0.0, 0.53, 0.74}{0.984} & \textcolor[rgb]{0.0, 0.53, 0.74}{0.963} & \textcolor[rgb]{0.0, 0.53, 0.74}{0.982}  & \textcolor[rgb]{0.0, 0.53, 0.74}{0.964} & \textcolor[rgb]{0.0, 0.53, 0.74}{0.980} &\textcolor[rgb]{0.0, 0.53, 0.74}{\textbf{0.972}} &\textcolor[rgb]{0.0, 0.53, 0.74}{\textbf{0.995}}\\
          \midrule
           \multirow{4}{*}{Nonliving-26}  & ResNet18 &0.906 &0.955 &0.781 &0.925 &0.739 &0.909 &0.543 &0.810 &0.854 &0.939 &0.914 &0.980 &0.958 &0.981 &0.939 &0.978 &0.929 &0.989 &\textbf{0.982} &\textbf{0.992} &0.970 &0.989 &0.978 &0.991\\
          & ResNet50 &0.916 &0.970 &0.832 &0.942 &0.776 &0.918 &0.638 &0.837 &0.893 &0.960 &0.848 &0.950 &0.805 &0.907 &0.873 &0.972 &0.907 &0.993 &0.962 &0.984 &0.956 &0.985 &\textbf{0.975} &\textbf{0.995}\\
          & WRN-50-2 &0.917 &0.977 &0.932 &0.971 &0.912 &0.959 &0.676 &0.861 &0.945 &0.969 &0.885 &0.942 &0.893 &0.939  &0.924 &0.973 &0.909 &0.991 &0.962 &0.979 &0.960 &0.988 &\textbf{0.978} &\textbf{0.992}\\
          \cline{2-26}
          & \textcolor[rgb]{0.0, 0.53, 0.74}{Average} & \textcolor[rgb]{0.0, 0.53, 0.74}{0.913} & \textcolor[rgb]{0.0, 0.53, 0.74}{0.967} & \textcolor[rgb]{0.0, 0.53, 0.74}{0.849} & \textcolor[rgb]{0.0, 0.53, 0.74}{0.946} & \textcolor[rgb]{0.0, 0.53, 0.74}{0.809} & \textcolor[rgb]{0.0, 0.53, 0.74}{0.929} & \textcolor[rgb]{0.0, 0.53, 0.74}{0.618} & \textcolor[rgb]{0.0, 0.53, 0.74}{0.836} & \textcolor[rgb]{0.0, 0.53, 0.74}{0.897} & \textcolor[rgb]{0.0, 0.53, 0.74}{0.956} & \textcolor[rgb]{0.0, 0.53, 0.74}{0.882} & \textcolor[rgb]{0.0, 0.53, 0.74}{0.957}& \textcolor[rgb]{0.0, 0.53, 0.74}{0.913}& \textcolor[rgb]{0.0, 0.53, 0.74}{0.974} & \textcolor[rgb]{0.0, 0.53, 0.74}{0.886} & \textcolor[rgb]{0.0, 0.53, 0.74}{0.943} & \textcolor[rgb]{0.0, 0.53, 0.74}{0.915} & \textcolor[rgb]{0.0, 0.53, 0.74}{0.991} & \textcolor[rgb]{0.0, 0.53, 0.74}{0.969} & \textcolor[rgb]{0.0, 0.53, 0.74}{0.985}  & \textcolor[rgb]{0.0, 0.53, 0.74}{0.962} & \textcolor[rgb]{0.0, 0.53, 0.74}{0.987} &\textcolor[rgb]{0.0, 0.53, 0.74}{\textbf{0.977}} &\textcolor[rgb]{0.0, 0.53, 0.74}{\textbf{0.992}}\\
         \bottomrule
    \end{tabular}}}
    \label{tab:main 3}
    \vspace{-8pt}
\end{table*}
\paragraph{\method{} significantly boosts performance under the natural shift.} Table~\ref{tab:main 2} illustrates the results of accuracy estimation under the natural shift on four datasets. Under the challenging and natural shift that is more complex than the other distribution shifts, we empirically observe $\threshold{} \leq \eta$. From Table~\ref{tab:main 2}, we observe a significant improvement compared with the other baselines. In particular, most methods have an $R^2$ and a $\rho$ under $0.9$ while \method{} reaches higher values. In addition, our method achieves the best performance on average on all four datasets. To visualize the estimation performance, we provide the scatter plots for \textit{Dispersion Score}, \textit{ProjNorm} and \method{} in Figure~\ref{fig:scatters} on Entity-18 with ResNet18. We find that \method{} scores present a robust linear relationship with ground-truth OOD errors, while the other state-of-the-art baselines tend to exhibit a biased estimation of high test errors. 
\begin{table*}[!t]
    \centering
    \caption{Method comparison on four benchmarks using ResNet18, ResNet50 and WRN-50-2 under \textbf{natural shift} with $R^2$ and $\rho$ metrics (the higher the better). The best results for each metric are in \textbf{bold}. Overall, \method{} surpasses all the other baselines.}
    \renewcommand\arraystretch{1.2}
    \resizebox{0.99\textwidth}{!}{
    \setlength{\tabcolsep}{1.2mm}{
    \begin{tabular}{cccccccccccccccccccccccccc}
        \toprule
        \multicolumn{26}{c}{Natural Shift}\\
        \toprule
        \multirow{2}{*}{Dataset} &\multirow{2}{*}{Model} &\multicolumn{2}{c}{Rotation} &\multicolumn{2}{c}{ConfScore} &\multicolumn{2}{c}{Entropy} &\multicolumn{2}{c}{AgreeScore} &\multicolumn{2}{c}{ATC} &\multicolumn{2}{c}{Fr\'{e}chet} &\multicolumn{2}{c}{Dispersion}&\multicolumn{2}{c}{ProjNorm} &\multicolumn{2}{c}{MDE} &\multicolumn{2}{c}{COT} &\multicolumn{2}{c}{Nuclear} &\multicolumn{2}{c}{\method{}}\\
        \cline{3-26}
        & &$R^2$ &$\rho$ &$R^2$ &$\rho$&$R^2$ &$\rho$&$R^2$ &$\rho$&$R^2$ &$\rho$&$R^2$ &$\rho$&$R^2$ &$\rho$ &$R^2$ &$\rho$&$R^2$ &$\rho$&$R^2$ &$\rho$&$R^2$ &$\rho$&$R^2$ &$\rho$\\
        \midrule
          \multirow{4}{*}{PACS} & ResNet18 &0.822 &0.895 &0.594 &0.755 &0.624 &0.755 &0.613 &0.832 &0.514 &0.650 &0.624 &0.804 &0.832 &0.825 &0.161 &0.419 &0.003 &0.153 &0.790 &0.783 &0.609 &0.874  &\textbf{0.827} &\textbf{0.909}\\
          & ResNet50 &0.860 &0.923 &0.070 &0.069 &0.061 &0.055 &0.463 &0.622 &0.192 &0.265 &0.463 &0.622 &0.073 &0.167 &0.244 &0.587 &0.059 &0.104 &0.891 &0.790 &0.611 &0.888 &\textbf{0.923} &\textbf{0.958}\\
          & WRN-50-2 &0.865 &0.902 &0.646 &0.678 &0.629 &0.671 &0.377 &0.858 &0.752 &0.832 &0.558 &0.832 &0.111 &0.167 &0.474 &0.650 &0.072 &0.244 &0.890 &0.888 &0.607 &0.867 &\textbf{0.924} &\textbf{0.972}\\
          \cline{2-26}
          & \textcolor[rgb]{0.0, 0.53, 0.74}{Average} & \textcolor[rgb]{0.0, 0.53, 0.74}{0.849} & \textcolor[rgb]{0.0, 0.53, 0.74}{0.906} & \textcolor[rgb]{0.0, 0.53, 0.74}{0.437} & \textcolor[rgb]{0.0, 0.53, 0.74}{0.501} & \textcolor[rgb]{0.0, 0.53, 0.74}{0.438} & \textcolor[rgb]{0.0, 0.53, 0.74}{0.494} & \textcolor[rgb]{0.0, 0.53, 0.74}{0.488} & \textcolor[rgb]{0.0, 0.53, 0.74}{0.770} & \textcolor[rgb]{0.0, 0.53, 0.74}{0.486} & \textcolor[rgb]{0.0, 0.53, 0.74}{0.582} & \textcolor[rgb]{0.0, 0.53, 0.74}{0.548} & \textcolor[rgb]{0.0, 0.53, 0.74}{0.337} & \textcolor[rgb]{0.0, 0.53, 0.74}{0.338} & \textcolor[rgb]{0.0, 0.53, 0.74}{0.275} & \textcolor[rgb]{0.0, 0.53, 0.74}{0.293} & \textcolor[rgb]{0.0, 0.53, 0.74}{0.552}  & \textcolor[rgb]{0.0, 0.53, 0.74}{0.045} & \textcolor[rgb]{0.0, 0.53, 0.74}{0.065} & \textcolor[rgb]{0.0, 0.53, 0.74}{0.857} & \textcolor[rgb]{0.0, 0.53, 0.74}{0.820} & \textcolor[rgb]{0.0, 0.53, 0.74}{0.609} & \textcolor[rgb]{0.0, 0.53, 0.74}{0.876} & \textcolor[rgb]{0.0, 0.53, 0.74}{\textbf{0.891}} & \textcolor[rgb]{0.0, 0.53, 0.74}{\textbf{0.946}}\\
          \midrule
        
    \multirow{4}{*}{Office-Home} & ResNet18 &0.822 &0.930 &0.795 &0.909 &0.761 &0.881 &0.054 &0.146 &0.571 &0.615 &0.605 &0.755 &0.453 &0.664  &0.064 &0.202 &0.331 &0.650 &0.863 &0.874 &0.692 &0.783 &\textbf{0.926} &\textbf{0.930}\\
          & ResNet50 &\textbf{0.851} &\textbf{0.944} &0.769 &0.895 &0.742 &0.853 &0.026 &0.216 &0.487 &0.734 &0.607 &0.685 &0.383 &0.727 &0.169 &0.475 &0.342 &0.622 &0.762 &0.846 &0.731 &0.895  &0.838 &0.916\\
          & WRN-50-2 &0.823 &\textbf{0.958} &0.741 &0.874 &0.696 &0.846 &0.132 &0.405 &0.383 &0.643 &0.589 &0.706 &0.456 &0.713 &0.172 &0.531 &0.342 &0.650 &\textbf{0.863} &0.874 &0.766 &0.874  &0.800 &0.895\\
          \cline{2-26}
          & \textcolor[rgb]{0.0, 0.53, 0.74}{Average} & \textcolor[rgb]{0.0, 0.53, 0.74}{0.832} & \textcolor[rgb]{0.0, 0.53, 0.74}{\textbf{0.944}} & \textcolor[rgb]{0.0, 0.53, 0.74}{0.768} & \textcolor[rgb]{0.0, 0.53, 0.74}{0.892} & \textcolor[rgb]{0.0, 0.53, 0.74}{0.733} & \textcolor[rgb]{0.0, 0.53, 0.74}{0.860} & \textcolor[rgb]{0.0, 0.53, 0.74}{0.071} & \textcolor[rgb]{0.0, 0.53, 0.74}{0.256} & \textcolor[rgb]{0.0, 0.53, 0.74}{0.480} & \textcolor[rgb]{0.0, 0.53, 0.74}{0.664} & \textcolor[rgb]{0.0, 0.53, 0.74}{0.601} & \textcolor[rgb]{0.0, 0.53, 0.74}{0.715}  & \textcolor[rgb]{0.0, 0.53, 0.74}{0.431} & \textcolor[rgb]{0.0, 0.53, 0.74}{0.702}  & \textcolor[rgb]{0.0, 0.53, 0.74}{0.135} & \textcolor[rgb]{0.0, 0.53, 0.74}{0.403} & \textcolor[rgb]{0.0, 0.53, 0.74}{0.339} & \textcolor[rgb]{0.0, 0.53, 0.74}{0.650} & \textcolor[rgb]{0.0, 0.53, 0.74}{0.781} & \textcolor[rgb]{0.0, 0.53, 0.74}{0.855} &\textcolor[rgb]{0.0, 0.53, 0.74}{0.730} &\textcolor[rgb]{0.0, 0.53, 0.74}{0.850}  &\textcolor[rgb]{0.0, 0.53, 0.74}{\textbf{0.854}} &\textcolor[rgb]{0.0, 0.53, 0.74}{0.913}\\
          \midrule
           \multirow{4}{*}{DomainNet} & ResNet18 &0.568 &0.692 &0.670 &0.736 &0.423 &0.609 &0.326 &0.668 &0.429 &0.597 &0.704 &0.903 &0.202 &0.516 &0.219 &0.443 &0.358 &0.445 &0.897 &0.910 &0.758 &0.789 &\textbf{0.902} &\textbf{0.937}\\
          & ResNet50 &0.588 &0.703 &0.570 &0.706 &0.344 &0.573 &0.455 &0.697 &0.245 &0.404 &0.746 &0.872 &0.002 &0.041 &0.220 &0.430 &0.379 &0.527 &0.903 &0.927  &0.809 &0.879 &\textbf{0.910} &\textbf{0.950}\\
          & WRN-50-2 &0.609 &0.712 &0.774 &0.874 &0.711 &0.845 &0.437 &0.698 &0.846 &0.918 &0.585 &0.831 &0.003 &0.034 &0.363 &0.466 &0.520 &0.713 &0.885 &0.935 &0.850 &0.911  &\textbf{0.893} &\textbf{0.978}\\
          \cline{2-26}
          & \textcolor[rgb]{0.0, 0.53, 0.74}{Average} & \textcolor[rgb]{0.0, 0.53, 0.74}{0.588} & \textcolor[rgb]{0.0, 0.53, 0.74}{0.702} & \textcolor[rgb]{0.0, 0.53, 0.74}{0.671} & \textcolor[rgb]{0.0, 0.53, 0.74}{0.722}  & \textcolor[rgb]{0.0, 0.53, 0.74}{0.493}  & \textcolor[rgb]{0.0, 0.53, 0.74}{0.676}  & \textcolor[rgb]{0.0, 0.53, 0.74}{0.406} & \textcolor[rgb]{0.0, 0.53, 0.74}{0.688} & \textcolor[rgb]{0.0, 0.53, 0.74}{0.507} & \textcolor[rgb]{0.0, 0.53, 0.74}{0.639} & \textcolor[rgb]{0.0, 0.53, 0.74}{0.678} & \textcolor[rgb]{0.0, 0.53, 0.74}{0.869} & \textcolor[rgb]{0.0, 0.53, 0.74}{0.069} & \textcolor[rgb]{0.0, 0.53, 0.74}{0.197} & \textcolor[rgb]{0.0, 0.53, 0.74}{0.234} & \textcolor[rgb]{0.0, 0.53, 0.74}{0.446} & \textcolor[rgb]{0.0, 0.53, 0.74}{0.419} & \textcolor[rgb]{0.0, 0.53, 0.74}{0.562} & \textcolor[rgb]{0.0, 0.53, 0.74}{0.894} & \textcolor[rgb]{0.0, 0.53, 0.74}{0.919} & \textcolor[rgb]{0.0, 0.53, 0.74}{0.805} & \textcolor[rgb]{0.0, 0.53, 0.74}{0.895} & \textcolor[rgb]{0.0, 0.53, 0.74}{\textbf{0.899}}  & \textcolor[rgb]{0.0, 0.53, 0.74}{\textbf{0.949}}\\
          \midrule
          \multirow{4}{*}{RR1-WILDS} & ResNet18 &0.821 &1.000 &0.951 &1.000 &0.836 &1.000 &0.929 &1.000 &0.342 &0.500 &0.936 &1.000 &0.843 &1.000 &0.859 &1.000 &0.927 &1.000 &0.969 &1.000 &0.885 &1.000 &\textbf{0.983} &\textbf{1.000}\\
          & ResNet50 &0.740 &1.000 &0.918 &1.000 &0.819 &1.000 &0.938 &1.000 &\textbf{0.986} &\textbf{1.000} &0.935 &1.000 &0.737 &1.000 &0.867 &1.000 &0.938 &1.000 &0.960 &1.000 &0.906 &1.000 &0.978 &\textbf{1.000}\\
          & WRN-50-2 &0.031 &0.500 &0.941 &1.000 &0.846 &1.000 &0.946 &1.000 &\textbf{0.988} &\textbf{1.000} &0.922 &1.000 &0.824 &1.000 &0.878 &1.000 &0.954 &1.000 &0.934 &1.000 &0.840 &1.000 &0.969 &\textbf{1.000}\\
          \cline{2-26}
          & \textcolor[rgb]{0.0, 0.53, 0.74}{Average} & \textcolor[rgb]{0.0, 0.53, 0.74}{0.530} & \textcolor[rgb]{0.0, 0.53, 0.74}{0.833} & \textcolor[rgb]{0.0, 0.53, 0.74}{0.937} & \textcolor[rgb]{0.0, 0.53, 0.74}{1.000} & \textcolor[rgb]{0.0, 0.53, 0.74}{0.833} & \textcolor[rgb]{0.0, 0.53, 0.74}{1.000} & \textcolor[rgb]{0.0, 0.53, 0.74}{0.938} & \textcolor[rgb]{0.0, 0.53, 0.74}{1.000} & \textcolor[rgb]{0.0, 0.53, 0.74}{0.779} & \textcolor[rgb]{0.0, 0.53, 0.74}{0.833}  & \textcolor[rgb]{0.0, 0.53, 0.74}{0.931} & \textcolor[rgb]{0.0, 0.53, 0.74}{1.000} & \textcolor[rgb]{0.0, 0.53, 0.74}{0.801} & \textcolor[rgb]{0.0, 0.53, 0.74}{0.833} & \textcolor[rgb]{0.0, 0.53, 0.74}{0.868} & \textcolor[rgb]{0.0, 0.53, 0.74}{1.000}  & \textcolor[rgb]{0.0, 0.53, 0.74}{0.940} & \textcolor[rgb]{0.0, 0.53, 0.74}{1.000} & \textcolor[rgb]{0.0, 0.53, 0.74}{0.953} & \textcolor[rgb]{0.0, 0.53, 0.74}{1.000} & \textcolor[rgb]{0.0, 0.53, 0.74}{0.877} & \textcolor[rgb]{0.0, 0.53, 0.74}{1.000} &\textcolor[rgb]{0.0, 0.53, 0.74}{\textbf{0.977}} &\textcolor[rgb]{0.0, 0.53, 0.74}{\textbf{1.000}}\\
         \bottomrule
    \end{tabular}}}
    \label{tab:main 2}
\end{table*}
\paragraph{Improved robustness.} Figure~\ref{fig:boxplot} presents a box plot showing the estimation robustness across different distribution shifts on all datasets except ImageNet, using ResNet18. Results for ImageNet are excluded due to the lack of \textit{Rotation} and \textit{AgreeScore} data for this dataset, as these two methods require retraining the networks. We observe that the estimation performance of \method{} is more stable than other baselines across three types of distribution shifts. Additionally, \method{} achieves the highest median estimation performance.

\subsection{Additional Experiments}
In this section, we discuss the results obtained with additional architectures, the ablation studies we conducted to validate our implementation choices, and the generalization capabilities of \method{}. 
\paragraph{Beyond ResNets.} To demonstrate the efficiency and versatility of \method{}, we conduct experiments on recent models such as Vision Transformers~\citep[ViT]{dosovitskiy2020image} and ConvNeXt~\citep{liu2022convnet}. We compare \method{} to its best competitors on $6$ datasets for $3$ distribution shifts in Figure~\ref{fig:all_arch}. The full results are gathered in Table~\ref{tab:arch} of Appendix~\ref{app:more_arch}. We note that \textit{ConfScore} is particularly strong with ConvNexts while \method{} works the best with ResNets and ViT. Again, we observe that \method{} is the best method overall. 

\paragraph{Ablation study.} To motivate our choices of implementation, we provide in Appendix~\ref{app:additional_exp} ablation studies on the $L_p$ norm and the Taylor order $n$ as well as a sensitive analysis on the calibration threshold $\eta$.

\begin{figure}[!t]
    \includegraphics[width=\linewidth]{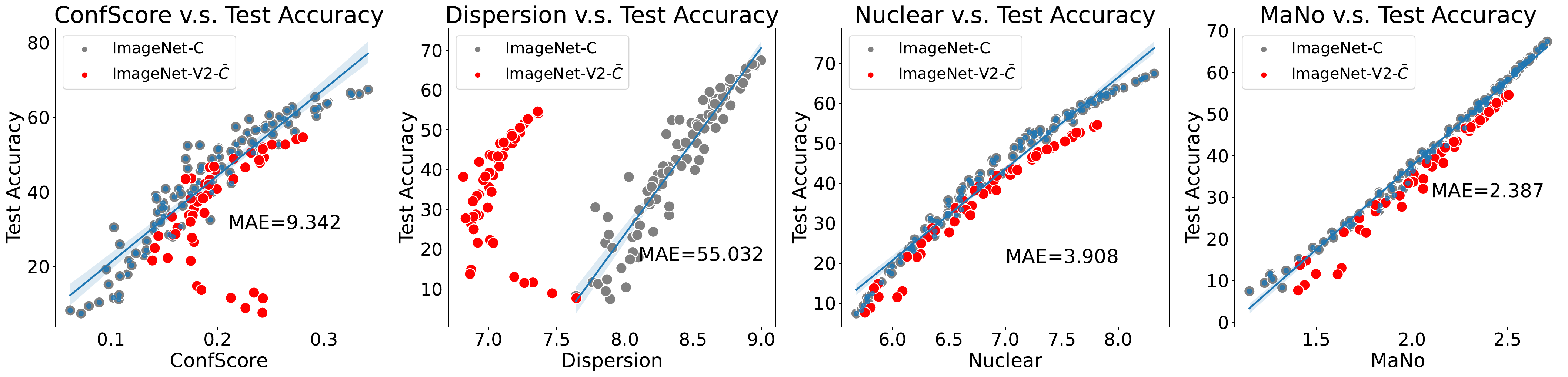}
    \vspace{-10pt}
    \caption{Comparison of generalization capability across four methods. Each subplot displays a linear regression model fitted on ImageNet-C, which is used to predict the accuracy on ImageNet-V2-$\Bar{C}$. The mean absolute error (MAE) is reported. All experiments are conducted using ResNet18.
    }
    \label{fig:imagenet_bar}
    \vspace{-8pt}
\end{figure}

\paragraph{Generalization capabilities of \method{}.}
To verify the generalization capabilities of \method{}, we utilize designed scores calculated from ImageNet-C and their corresponding accuracy to fit a linear regression model. This model is then used to predict the test accuracy on ImageNet-V2-$\Bar{C}$, which is generated using the $10$ new corruptions provided by~\citep{mintun2021interaction} on ImageNet-V2 \citep{recht2019imagenet}. These new corruptions are perceptually dissimilar from those in ImageNet-C, including warps, blurs, color distortions, noise additions, and obscuring effects. Figure~\ref{fig:imagenet_bar} shows that ConfScore and Dispersion give two distinct trends, while Nuclear exhibits some deviations for ImageNet-V2-$\Bar{C}$. In comparison, our \method{} exhibits a consistent prediction pattern for both ImageNet-C and ImageNet-V2-$\Bar{C}$, aligning well with the linear regression model trained on ImageNet-C. Additionally, experimental results on ImageNet-C and ImageNet-$\Bar{C}$, generated from the validation set of ImageNet, are provided in Appendix~\ref{app:generalization_capability}, further demonstrating the superiority of \method{}. 

\subsection{Discussion}
\label{sec:discussion}
In this section, we discuss how \method{} can be applied in practice, the benefit of combining \normalization{} with other estimation baselines, and the limitations of our approach.
\paragraph{Real-world applications.} In Appendix~\ref{app:mano_practice}, we discuss how \method{} can be used in real-world applications. In particular, our additional results with the Mean Absolute Error (MAE) metric confirm the superiority of \method{}.

\paragraph{Can \normalization{} enhance other logit-based methods?} 
\begin{wraptable}[10]{r}{0.52\textwidth}
\vspace{-12pt}
\tiny
\caption{Impact of \normalization{} on other logit-based methods. \normalization{} significantly boosts the performance of Nuclear. The metric used is $R^2$.}
\renewcommand\arraystretch{1}
\resizebox{\linewidth}{!}{
\setlength{\tabcolsep}{1mm}{
\begin{tabular}{ccccccc}
\toprule
\multirow{2}{*}{Dataset} &\multicolumn{2}{c}{ConfScore} &\multicolumn{2}{c}{\method{}} &\multicolumn{2}{c}{Nuclear} \\
\cline{2-7}
&w/o &w/  &w/o &w/ &w/o &w/\\
\midrule
PACS &\textbf{0.594} &0.574 &0.541 &\textbf{0.827} &0.609 &\textbf{0.851} \\
\midrule
Office-Home &0.795 &\textbf{0.829} &\textbf{0.929} &0.926 &0.692 &\textbf{0.826}\\
\bottomrule
\end{tabular}}}
\label{tab:softtrun}
\end{wraptable}
To study this, we conducted an ablation study by equipping \normalization{} with \textit{Nuclear} \citep{deng2023confidence}, \textit{ConfScore} \citep{hendrycks2016baseline}, and our \method{}. In Table~\ref{tab:softtrun}, we observe that \normalization{} significantly enhances the estimation performance $R^2$ of Nuclear. For example, Nuclear is improved from $0.692$ to $0.826$ on poorly-calibrated Office-Home.

\paragraph{Limitations.} Despite its soundness and strong empirical performance, we acknowledge that our method has potential areas for improvement. One of these is the dependence on $\eta$  of the selection criterion in Eq.~\eqref{eq:criterion}. We elaborate on this discussion in Appendix~\ref{app:limitations}. In future work, we will explore a smoother way to automatically select the optimal normalization function without requiring hyperparameters. Additionally, if multiple validation sets are provided, as in \citep{deng2021labels, deng2021does}, we could select $\eta$ based on those sets.

\section{Conclusion}
In this paper, we introduce \method{}, a simple yet effective training-free method to estimate test accuracy in an unsupervised manner using the Matrix Norm of neural network predictions on test data. Our approach is inspired by the LDS assumption that optimal decision boundaries should lie in low-density regions. To mitigate the negative impact of different distribution shifts on estimation performance, we first demonstrate the failure of \texttt{softmax} normalization under poor calibration, due to the accumulation of overconfident errors. We then propose a normalization strategy based on Taylor polynomial approximation, balancing logits information and error accumulation. Extensive experiments show that \method{} consistently outperforms previous methods across various distribution shifts. This work highlights that logits imply the feature-to-boundary distance and considers the impact of calibration on estimation performance. We hope our insights inspire future research to explore the relationship between model outputs and generalization.

\section*{Acknowledgements}
Ambroise Odonnat would like to thank Alexandre Ram\'e and Youssef Attia El Hili for the fruitful discussions that led to this work. The authors thank the anonymous reviewers and meta-reviewers for their time and constructive feedback. This work was enabled thanks to open-source software such as Python~\citep{van1995python}, PyTorch~\citep{pytorch} and Matplotlib~\citep{hunter2007matplotlib}. This research is supported by the National Research Foundation Singapore and DSO National Laboratories under the AI Singapore Programme (AISGAward No: AISG2-GC-2023-009).

\bibliography{references}
\bibliographystyle{apalike}

\newpage
\appendix
\textbf{\LARGE Appendix}
\paragraph{Roadmap.} We provide the pseudo-code of \method{} in Appendix~\ref{app:pseudo_code}. We discuss related work in Appendix~\ref{app:related_work} and provide some background on Tsallis entropies in Appendix~\ref{app:tsallis_entropy}. Appendix~\ref{app:proofs} contains detailed proofs of our theoretical results. Additional discussion and theoretical insights into Section~\ref{sec:normalization} are given in Appendix~\ref{app:discuss_softrun}. In Appendix~\ref{app:more_arch}, we conduct experiments with ViT and ConvNext architectures and provide a thorough ablation study and sensitivity analysis in Appendix~\ref{app:additional_exp}. Finally, we explain how \method{} can be used in practice in Appendix~\ref{app:mano_practice}. We display the corresponding table of contents below.
\addtocontents{toc}{\protect\setcounter{tocdepth}{2}}

\renewcommand*\contentsname{\Large Table of Contents}

\tableofcontents
\clearpage

\section{Pseudo-Code of \method{}}
\label{app:pseudo_code}
Algorithm~\ref{alg:our_method} summarizes \method{} introduced in Section~\ref{sec:method}, which is a lightweight, training-free method for unsupervised accuracy estimation using the neural network's outputs. We open-sourced the code of \method{} at \url{https://github.com/Renchunzi-Xie/MaNo}.

\begin{algorithm}[!h] 
\caption{Our proposed algorithm, \method{}, for unsupervised accuracy estimation.}
\label{alg:our_method}
    \textbf{Input:} Model $f$ pre-trained on $\mcal{D}_\mrm{train}$, test dataset $\mcal{D}_\mrm{test}=\{\mbf{x}_i\}_{i=1}^N$. \\
    \textbf{Parameters:} Hyperparameter $p > 1$. \\
    \textbf{Initialization:} Empty prediction matrix $\mbf{Q} \in \RR^{N \times K}$. \\
    \underline{\textit{Criterion}}: compute criterion $\threshold{}$ and select $\sigma$ following Eq.~\eqref{eq:normalization} and Eq.~\eqref{eq:criterion}.\\
    \For{$i \in \llbracket 1, N \rrbracket$}{
    \underline{\textit{Inference}}: recover logits $\mbf{q}_i = f(\mbf{x}_i) \in \RR^K$. \\
    \underline{\textit{Normalization}}: obtain normalized logits $\sigma(\mbf{q}_i) \in \Delta_K$. \\ 
    \underline{\textit{Update}}: fill the prediction matrix $\mbf{Q}_i \gets \sigma(\mbf{q}_i)$ following Eq.~\eqref{eq:row_matrix}.
    }
    \textbf{Output} Estimation score $\score{f, \mcal{D}_\mrm{test}}=\frac{1}{\sqrt[\leftroot{-2}\uproot{2}p]{NK}}\lVert\mbf{Q} \rVert_p$ following Eq.~\eqref{eq:estimation_score}.
\end{algorithm}

\section{Extended Related Work}
\label{app:related_work}
\paragraph{Unsupervised accuracy estimation.} This task aims to estimate model generalization performance on unlabeled test sets. To achieve this, several directions have been proposed.
\underline{\textit{(1) Utilizing model outputs:}} One popular research direction is to use the model outputs on distribution-shifted data to construct a linear relationship with the test accuracy~\citep{hendrycks2016baseline, garg2022leveraging, deng2023confidence, guillory2021predicting,xie2024characterising}. Some of these approaches are limited by requiring access to the training set \citep{chuang2020estimating,chen2021detecting}. The most recent work \citep{deng2023confidence} uses the nuclear norm of the softmax probability matrix as a training-free accuracy estimator. However, it significantly suffers from the overconfidence issues~\citep{wei2022mitigating}, leading to fluctuating estimation performance across natural distribution shifts. Our work focuses on addressing this issue by balancing logit-information completeness and overconfidence-information accumulation.
\underline{\textit{(2) Considering distribution discrepancy:}} another direction examines the negative relation between test accuracy and the distribution discrepancy between the training and test datasets~\citep{deng2021labels, lu2023characterizing, yu2022predicting,tu2023bag}. However, commonly-used distribution distances do not guarantee stable accuracy estimation under different distribution shifts~\citep{guillory2021predicting, xie2023importance}, and some of these methods are time-consuming on large-scale datasets due to the requirement of training data~\citep{deng2021labels}.
\underline{\textit{(3) Constructing unsupervised losses:}} methods such as data augmentation and multiple-classifier agreement have also been introduced~\citep{jiang2021assessing, madani2004co, platanios2016estimating, platanios2017estimating}. However, they usually require special model architectures, undermining their practical applicability.

\paragraph{Distance to decision boundaries.} The idea of treating distance to the decision boundary as an indicator of confidence originates from classical support vector machines \citep{Vapnik:1998}. Decision boundaries of deep neural networks have been studied in various contexts. For example, some works explore the geometric properties of deep neural networks either in the input space \citep{fawzi2018empirical, poole2016exponential, montufar2014number, karimi2019characterizing} or in the weight space \citep{choromanska2015loss, dauphin2014identifying, chaudhari2019entropy, dinh2017sharp, freeman2016topology}. Some works apply the properties of decision boundaries to address practical questions, such as adversarial defense \citep{he2018decision, heo2019knowledge, croce2020minimally}, OOD detection \citep{lee2020multi}, and domain generalization \citep{yousefzadeh2021deep, li2022new}. As we discuss in Section \ref{sec:explain}, those approaches that use the distance to the decision boundary as an \textit{unsupervised} indicator of confidence, rely on the low-density separation assumption \citep{chapelle05lds}, which states that the classifier must mistake mostly in the low margin zone \citep{feofanov2019transductive,feofanov2024multi}. In our work, under this assumption, similar to \citet{li2018decision}, we use the distance between the learned intermediate feature to each decision boundary in the last hidden space.

\paragraph{Normalization in deep learning.} Normalization is a crucial technique extensively utilized across various fields in deep learning, including domain generalization \citep{wang2021tent, fan2021adversarially,seo2020learning}, metric learning \citep{sohn2016improved, wu2018unsupervised, oord2018representation}, face recognition \citep{ranjan2017l2, liu2017sphereface, wang2017normface, deng2019arcface, zhang2019adacos} and self-supervised learning \citep{chen2020simple}. For example, TENT \citep{wang2021tent} normalizes features of test data using the mean value and standard deviation estimated from the target data. $L_2$-constrained softmax \citep{ranjan2017l2} introduces $L_2$ normalization on features. These normalization techniques are primarily employed to adapt new samples to familiar domains, calculate similarity, and speed up convergence. However, our proposed normalization focuses on reducing the negative implications of poorly calibrated scenarios.

\section{Background on Tsallis Entropies}
\label{app:tsallis_entropy}
The definition of Tsallis $\alpha$-entropies~\citep{tsallis1988entropy} is given below.
\begin{boxdef}[Tsallis $\alpha$-entropies]
    Let $\mathbf{p} \in \Delta_K$ be a probability distribution. Let $\alpha > 1$ and $k\geq 0$. The Tsallis $\alpha$-entropy is defined as:
    \begin{equation*}
        \mathbf{H}_{\alpha}^\mathrm{T}(\mathbf{p}) = k(\alpha-1)^{-1}(1 - \lVert\mbf{p}\rVert_{\alpha}^{\alpha}).
    \end{equation*}
\end{boxdef}

\begin{wrapfigure}[12]{r}{0.3\textwidth}
    \centering
    \includegraphics[width=\linewidth]{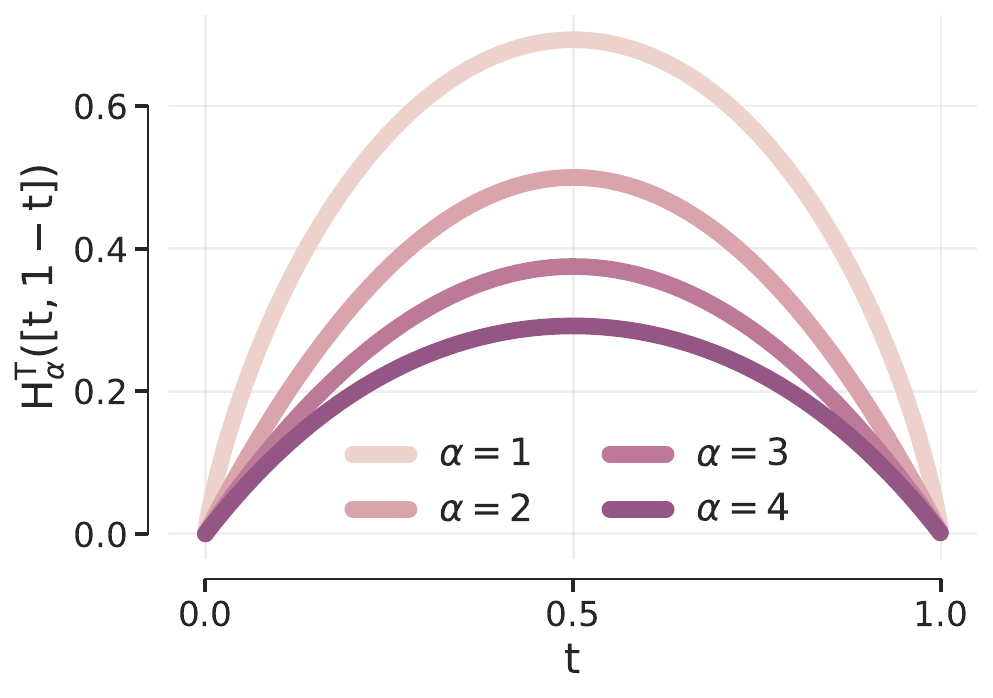}
    \caption{Tsallis $\alpha$-entropies of $[t, 1-t]$ for $t \in [0,1]$.}
    \label{fig:tsallis_entropy}
\end{wrapfigure}

It is common to take $k=1$ or $k=\frac{1}{\alpha}$ following~\citet{blondel19Fyclassifiers}. The Tsallis $\alpha$-entropy generalizes the Boltzmann-Gibbs theory of statistic mechanics to nonextensive systems. It has been used as a measure of disorder and uncertainty in many applications~\citep{murray2004entropy, renato_2014_orbit, sneddon_2007_tsallis, teimoori2024inflation}, including in Machine Learning~\citep{blondel19Fyclassifiers, blondel21FYlearning, muzellec17tsallis}. Moreover, they generalize two widely-known measures of uncertainty. Indeed, the limit case $\alpha \to 1$ leads to the Shannon entropy $\mathbf{H}_S$~\citep[see][Appendix A.1]{peters2019sparse}, \textit{i.e.},
$\lim_{\alpha \to 1}\mathbf{H}_{\alpha}^\mathrm{T}(\mathbf{p}) = \mathbf{H}_S(\mathbf{p}) = -\sum_{j =1}^K p_j \ln(p_j)$, while taking $\alpha=2$ leads to the Gini index $\mbf{G}$, a popular impurity measure fo decision trees~\citep{gini1912variabilità}, \textit{i.e.}, $\mathbf{H}_2^\mathrm{T}(\mathbf{p}) = \frac{1}{2}(1 - \lVert p\rVert_2^2) = \mathbf{G}(\mathbf{p})$. Tsallis entropies measure the uncertainty: the higher the entropy the greater the uncertainty. From a probabilistic perspective, the entropy will take high values for \emph{uncertain} probability distributions, \textit{i.e.}, close to the uniform distribution. We visualize the evolution of the Tsallis entropy for varying parameters $\alpha$ in Figure~\ref{fig:tsallis_entropy}, where the case $\alpha=1$ corresponds to the Shannon entropy. 

\section{Proofs}
\label{app:proofs}
In this section, we detail the proofs of our theoretical results.

\paragraph{Notations.} Scalar values are denoted by regular letters (e.g., parameter $\lambda$), vectors are represented in bold lowercase letters (e.g., vector $\mbf{x}$) and matrices are represented by bold capital letters (e.g., matrix $\mbf{A}$). The $i$-th row of the matrix $\mbf{A}$ is denoted by $\mbf{A}_{i}$, its $j$-th column is denoted by $\mbf{A}_{\cdot, j}$ and its elements are denoted by $\mbf{A}_{ij}$. The trace of a matrix $\mbf{A}$ is denoted by $\tr{\mbf{A}}$ and its transpose by $\mbf{A}^\top$. The $L_p$ norm of a vector $\mbf{x}$ is denoted by $\lVert \mbf{x}\rVert_p$, and by abuse of notation we denote it by $\lVert\mbf{A}\rVert_p$ for a matrix $\mbf{A}$ with $\lVert\mbf{A}\rVert_p^p = \sum_{i}\lVert \mbf{A}_i\rVert_p^p = \sum_{ij}\lvert \mbf{A}_{ij}\rvert^p$. Let $\Delta_K \coloneqq \{\mbf{p} \in [0,1]^K | \sum_{i=1}^K\mbf{p}_i = 1\}$ be the $K$-dimensional probability simplex.

\subsection{Impact of Prediction Errors}
\label{app:impact-pred-err}
Let $\mbf{x} \in \mcal{D}_\mrm{test}$ be a test sample with ground-truth label $y \in \{1, \dots, K\}$. In multi-class classification, the \texttt{softmax} operator is used to approximate the posterior probability $p(y | \mbf{x})$~\citep[see][chap.4, p.198]{bishop2006pattern}. Reusing the notations of Section~\ref{sec:normalization}, because of the distribution shifts between source and target, logits are subject to a prediction bias $\bm{\varepsilon} = (\varepsilon_k)_k$ and write $f(\mbf{x}) = \mbf{q}^* + \bm{\varepsilon}$ where $\mbf{q}^*$ are ground-truth logits. In this section, we study the impact of such bias on the approximation of the posterior $p(y | \mbf{x})$.

\paragraph{Impact on the posterior approximation.} Proposition~\ref{prop:kl_bounds} shows the impact of the prediction bias on the KL divergence between the true class posterior probabilities, assumed modeled as $\mbf{p} = \softmax{\mbf{q}^*}\in \Delta_K$, and the estimated ones $\mbf{s} = \softmax{f(\mbf{x})}\in \Delta_K$. In particular, it states that 
\begin{equation}
\label{eq:kl_bounds}
        0 \leq \mrm{KL}\mleft(\mbf{p} || \mbf{s}\mright) \leq \bm{\varepsilon}_{+}^T\mbf{p},
    \end{equation}
where $\bm{\varepsilon}_{+} = (\max_{l}\{\varepsilon_l\} - \varepsilon_k)_k \in \RR^K_+$. The proof is given below.
\begin{proof}
We denote $\mbf{q} = f(\mbf{x}) \in \RR^K$ the neural network's outputs on a given test sample $\mbf{x}$.
    We first remark that
    \begin{equation}
    \label{eq:kl_equality}
    \begin{split}
        \mrm{KL}(\mbf{p} || \mbf{s}) &=  \sum_k \mbf{p}_k \ln\mleft( \frac{\mbf{p}_k}{\mbf{s}_k}\mright)\\ 
        & = \sum_k \mbf{p}_k \ln \mleft(
\frac{\exp(\mbf{q}^*_{k})}{\sum_{j=1}^K \exp(\mbf{q}^*_{j})} \cdot \frac{\sum_{j=1}^K \exp(\mbf{q}^*_{j} + \varepsilon_j)}{\exp(\mbf{q}^*_{k} + \varepsilon_k)}
 \mright) \\
 & = \sum_k \mbf{p}_k \ln \mleft( \exp(-\varepsilon_k) \cdot \frac{\sum_{j=1}^K \exp(\mbf{q}^*_{j} + \varepsilon_j)}{\sum_{j=1}^K \exp(\mbf{q}^*_{j})}
 \mright). \\
 \end{split}
    \end{equation}
To obtain the upper-bound, we notice that 
\begin{equation*}
    \exp(\mbf{q}^*_{j} + \varepsilon_j) = \exp(\mbf{q}^*_{j}) \exp(\varepsilon_j) \leq \exp(\mbf{q}^*_{j}) \cdot \max_l \{\exp(\varepsilon_l)\}.
\end{equation*}
This leads to 
\begin{equation}
    \frac{\sum_{j=1}^K \exp(\mbf{q}^*_{j} + \varepsilon_j)}{\sum_{j=1}^K \exp(\mbf{q}^*_{j})} \leq \frac{\max_l \{\exp(\varepsilon_l)\} \sum_{j=1}^K \exp(\mbf{q}^*_{j})}{\sum_{j=1}^K \exp(\mbf{q}^*_{j})} = \max_l \{\exp(\varepsilon_l)\}.
\end{equation}
Using the fact that all the terms are positive and that $\ln$ and $\exp$ are increasing functions, we obtain from Eq.~\eqref{eq:kl_equality} that
\begin{equation}
\label{eq:ub}
\begin{split}
    \sum_k \mbf{p}_k \ln \mleft( \exp(-\varepsilon_k) \cdot \frac{\sum_{j=1}^K \exp(\mbf{q}^*_{j} + \varepsilon_j)}{\sum_{j=1}^K \exp(\mbf{q}^*_{j})}
 \mright) & \leq \sum_k \mbf{p}_k \ln ( \exp(-\varepsilon_k) \cdot \max_l \{\exp(\varepsilon_l)\}) \\
 & \leq \sum_k \mbf{p}_k [\ln ( \exp( \max_l \{\varepsilon_l\})) -\varepsilon_k] \\
 & \leq \sum_k \mbf{p}_k [\max_l \{\varepsilon_l\} - \varepsilon_k ]\\
 & = \bm{\varepsilon}_{+}^T\mbf{p}.
 \end{split}
\end{equation}
Combining Eq.~\eqref{eq:kl_equality} and Eq.~\eqref{eq:ub} gives the upper bound.
\end{proof}
The quantities $\bm{\varepsilon}_{+}$ is a linear transformation of the prediction bias $\bm{\varepsilon} \in\RR^K$ and has nonnegative entries, which means each class is overestimated, representing an \emph{overconfident} model. Proposition~\ref{prop:kl_bounds} shows that the discrepancy between the error approximation of the posterior probabilities is controlled by the alignment between the posterior and this extreme prediction bias. In addition, by a simple application of Cauchy-Schwartz in Eq.~\eqref{eq:kl_bounds} and using the fact that $\lvert \mbf{p} \rVert = \sum_{k=1} \mbf{p}_k^2 \leq \sum_{k=1} \mbf{p}_k = 1$, we have $\mrm{KL}\mleft(\mbf{p} || \mbf{s}\mright) \leq \lVert\bm{\varepsilon}_{+}\rVert_2$. In particular, in the perfect situation where $\bm{\varepsilon} = \bm{0}$, $\bm{\varepsilon}_{+}$ is equal to $0$ and the \texttt{softmax} probabilities perfectly approximate the posterior. In summary, Proposition~\ref{prop:kl_bounds} indicates that not only the norm of the prediction bias but also its alignment to the posterior is responsible for the approximation error of the posterior. In our setting, it means that logits-methods need a low prediction bias on classes on which the model is confident such that \texttt{softmax} probabilities can be reliably used to estimate accuracy. This follows our analysis and empirical verification from Section~\ref{sec:normalization}.

\paragraph{A real-world example.} Although we usually tend to think that a high prediction bias shifts the predicted posterior towards the uniform distribution, in the general case, other situations may happen that hinder the quality of the accuracy estimation. For example, one may think of a letter recognition task with a neural network pre-trained on the Latin alphabet and tested on the Cyrillic one. In this case, some prediction probabilities will be adversarial as the neural network will not be aware of the semantic differences between the Latin \say{B} and the Cyrillic \say{B}, therefore predicting a wrong class with high probability.

\subsection{Distance to the Hyperplane}
\label{app:lem_distance_boundary}
\begin{boxlem}[\citet{stackexchange}]
\label{lem:distance_to_boundary}
    Let $\bm{\omega} \in \RR^n$ be non zero and $b \in \RR$. The distance between any point $\mbf{z} \in \RR^n$ and the hyperplane $\{\mbf{x} | \bm{\omega}^\top\mbf{x} + b = 0\}$ writes $\mrm{d}(\bm{\omega}, \mbf{z}) = \lvert \bm{\omega}^\top\mbf{x} + b\rvert/\norm{\bm{\omega}}$.
\end{boxlem}
\begin{proof}
    The proof follows the geometric intuition from~\citet{stackexchange}. We recall it here for the sake of self-consistency. The distance between $\mbf{z}$ and the hyperplane $\mcal{H} = \{\mbf{x} \in \RR^n| \bm{\omega}^\top\mbf{x} + b = 0\}$ is equal to the distance between $\mbf{z}$ and its orthogonal projection on $\mcal{H}$. We consider the line $L = \{\mbf{z} + t\bm{\omega} | t \in \RR\}$ that is orthogonal to $\mcal{H}$ and passes through $\mbf{z}$.  The desired orthogonal projection is simply the point $\mbf{z} + t^*\bm{\omega}$ such that $L$ and $\mcal{H}$ intersects, \textit{i.e.}, such that 
    \begin{align*}
        \bm{\omega}^\top\mleft(\mbf{z} + t^*\bm{\omega}\mright) + b = 0 &\Leftrightarrow \bm{\omega}^\top\mbf{z} + b = -t^*\norm{\bm{\omega}}^2 \\
        &\Leftrightarrow t^* = -\frac{\bm{\omega}^\top\mbf{z} + b}{\norm{\bm{\omega}}^2} \tag{$\norm{\bm{\omega}} \neq 0$}.
    \end{align*} 
    It follows that the distance between $\mbf{z}$ and $\mcal{H}$ writes
    \begin{equation*}
        \mrm{d}(\bm{\omega}, \mbf{z}) = \norm{\mbf{z} - \mbf{z} + t^*\bm{\omega}} = \norm{-\frac{\bm{\omega}^\top\mbf{z} + b}{\norm{\bm{\omega}}^2} \times \bm{\omega}} = \frac{\lvert\bm{\omega}^\top\mbf{z} + b\rvert}{\norm{\bm{\omega}}}.
    \end{equation*}
\end{proof}

\subsection{Proof of Theorem~\ref{thm:connection_uncertainty}}
\label{app:thm_connection_uncertainty}
\begin{proof}
    Reusing the notations introduced in Section~\ref{sec:method} and Algorithm~\ref{alg:our_method}, we have that
    \begin{align*}
        \score{f, \mcal{D}_\mrm{test}}^p = \frac{1}{NK}\lVert\mbf{Q}\rVert_p^p &= \frac{1}{NK}\sum_{i=1}^N \lVert \mbf{Q}_i\rVert_p^p \tag{Definition of $\lVert\mbf{Q}\rVert_p$}\\
        &= \frac{1}{NK}\sum_{i=1}^N \lVert \probmap{\mbf{q}_i} \rVert_p^p \tag{Definition of $\mbf{Q}_i$ in Algorithm~\ref{alg:our_method}}\\
        &= \frac{1}{NK}\sum_{i=1}^N 1 - (1 - \lVert\probmap{\mbf{q}_i}\rVert_p^p) \\
        &= \frac{1}{K} - \frac{p(p-1)}{NK}\sum_{i=1}^N \frac{1}{p(p-1)}(1 - \lVert\probmap{\mbf{q}_i}\rVert_p^p) \\
        &= \frac{1}{K} - \frac{p(p-1)}{NK}\sum_{i=1}^N \mathbf{H}_p^\mathrm{T}\mleft(\probmap{\mbf{q}_i}\mright) \tag{Definition of $\mathbf{H}_p^\mathrm{T}$} \\
        &= b - a \mleft(\frac{1}{N}\sum_{i=1}^N \mathbf{H}_p^\mathrm{T}\mleft(\probmap{\mbf{q}_i}\mright)\mright) ,\\
    \end{align*}
where $a = \frac{p(p-1)}{K} > 0$ and $b=\frac{1}{K}$. Rearranging the terms concludes the proof.
\end{proof}

\subsection{Properties of $\phi$}
\label{app:phi_proprieties}
The \texttt{softmax} can be decomposed as $\softmax{\mbf{q}} = \exp(\mbf{q})/\sum_{k=1}^K \exp(\mbf{q})_k = (\phi \circ \exp)(\mbf{q})$, where $\phi \colon \RR^K_+ \to \Delta_K$ writes $\phi(\mbf{u})= \mbf{u} / \sum_{k=1}^K\mbf{u}_k = \mbf{u} / \lVert\mbf{u}\rVert_1$. We extend the domain of $\phi$ to $\RR^K_+$ by setting $\phi(\mbf{0}) = \frac{1}{K}\mathbbm{1}_K$. The following proposition states the properties of $\phi$.
\begin{boxprop}[Properties of $\phi$] 
\label{prop:property_phi}
\text{}
\begin{enumerate}[wide, labelwidth=!, labelindent=0pt]
    \item \textbf{Generalized injectivity.} $\forall \mbf{u}, \mbf{v} \in \RR^K_+ \setminus\{\mbf{0}\}, \quad \phi(\mbf{u}) = \phi(\mbf{v}) \iff \exists \alpha \in \RR^*, \text{ s.t. } \mbf{u} = \alpha \mbf{v}$
    \item \textbf{Evaluation on constant inputs.} Let $\mbf{u} = \alpha \mathbbm{1}_K$ with $\alpha \geq 0$. Then, we have $\phi(\mbf{u}) = \frac{1}{K}\mathbbm{1}_K.$
\end{enumerate}
\end{boxprop}
\begin{proof}
We start by proving the first part of Proposition~\ref{prop:property_phi}. Let $\mbf{u}, \mbf{v} \in \RR^K_+ \setminus{\mbf{0}}$. We have
\begin{align*}
    \phi(\mbf{u}) = \phi(\mbf{v}) & \iff \frac{\mbf{u}}{\lVert \mbf{u} \rVert_1} = \frac{\mbf{v}}{\lVert \mbf{v} \rVert_1} \iff \mbf{u} = \underbrace{\frac{\lVert \mbf{u} \rVert_1}{\lVert \mbf{v}\rVert_1}}_{\alpha > 0} \times \mbf{v}.
\end{align*}
Then, we prove the second part of the proposition. Let $\alpha \geq 0$ and consider $\mbf{u}  = \alpha \mathbbm{1}_K \in \RR^K_+$. If $\alpha = 0$, then $\mbf{u} = \mbf{0}$ and by definition, $\phi(\mbf{u}) = \phi(\mbf{0}) = \frac{1}{K}\mathbbm{1}_K$. Assuming $\alpha > 0$, we have
\begin{equation*}
        \phi(\mbf{u}) = \frac{\mbf{u}}{\lVert \mbf{u}\rVert_1} = \frac{\mbf{u}}{\sum_{k=1}^K\mbf{u}_k} = \frac{\alpha}{\sum_{k=1}^K \alpha} \times \mathbbm{1}_K = \frac{1}{K} \mathbbm{1}_K.
\end{equation*}
\end{proof}
The first part of the proposition is dubbed \say{generalized injectivity} as the injectivity can be retrieved by fixing $\alpha=1$ in Proposition~\ref{prop:property_phi}. It ensures that $\phi$ only has \emph{equal} outputs if the inputs are \emph{similar}. To illustrate that, consider the logits $\mbf{q}, \bm{\delta} \in \RR^K$. From Proposition~\ref{prop:property_phi}, having $\softmax{\mbf{q}} = \softmax{\bm{\delta}}$ is equivalent to having $\exp(\mbf{q}) = \alpha \exp(\bm{\delta})$ for some $\alpha \neq 0$. By positivity of both sides, it implies $\alpha > 0$, and taking the logarithm leads to $\mbf{q} = \bm{\delta} + \ln{\alpha}$. It means that $\mbf{q}$ equals $\bm{\delta}$ up to a fixed constant. From a learning perspective, those logits will thus have the same predicted label and normalized logits. Proposition~\ref{prop:property_phi} shows that using $\phi$ preserves the information from the neural network. In addition, if the neural network's output is not informative,\textit{i.e.}, all entries are equal, then the link function gives equal probability to all classes.

\section{Theoretical Insights into Criterion $\threshold{}$}
\label{app:discuss_softrun}

In Section~\ref{sec:normalization}, we explained the main drawbacks of using the softmax normalization in the presence of the prediction bias proposing a new alternative normalization, \normalization{}, that we recall is
\begin{equation*}
    v(\mbf{q}_i) = \begin{cases}
1+\mbf{q}_i+\frac{\mbf{q}_i^2}{2}, &\text{if } \threshold{} \leq \eta \qquad \text{(Taylor)}\\
\exp(\mbf{q}_i), &\text{if } \threshold{} > \eta \qquad \text{(softmax)}
    \end{cases},
\end{equation*} 
where $\threshold{}$ is the selection criterion defined as
\begin{equation*}
\threshold{} = -\frac{1}{NK} \sum_{j=1}^N\sum_{k=1}^K \log(\frac{\exp(\mbf{q}_j)_k}{\sum_{j=1}^K\exp(\mbf{q}_i)_j)}).
\end{equation*}
In this section, we first would like to give more insights on the choice of $\threshold{}$ and the selection rule. Then, we motivate our choice of the hyperparameter $\eta$. Finally, we discuss the potential limitations of our approach.

\subsection{Choice of Criterion $\threshold{}$}

\begin{figure}[!t] 
    \centering
    \includegraphics[width=0.7\linewidth, clip=True, trim=1cm 0.5cm 0 0.5cm]{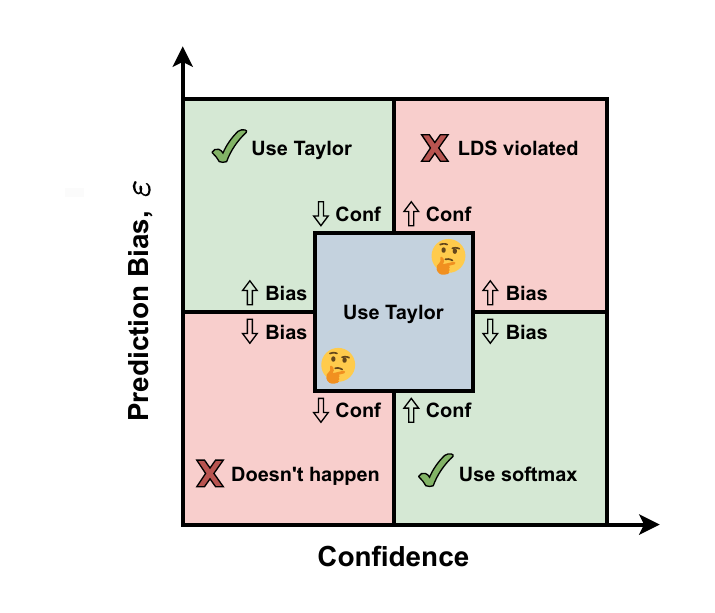}
    \caption{A schematic illustration of possible cases that may or may not happen and what normalization technique we use depending on the model's confidence and the value of the prediction bias. The two \textit{unadmissible} scenarios correspond to the red sub-squares.}
    \label{fig:pred_bias_vs_conf}
\end{figure}

\paragraph{Intuition.} We first provide some high-level intuition behind the selection criterion of Section~\ref{sec:normalization}. As we discussed before, the main idea of the logit-based approach is to rely on the model's confidence whose reliability depends on the prediction bias induced by possible distribution shift.  Depending on exact values of confidence and bias, we can roughly distinguish the five following cases illustrated in Figure \ref{fig:pred_bias_vs_conf} and described as follows:

\begin{enumerate}[leftmargin=*]
    \item \textit{High confidence, high bias.} The model is self-confident but practically makes a lot of mistakes. This corresponds to the case when assumptions are not met (Section \ref{sec:explain}), so logits are generally uninformative, and no normalization technique can really fix it. Thus, in practice, we have to make sure that the low-density separation (LDS) assumption holds, which is generally the case for well-calibrated models trained on diverse training sets. However, a user has to be careful when applying test-time adaptation methods \citep{wang2021tent,rusak2022if,chen2022contrastive} to an original pre-trained model, since these approaches perform unsupervised confidence maximization making LDS not guaranteed anymore.
    \item \textit{Low confidence, low bias.} The model tends to output confidence that is low but overall precise. This situation is unlikely to happen mainly for the following two reasons. Firstly, the classification problem is poorly posed as the considered case implies that the true posterior probabilities are close to uniform ones. Secondly, it is known that deep neural networks tend to be overconfident in their predictions \citep{wei2022mitigating}, which we also observed in our experiments. Thus, we do not consider this case and leave it as a subject of future work.
    \item \textit{High confidence, low bias.} The model tends to be self-confident being overall precise. This is a favorable case as logits are very reliable, so we can use softmax normalization without being afraid to be too optimistic.
    \item \textit{Low confidence, high bias.} The model is not confident in their predictions and it indeed makes a lot of mistakes due to the high prediction bias. This is also a favorable scenario as low confidence correlates with low performance. In this case, we want posterior probabilities to be close to uniform, and we use the Taylor normalization due to its smoother behavior. 
    \item \textit{Grey zone.} It corresponds to a mixed scenario when different examples may refer to different cases. As it is generally difficult to know what normalization technique would be the most relevant, we would opt for a more conservative solution in this situation. This is where the Taylor normalizer becomes useful as it does not exacerbate prediction bias to the same degree as the softmax (see Section \ref{sec:softmax-failure} for more details).
\end{enumerate}

\paragraph{Formulation of $\threshold{}$.} 
As we discussed in the main body of the paper, the criterion $\threshold{}$ is equal, up to a constant, to the average KL divergence between the uniform distribution and the predicted \texttt{softmax} probabilities~\citep{tian2021exploring}. This implies that the criterion reflects the model's confidence being high when predicted probabilities are far away from the uniform ones. As we rely on the LDS assumption, high values of $\threshold{}$ correspond to the $3^{\mrm{rd}}$ case (\textit{high confidence, low bias}), and the softmax normalization is selected. Conversely, when the $\threshold{}$ is lower than the threshold $\eta$, we apply Taylor, which corresponds to either the $4^{\mrm{th}}$ case (\textit{low confidence, high bias}) or the $5^{\mrm{th}}$ case (\textit{grey zone}). 
 
\paragraph{Connection between criterion and misclassification error.}
The next proposition provides further insights into our selection process. Let $\mbf{u} = \frac{1}{K}\mathbbm{1}_K\in \Delta_K$ be the uniform probability. The test dataset writes $\mcal{D}_\mrm{test} =\{\mbf{x}_i \}_{i=1}^N$ with corresponding logits $\mbf{q}_i = f(\mbf{x}_i)$ and ground-truth labels $\mcal{Y}_{\mrm{test}}=\{y_i\}_{i=1}^N$ (unavailable in practice). We denote the \texttt{softmax} probabilities by $\mbf{s}^i = \softmax{\mbf{q}_i} = \exp(\mbf{q}_i) / \sum_{k=1}^K \exp(\mbf{q}_i)_k \in \Delta_K$. We introduce the entropy of a probability vector as $\mrm{H}(\mbf{p}) = - \frac{1}{K}\sum_{k=1}^K \mbf{p}_k \ln(\mbf{p}_k)$. In particular, it is a measure of uncertainty and takes a high value when the model is uncertain, \textit{i.e.}, outputs probabilities close to the uniform. We establish in the following proposition the connection between the criterion $\threshold{}$, the miscalibration error, the model's confidence, and its entropy. 
\begin{boxprop}[$\threshold{}$, misclassification error, confidence and entropy] 
\label{prop:criterion_error_uncertainty}
We have
\begin{equation*}
\underbrace{\xi\mleft(\mcal{D}_\mrm{test},\mcal{Y}_{\mrm{test}}\mright)}_{\text{misclassification}} + \underbrace{\mcal{U}\mleft( \mcal{D}_\mrm{test}\mright)}_{\text{confidence}} + \underbrace{\mcal{H}\mleft( \mcal{D}_\mrm{test}\mright)}_{\text{entropy}} \leq \underbrace{\threshold{}}_{\text{criterion} } + \ln(e + \frac{1}{K}),
\end{equation*}
where $\xi\mleft(\mcal{D}_\mrm{test}, \mcal{Y}_\mrm{test}\mright) = \frac{1}{N}\sum_{i=1}^N \mleft(1 - \mbf{s}^i_{y_i}\mright)$, $\mcal{U}\mleft(\mcal{D}_\mrm{test}\mright) = \frac{1}{N}\sum_{i=1}^N \kld{\mbf{u}}{\mbf{s}^i}$, and $\mcal{H}\mleft(\mcal{D}_\mrm{test}\mright) = \frac{1}{N}\sum_{i=1}^N \mrm{H}(\mbf{s}^i)$.
\end{boxprop}
\begin{proof}
We first present the following lemma that introduces the change of measure inequality \citep{banerjee2006,seldin2010pac}.
\begin{boxlem}[Change of measure inequality~\citep{seldin2010pac}]
\label{lem:change-of-measure-lem}
Let $Z$ be a random variable on $\{1, \dots, K\}$ and $\bm{\mu}=(\mu_k
)_{k}\in \Delta_K$ and $\bm{\nu}=(\nu_k
)_{k} \in \Delta_K$ be two probability distributions. For any measurable function $\psi \colon \mcal{Z} \to \RR$, the following inequality holds:
\begin{equation*}
    \sum_{k=1}^K \mu_k \psi(k) \leq \kld{\bm{\mu}}{\bm{\nu}} + \ln\mleft(\sum_{k=1}^K \nu_k\exp(\psi(k))\mright).
\end{equation*}
\end{boxlem}
\begin{proof}
We have
    \begin{align*}
        \sum_{k=1}^K \mu_k \psi(k) &= \sum_{k=1}^K \mu_k \ln \mleft(\frac{\mu_k}{\nu_k}\exp(\psi(k))\frac{\nu_k}{\mu_k} \mright) \\
        &= \sum_{k=1}^K \mu_k \ln \mleft(\frac{\mu_k}{\nu_k}\mright) + \sum_{k=1}^K \mu_k \ln\mleft(\exp(\psi(k))\frac{\nu_k}{\mu_k} \mright) \\
        &= \kld{\mu}{\nu} + \sum_{k=1}^K \mu_k \ln\mleft(\exp(\psi(k))\frac{\nu_k}{\mu_k} \mright) \tag{Definition of $\kld{\cdot}{\cdot}$}\\
        &\leq \kld{\mu}{\nu} + \ln\mleft(\sum_{k=1}^K \mu_k \exp(\psi(k))\frac{\nu_k}{\mu_k} \mright) \tag{Jensen inequality} \\
        &= \kld{\mu}{\nu} + \ln\mleft(\sum_{k=1}^K \nu_k \exp(\psi(k))\mright).
    \end{align*}
\end{proof}
We now proceed to the proof of Proposition~\eqref{prop:criterion_error_uncertainty}. For a given test sample $\mbf{x}_i \in \mcal{D}_\mrm{test}$, we first notice that 
    \begin{equation*}
        \kld{\mbf{u}}{\mbf{s}^i} = \sum_{k=1}^K \mbf{u}_k \ln \mleft(\frac{\mbf{u}_k}{\mbf{s}^i_{k}}\mright) = \frac{1}{K} \sum_{k=1}^K \ln(\frac{1}{K}) - \ln(\mbf{s}^i_{k}) = -\ln(K) - \frac{1}{K} \sum_{k=1}^K \ln(\mbf{s}^i_{k}).
    \end{equation*}
Similarly, we obtain $\kld{\mbf{s}^i}{\mbf{u}} =\sum_{k=1}^K \mbf{s}^i_{k} \ln(\mbf{s}^i_{k}) + \ln(K)$. Combining those results leads to
\begin{align*}
    &\kld{\mbf{u}}{\mbf{s}^i} + \kld{\mbf{s}^i}{\mbf{u}} = - \frac{1}{K} \sum_{k=1}^K \ln(\mbf{s}^i_{k}) + \sum_{k=1}^K \mbf{s}^i_{k} \ln(\mbf{s}^i_{k}) \\
    \iff &  \kld{\mbf{s}^i}{\mbf{u}} = - \kld{\mbf{u}}{\mbf{s}^i} - \frac{1}{K} \sum_{k=1}^K \ln(\mbf{s}^i_{k}) + \sum_{k=1}^K \mbf{s}^i_{k} \ln(\mbf{s}^i_{k}).
\end{align*}
Consider the function $\psi(k) = \II{y_i\neq k}$ that takes the value $1$ when $y_i\neq k$ and $0$ otherwise. Using Lemma~\ref{lem:change-of-measure-lem} with the measures $\bm{\mu} = \mbf{s}^i, \bm{\nu} = \mbf{u}$ and $\psi$, and the previous equation, we obtain
\begin{align*}
    & \sum_{k=1}^K \mbf{s}^i_{k} \II{y_i\neq k} \leq \kld{\mbf{s}^i}{\mbf{u}}  + \ln\mleft(\sum_{k=1}^K \mbf{u}_k\exp(\II{y_i\neq k})\mright) \\
    \iff & \sum_{k=1}^K \mbf{s}^i_{k} \II{y_i\neq k} \leq - \kld{\mbf{u}}{\mbf{s}^i} - \frac{1}{K} \sum_{k=1}^K \ln(\mbf{s}^i_{k}) + \sum_{k=1}^K \mbf{s}^i_{k} \ln(\mbf{s}^i_{k}) + \ln\mleft(\sum_{k=1}^K \mbf{u}_k\exp(\II{y_i\neq k})\mright) \\
    \iff & 1 - \mbf{s}^i_{y_i} \leq - \kld{\mbf{u}}{\mbf{s}^i} - \frac{1}{K} \sum_{k=1}^K \ln(\mbf{s}^i_{k}) - \mrm{H}(\mbf{s}^i) + \ln\mleft(\frac{1}{K}(Ke + 1)\mright) \tag{$\sum_{k=1}^K \mbf{s}^i_{k} = 1$}\\
    \iff & 1-\mbf{s}^i_{y_i} + \kld{\mbf{u}}{\mbf{s}^i} + \mrm{H}(\mbf{s}^i) \leq - \frac{1}{K} \sum_{k=1}^K \ln(\mbf{s}^i_{k})  + \ln\mleft(e + \frac{1}{K}\mright).
\end{align*}
Summing over all the test samples and dividing by $N$ leads to
\begin{equation*}
    \frac{1}{N}\sum_{i=1}^N (1 -\mbf{s}^i_{y_i}) + \frac{1}{N} \sum_{i=1}^N \kld{\mbf{u}}{\mbf{s}^i} + \frac{1}{N} \sum_{i=1}^N \mrm{H}(\mbf{s}^i) \leq \underbrace{-\frac{1}{NK}\sum_{i=1}^N\sum_{k=1}^K \ln(\mbf{s}^i_{k})}_{=\threshold{}} + \ln\mleft(e + \frac{1}{K}\mright), 
\end{equation*}
which concludes the proof by using the notations introduced in Proposition~\ref{prop:criterion_error_uncertainty}.
\end{proof}
\paragraph{Interpretation.} The term $\xi\mleft(\mcal{D}_\mrm{test}, \mcal{Y}_\mrm{test}\mright)$, dubbed misclassification error, is the average error on the test set between the optimal probability on the true label (\textit{\textit{i.e.}}, $1$) and the predicted probability~$\mbf{s}^i_{y}$. It takes high values when the model makes a lot of mistakes, assigning low confidence to the true class labels, and low values otherwise. $\mcal{U}\mleft(\mcal{D}_\mrm{test}\mright)$ is the average KL divergence on the test set between the predicted probabilities and the uniform distribution and it measures the model's confidence~\citep{tian2021exploring}. It takes high values when the predicted probabilities are far from the uniform (confidence) and low values when they are close to the uniform (uncertain). $\mcal{H}\mleft(\mcal{D}_\mrm{test}\mright)$ is the average entropy on the test set of the predicted probabilities. It takes high values when predicted probabilities are close to the uniform and low values otherwise. Proposition~\ref{prop:criterion_error_uncertainty} implies that when the model makes few mistakes ($\xi\mleft(\mcal{D}_\mrm{test}, \mcal{Y}_\mrm{test}\mright)$ is low) and is confident ($\mcal{U}\mleft(\mcal{D}_\mrm{test}\mright)$ is high and $\mcal{H}\mleft(\mcal{D}_\mrm{test}\mright)$ is low), then the criterion $\threshold{}$ takes high values. This matches the empirical evidence from from~\citet{tian2021exploring}. Proposition~\ref{prop:criterion_error_uncertainty} is harder to analyze in other scenarios, \textit{i.e}, when the misclassification error, the confidence, or the entropy behaves differently, mostly because of the interplay between $\mcal{U}\mleft(\mcal{D}_\mrm{test}\mright)$ and $\mcal{H}\mleft(\mcal{D}_\mrm{test}\mright)$. However, we experimentally show the benefits of $\threshold{}$ and \normalization{} in Section~\ref{sec:experiments} where \method{} achieves superior performance against $11$ commonly used baselines for various architectures and types of shifts on $12$ datasets.

\subsection{Choice of hyperparameter $\eta$}
\label{app:thm_eta}
In all our experiments, we take $\eta=5$ for the selection criterion in Eq.~\eqref{eq:criterion}. We motivate this choice in what follows. In our setting, we consider test samples $\mbf{x}_i \in \mcal{D}_\mrm{test}$ drawn i.i.d. from the test distribution $p_T$. As the model $f$ pre-trained on $\mcal{D}_\mrm{train}$ is a deterministic function, the logits $\mbf{q}_i$ are i.i.d. random variables and the decision threshold $\threshold = -\frac{1}{NK} \sum_{i=1}^N \sum_{k=1}^K \ln\mleft(\exp(\mbf{q}_i)_k/\sum_{j=1}^K\exp(\mbf{q}_i)_j\mright)$ is a random variable with mean $\mu$ and variance $\nu$. Applying the Chebyshev's inequality leads to
\begin{equation}
    \label{eq:chebyshev}
    \PP\mleft(\left\lvert\threshold{} - \mu\right\rvert > \nu\eta\mright) \leq \frac{1}{\eta^2}.
\end{equation}
The threshold $\threshold$ is used to determine how calibrated the model is on a given test dataset $\mcal{D}_\mrm{test}$. Figure~\ref{fig:norm_selection} shows that our proposed normalization is optimal in poorly calibrated datasets and performs slightly below the \texttt{softmax} in calibrated situations. Hence, we can afford to be conservative and we want to consider the model calibrated only for \emph{extreme} values of $\threshold{}$. From Eq.~\eqref{eq:chebyshev}, taking $\eta=5$ ensures that the probability that  $\threshold{}$ deviates from its mean by several standard deviations with probability smaller than $5\%$ ($\frac{1}{25} < 0.05$). It should be noted that we do not claim the optimality of this choice nor the optimality of our automatic selection in Eq.~\eqref{eq:criterion}. However, it is particularly difficult to define decision rules in unsupervised and semi-supervised settings~\citep{amini2023selftraining}. Moreover, using Eq.~\eqref{eq:criterion}, \method{} remains suitable even when test labels are not available which is often the case in real-world applications, and we demonstrate state-of-the-art performance for various architecture and types of shifts in Section~\ref{sec:experiments}. For the sake of self-consistency, we also provide a sensitivity analysis on the values of $\eta$ in Appendix~\ref{app:eta}.

\subsection{Potential Limitations} 
\label{app:limitations}
It should be noted that the selection criterion of Eq.~\eqref{eq:criterion} remains somehow heuristic and might depend on the model, the data, or the threshold $\eta$. As stated above, the chosen value of $\eta$ is motivated by a probabilistic argument and by our experiments. However, as it can be seen in Eq.~\eqref{eq:chebyshev}, the mean and standard deviation of  $\threshold{}$ can impact the validity of $\eta$. In particular, this could be the case when applying \method{} on other data modalities than images or in other learning settings (\textit{e.g.}, classification with a huge number of classes, regression tasks, auto-regressive settings). We believe this is the subject of future work to improve the robustness and versatility of our method.

\paragraph{Impact of the number of classes.} As a first research direction, we provide a motivating example with synthetic data on the impact of the number of classes $K$ on the values of $\threshold{}$. We uniformly draw random vectors of $\RR^K$ in $[-5, 5]$ to mimic the logits obtained from $100000$ independent models. 
 \begin{wrapfigure}{r}{0.5\textwidth}
    \centering
\includegraphics[width=1\linewidth]{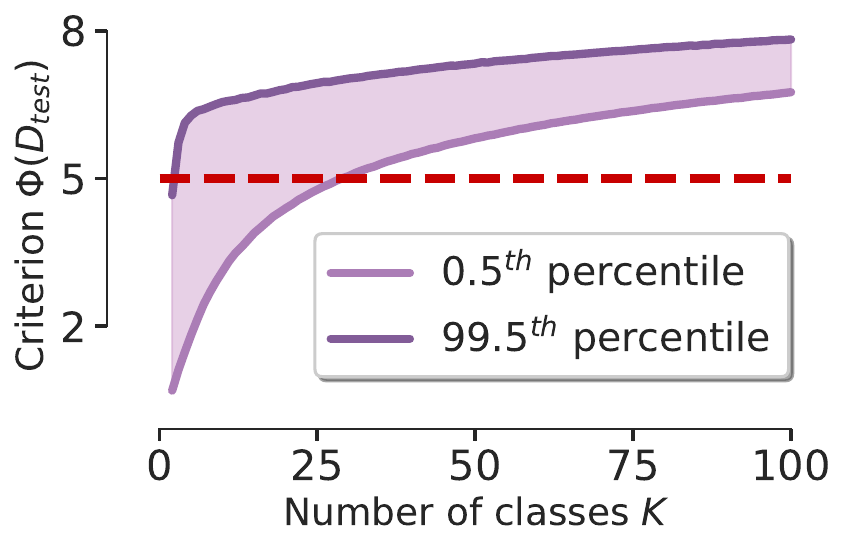}
        \label{fig:criterion}
    \caption{Evolution of the $99\%$ confidence interval on $\threshold{}$ with the number of classes $K$. As $K$ increases, $\threshold{}$ will likely be higher than $\eta = 5$.}
    \label{fig:conf_interval}
\end{wrapfigure}
We compute the corresponding $\threshold{}$ for each model and recover the $0.5^{\mrm{th}}$ and $99.5^{\mrm{th}}$ percentile to obtain a $99\%$ confidence interval. We repeat this experiment for $K \in \llbracket 2, 100 \rrbracket$. The evolution of the confidence interval is displayed in Figure~\ref{fig:conf_interval}. We observe that as soon as $K > 3$, the upper bound of the confidence interval has a very slow increase. However, the lower bound increases quickly in the beginning until $K \sim 25$ and then adopts the same increase pace as the upper bound. In summary, the range of values of $\threshold{}$ becomes thinner and more concentrated on high values for $K > 25$. In particular, we observe that $99\%$ of the models have an associated $\threshold{}$, higher than $\eta =5$, which means that in this situation, the \texttt{softmax} would always be selected as a normalization $\sigma$. While the conclusions from this experiment are not directly applicable to our real experimental setting (in particular, Taylor and \texttt{softmax} cases of Eq.~\eqref{eq:criterion} occur both for datasets with $K> 25$ and $K \leq 25$), we believe it motivates further work to make $\eta$ more robust to the values of $\threshold{}$. In particular, one could propose to compute $\eta$ based on statistics of the data or as a function of the number of classes. 

\paragraph{Dispersion of the softmax.} In the previous paragraph, we showed that the number of classes $K$ can have an impact on the values taken by criterion $\threshold{}$. As the criterion relies on the softmax probabilities of the model on test data, it is natural to investigate the impact of $K$ on the \texttt{softmax} function. The lemma below shows that the \texttt{softmax} must disperse on all entries as the number of classes $K$ increases. It means that the total weights on the \texttt{softmax} entries (equal to $1$) cannot be concentrated on a few entries as the number of classes $K$ increases.
\begin{boxlem}
    \label{lem:ineq_softmax}
    Let $\bm{\theta} \in \RR^K$ be logits of a neural network $f$ such that $\lVert \bm{\theta} \rVert_1 \leq c$ for some $c > 0$. Then, as the number of classes grows, \textit{i.e.}, $K \to \infty$, we have
    \begin{equation*}
        \softmax{\bm{\theta}} = \mcal{O} \mleft( \frac{1}{K} \mright),
    \end{equation*}
    where the equality holds at the component level.
\end{boxlem}
\begin{proof}
    Following the proof of~\citet[Lemma D.7]{zekri2024largelanguagemodelsmarkov}, we can show for all $i \in [K]$ that
    \begin{align*}
        \frac{\exp{\mleft(-c\mright)}}{\sum_{j=1}^K \exp{\mleft(c\mright)}} &\leq \frac{\exp{\mleft(\bm{\theta}_i\mright)}}{\sum_{j=1}^m \exp{\mleft(\bm{\theta}_j\mright)}} \leq \frac{\exp{\mleft(c\mright)}}{\sum_{j=1}^K \exp{\mleft(-c\mright)}} \iff 
        \frac{a}{K} \leq \softmax{\bm{\theta}}_i \leq \frac{b}{K},
    \end{align*}
where $a = \exp{(-2c)}, b = \exp{(2c)}$ are constant. This concludes the proof.
\end{proof}
Lemma~\ref{lem:ineq_softmax} implies that, as the number of classes grows, the highest value an individual entry can have decreases. Hence, the number of classes impacts the distribution of the weights among the \texttt{softmax} entries (recalling that it must sum at $1$ as it is a probability vector). As by definition, $\threshold{}$ depends on the \texttt{softmax} probability distributions, this will impact its value. We believe that empirically and theoretically studying this phenomenon could be insightful in deriving more robust selection criteria and threshold values. We note that Lemma~\ref{lem:ineq_softmax} is similar but more general than~\citet[Lemma 2.1]{velickovic2024softmax} as our global bounding condition on $\bm{\theta}$ encompasses their entry-wise condition. 

\section{Beyond ResNets: Experiments with Vision Transformers and ConvNeXts}
\label{app:more_arch}
To evaluate the efficiency of \method{} across diverse model architectures, we conducted additional experiments with the Vision Transformer~\citep[ViT]{dosovitskiy2020image} and the ConvNeXt~\citep{liu2022convnet} architectures. The numerical results are gathered in Table~\ref{tab:arch}. Two methods stand out from the rest of the baselines: \textit{ConfScore} and \method{}. In particular, \textit{ConfScore} is particularly strong with the ConvNeXt architecture while \method{} is better with the Vision Transformer. Overall, \method{} leads to a better accuracy estimation on average.

\begin{table*}[!h]
    \centering
    \caption{Method comparison using Vision Transformer (ViT) and ConvNext under \textbf{synthetic, subpopulation and natural shifts} with $R^2$ and $\rho$ metrics (the higher the better). The best results for each metric are in \textbf{bold}. Overall, \method{} surpasses its competitors while \textit{ConfScore} appears to be stronger with ViT and ConvNext than with ResNets.
    }
    \scalebox{0.65}{
    \begin{tabular}{cccccccccccccccc}
        \toprule
        \multirow{2}{*}{Dataset} &\multirow{2}{*}{Network} &\multicolumn{2}{c}{ConfScore} &\multicolumn{2}{c}{Entropy} &\multicolumn{2}{c}{ATC} &\multicolumn{2}{c}{MDE} &\multicolumn{2}{c}{COT} &\multicolumn{2}{c}{Nuclear} &\multicolumn{2}{c}{\method{}}\\
        \cline{3-16}
        &  &$R^2$ &$\rho$ &$R^2$ &$\rho$ &$R^2$ &$\rho$&$R^2$ &$\rho$&$R^2$ &$\rho$&$R^2$ &$\rho$&$R^2$ &$\rho$\\
        \midrule
         \multirow{3}{*}{CIFAR 10} &ViT &0.985 &0.996 &0.980 &0.996 &0.991 &0.997 &0.871 &0.873 &0.950 &\textbf{0.997} &0.937 &0.988 &\textbf{0.991} &0.996\\
          &ConvNeXt &0.936 &0.996 &0.924 &0.995 &0.911 &0.994 &0.002 &0.534 &0.978 &0.994 &\textbf{0.991} &0.994 &0.916 &\textbf{0.995}\\
          \cline{2-16}
          & \textcolor[rgb]{0.0, 0.53, 0.74}{Average} &\textcolor[rgb]{0.0, 0.53, 0.74}{0.961} &\textcolor[rgb]{0.0, 0.53, 0.74}{0.996}&\textcolor[rgb]{0.0, 0.53, 0.74}{0.952} &\textcolor[rgb]{0.0, 0.53, 0.74}{0.996} &\textcolor[rgb]{0.0, 0.53, 0.74}{0.951} &\textcolor[rgb]{0.0, 0.53, 0.74}{0.996} &\textcolor[rgb]{0.0, 0.53, 0.74}{0.435} &\textcolor[rgb]{0.0, 0.53, 0.74}{0.704} &\textcolor[rgb]{0.0, 0.53, 0.74}{\textbf{0.964}} &\textcolor[rgb]{0.0, 0.53, 0.74}{0.996} &\textcolor[rgb]{0.0, 0.53, 0.74}{0.964} &\textcolor[rgb]{0.0, 0.53, 0.74}{0.991} &\textcolor[rgb]{0.0, 0.53, 0.74}{0.954} &\textcolor[rgb]{0.0, 0.53, 0.74}{\textbf{0.996}}\\
          \midrule
         \multirow{3}{*}{CIFAR 100} &ViT &0.983 &\textbf{0.997} &0.981 &0.995 &0.987 &0.995 &0.974 &0.983 &\textbf{0.993} &0.996 &0.977 &0.995 &0.989 &0.996\\
          &ConvNeXt &0.976 &\textbf{0.995} &0.957 &0.992 &0.976 &0.993 &0.617 &0.399 &0.981 &0.996 &\textbf{0.982} &0.994 &0.954 &0.994\\
          \cline{2-16}
          & \textcolor[rgb]{0.0, 0.53, 0.74}{Average} &\textcolor[rgb]{0.0, 0.53, 0.74}{0.961} &\textcolor[rgb]{0.0, 0.53, 0.74}{\textbf{0.996}} &\textcolor[rgb]{0.0, 0.53, 0.74}{0.952} &\textcolor[rgb]{0.0, 0.53, 0.74}{\textbf{0.996}} &\textcolor[rgb]{0.0, 0.53, 0.74}{0.982} &\textcolor[rgb]{0.0, 0.53, 0.74}{0.994} &\textcolor[rgb]{0.0, 0.53, 0.74}{0.795}&\textcolor[rgb]{0.0, 0.53, 0.74}{0.691} &\textcolor[rgb]{0.0, 0.53, 0.74}{\textbf{0.987}} &\textcolor[rgb]{0.0, 0.53, 0.74}{\textbf{0.996}} &\textcolor[rgb]{0.0, 0.53, 0.74}{0.977} &\textcolor[rgb]{0.0, 0.53, 0.74}{0.995} &\textcolor[rgb]{0.0, 0.53, 0.74}{0.971} &\textcolor[rgb]{0.0, 0.53, 0.74}{0.995}\\
          \midrule
         \multirow{3}{*}{PACS} & ViT &0.711 &0.783 &0.631 &0.727 &0.426 &0.503 &0.180 &0.209 &0.742 &0.797 &\textbf{0.823} &\textbf{0.860} &0.705 &0.755\\
          &ConvNeXt &\textbf{0.900} &\textbf{0.895} &0.872 &0.853 &0.727 &0.580 &0.004 &0.062 &0.814 &0.748 &0.834 &0.790 &0.874 &0.755\\
          \cline{2-16}
          & \textcolor[rgb]{0.0, 0.53, 0.74}{Average} &\textcolor[rgb]{0.0, 0.53, 0.74}{0.806} &\textcolor[rgb]{0.0, 0.53, 0.74}{\textbf{0.839}} &\textcolor[rgb]{0.0, 0.53, 0.74}{0.752} &\textcolor[rgb]{0.0, 0.53, 0.74}{0.790} &\textcolor[rgb]{0.0, 0.53, 0.74}{0.577} &\textcolor[rgb]{0.0, 0.53, 0.74}{0.541} &\textcolor[rgb]{0.0, 0.53, 0.74}{0.092} &\textcolor[rgb]{0.0, 0.53, 0.74}{0.073} &\textcolor[rgb]{0.0, 0.53, 0.74}{0.778} &\textcolor[rgb]{0.0, 0.53, 0.74}{0.772} &\textcolor[rgb]{0.0, 0.53, 0.74}{\textbf{0.829}} &\textcolor[rgb]{0.0, 0.53, 0.74}{0.825} &\textcolor[rgb]{0.0, 0.53, 0.74}{0.789} &\textcolor[rgb]{0.0, 0.53, 0.74}{0.755}\\
          \midrule
         \multirow{3}{*}{Office-Home} &ViT &0.947 &0.958 &0.928 &0.979 &0.896 &0.902 &0.217 &0.755 &0.642 &0.856 &0.861 &0.958 &\textbf{0.953} &\textbf{0.979}\\
          &ConvNeXt &\textbf{0.784} &0.860 &0.649 &0.825 &0.769 &\textbf{0.923} &0.040 &0.475 &0.642 &0.856 &0.514 &0.514 &0.733 &0.818\\
          \cline{2-16}
          & \textcolor[rgb]{0.0, 0.53, 0.74}{Average} &\textcolor[rgb]{0.0, 0.53, 0.74}{\textbf{0.865}} &\textcolor[rgb]{0.0, 0.53, 0.74}{0.909} &\textcolor[rgb]{0.0, 0.53, 0.74}{0.788} &\textcolor[rgb]{0.0, 0.53, 0.74}{0.902} &\textcolor[rgb]{0.0, 0.53, 0.74}{0.832} &\textcolor[rgb]{0.0, 0.53, 0.74}{\textbf{
          0.912}} &\textcolor[rgb]{0.0, 0.53, 0.74}{0.128} &\textcolor[rgb]{0.0, 0.53, 0.74}{0.615} &\textcolor[rgb]{0.0, 0.53, 0.74}{0.642} &\textcolor[rgb]{0.0, 0.53, 0.74}{0.856} &\textcolor[rgb]{0.0, 0.53, 0.74}{0.687} &\textcolor[rgb]{0.0, 0.53, 0.74}{0.736} &\textcolor[rgb]{0.0, 0.53, 0.74}{0.843} &\textcolor[rgb]{0.0, 0.53, 0.74}{0.898}\\
          \midrule
         \multirow{3}{*}{Entity-13} &ViT &0.930 &0.950 &0.925 &0.950 &0.950 &0.971 &0.816 &0.884 &0.923 &0.958 &0.873 &0.882 &\textbf{0.958} &\textbf{0.971}\\
          &ConvNeXt &\textbf{0.943} &\textbf{0.970} &0.931 &0.960 &0.901 &0.902 &0.868 &0.805 &0.937 &0.938 &0.941 &0.942 &0.930 &0.963\\
          \cline{2-16}
          & \textcolor[rgb]{0.0, 0.53, 0.74}{Average} &\textcolor[rgb]{0.0, 0.53, 0.74}{0.937} &\textcolor[rgb]{0.0, 0.53, 0.74}{0.960} &\textcolor[rgb]{0.0, 0.53, 0.74}{0.928} &\textcolor[rgb]{0.0, 0.53, 0.74}{0.955} &\textcolor[rgb]{0.0, 0.53, 0.74}{0.926} &\textcolor[rgb]{0.0, 0.53, 0.74}{0.937} &\textcolor[rgb]{0.0, 0.53, 0.74}{0.842} &\textcolor[rgb]{0.0, 0.53, 0.74}{0.844} &\textcolor[rgb]{0.0, 0.53, 0.74}{0.930} &\textcolor[rgb]{0.0, 0.53, 0.74}{0.948} &\textcolor[rgb]{0.0, 0.53, 0.74}{0.907} &\textcolor[rgb]{0.0, 0.53, 0.74}{0.912} &\textcolor[rgb]{0.0, 0.53, 0.74}{\textbf{0.944}} &\textcolor[rgb]{0.0, 0.53, 0.74}{\textbf{0.967}}\\
          \midrule
         \multirow{3}{*}{Entity-30} &ViT &0.950 &0.972 &0.937 &0.968 &0.948 &0.970 &0.819 &0.908 &0.939 &0.971 &0.905 &0.927&\textbf{0.959} &\textbf{0.975}\\
          &ConvNeXt &\textbf{0.968} &0.988 &0.955 &0.981 &0.916 &0.936 &0.959 &0.961 &0.942 &0.976 &0.942 &0.959 &0.960 &\textbf{0.990}\\
          \cline{2-16}
          & \textcolor[rgb]{0.0, 0.53, 0.74}{Average} &\textcolor[rgb]{0.0, 0.53, 0.74}{0.959} &\textcolor[rgb]{0.0, 0.53, 0.74}{0.980} &\textcolor[rgb]{0.0, 0.53, 0.74}{0.946} &\textcolor[rgb]{0.0, 0.53, 0.74}{0.975} &\textcolor[rgb]{0.0, 0.53, 0.74}{0.932} &\textcolor[rgb]{0.0, 0.53, 0.74}{0.953} &\textcolor[rgb]{0.0, 0.53, 0.74}{0.889} &\textcolor[rgb]{0.0, 0.53, 0.74}{0.934} &\textcolor[rgb]{0.0, 0.53, 0.74}{0.941} &\textcolor[rgb]{0.0, 0.53, 0.74}{0.973} &\textcolor[rgb]{0.0, 0.53, 0.74}{0.924} &\textcolor[rgb]{0.0, 0.53, 0.74}{0.943} &\textcolor[rgb]{0.0, 0.53, 0.74}{\textbf{0.960}} &\textcolor[rgb]{0.0, 0.53, 0.74}{\textbf{0.982}}\\
         \bottomrule
     \end{tabular}
     }
    \label{tab:arch}
\end{table*}

\section{Sensitivity Analysis and Ablation Study}
\label{app:additional_exp}
\subsection{Choice of $L_p$ Norm} 
\label{app:choice_lp}
\begin{figure}[!t]
    \centering
    \subfigure[Impact of the $L_p$ norm]{
        \centering
        \includegraphics[width=0.3\textwidth]{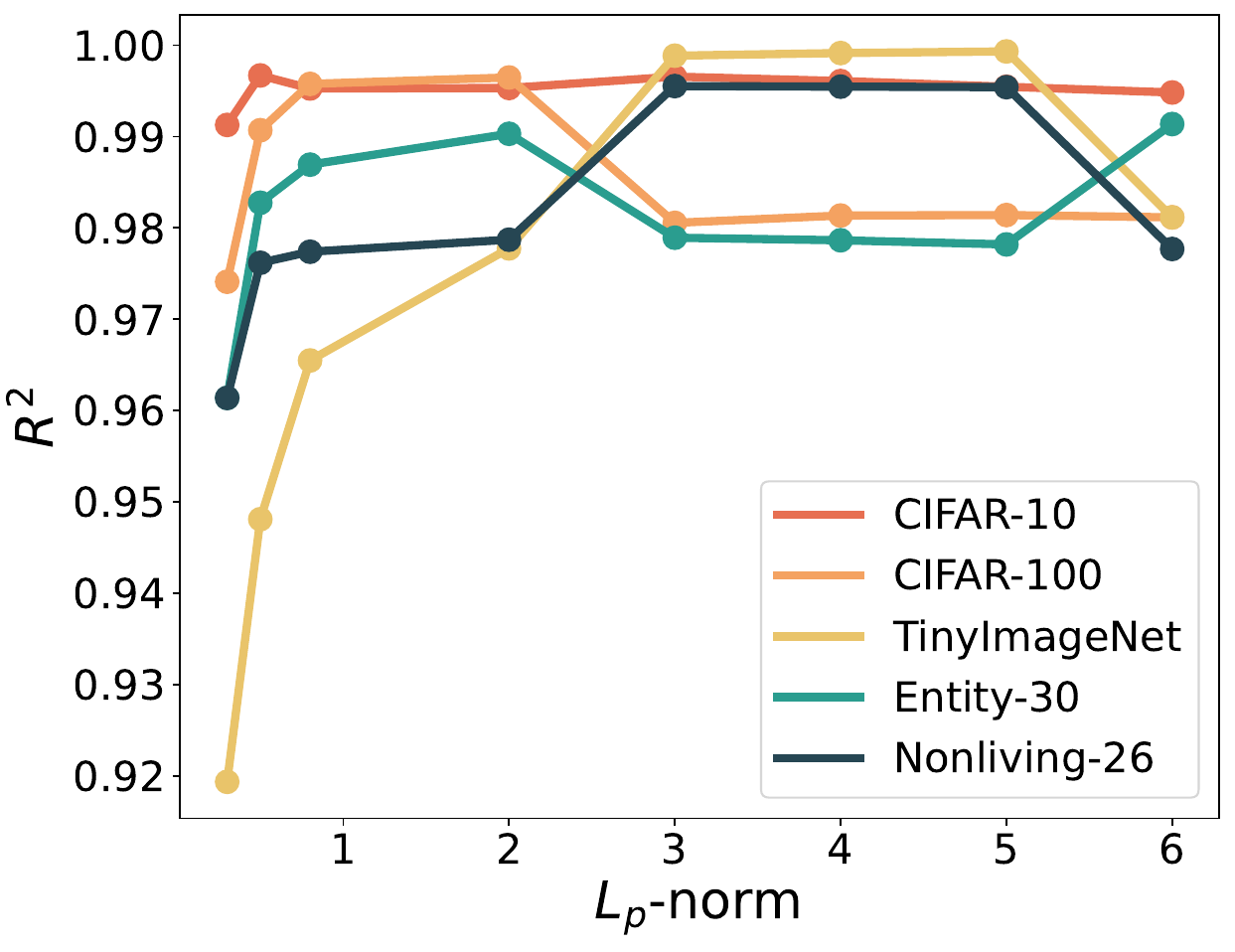}
        \label{fig:l_p}}
    \subfigure[Approximation order $n$]{
        \centering
        \includegraphics[width=0.3\textwidth]{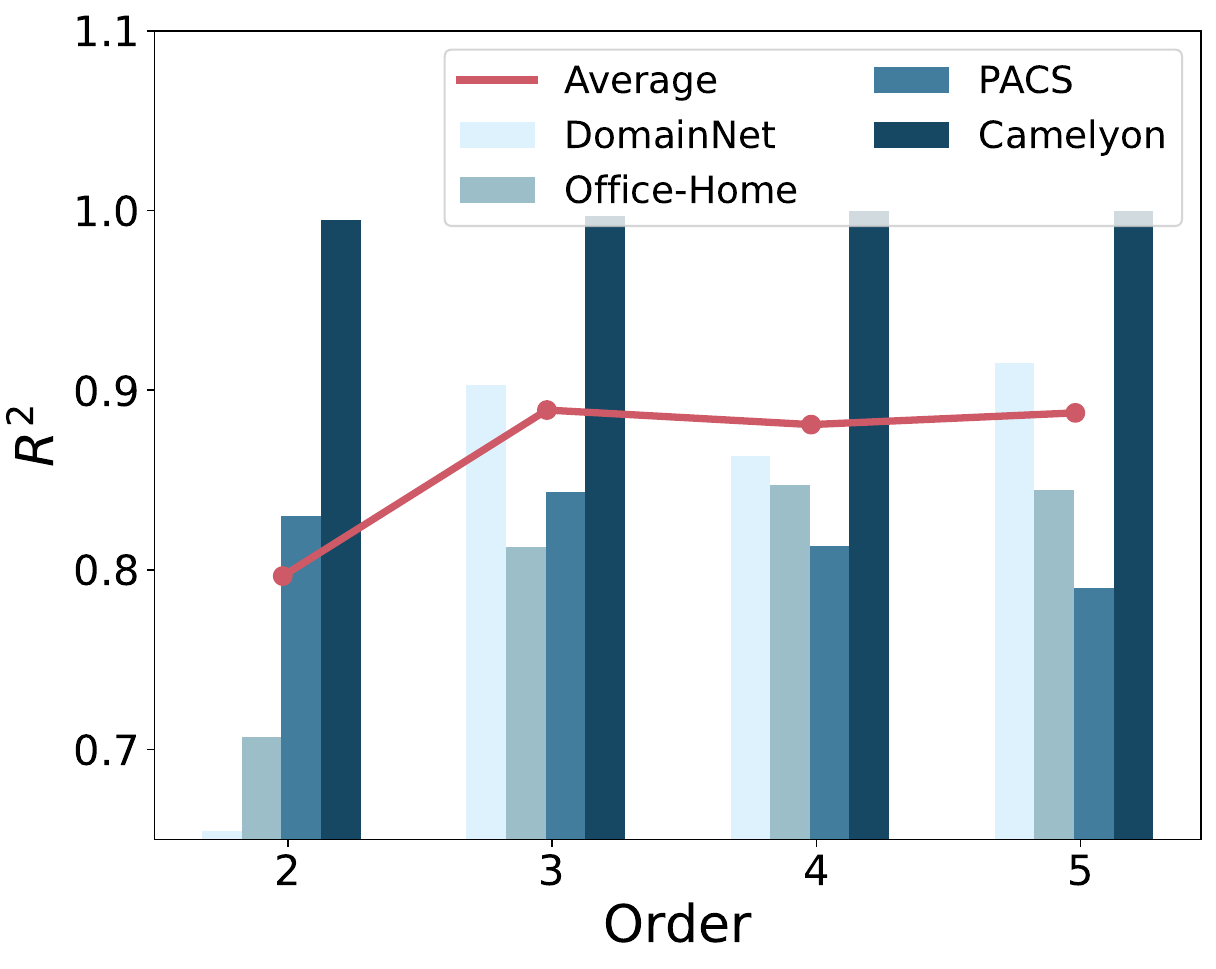}
        \label{fig:taylor}}
     \subfigure[Benefits of \normalization{}]{
        \centering
        \includegraphics[width=0.3\textwidth]{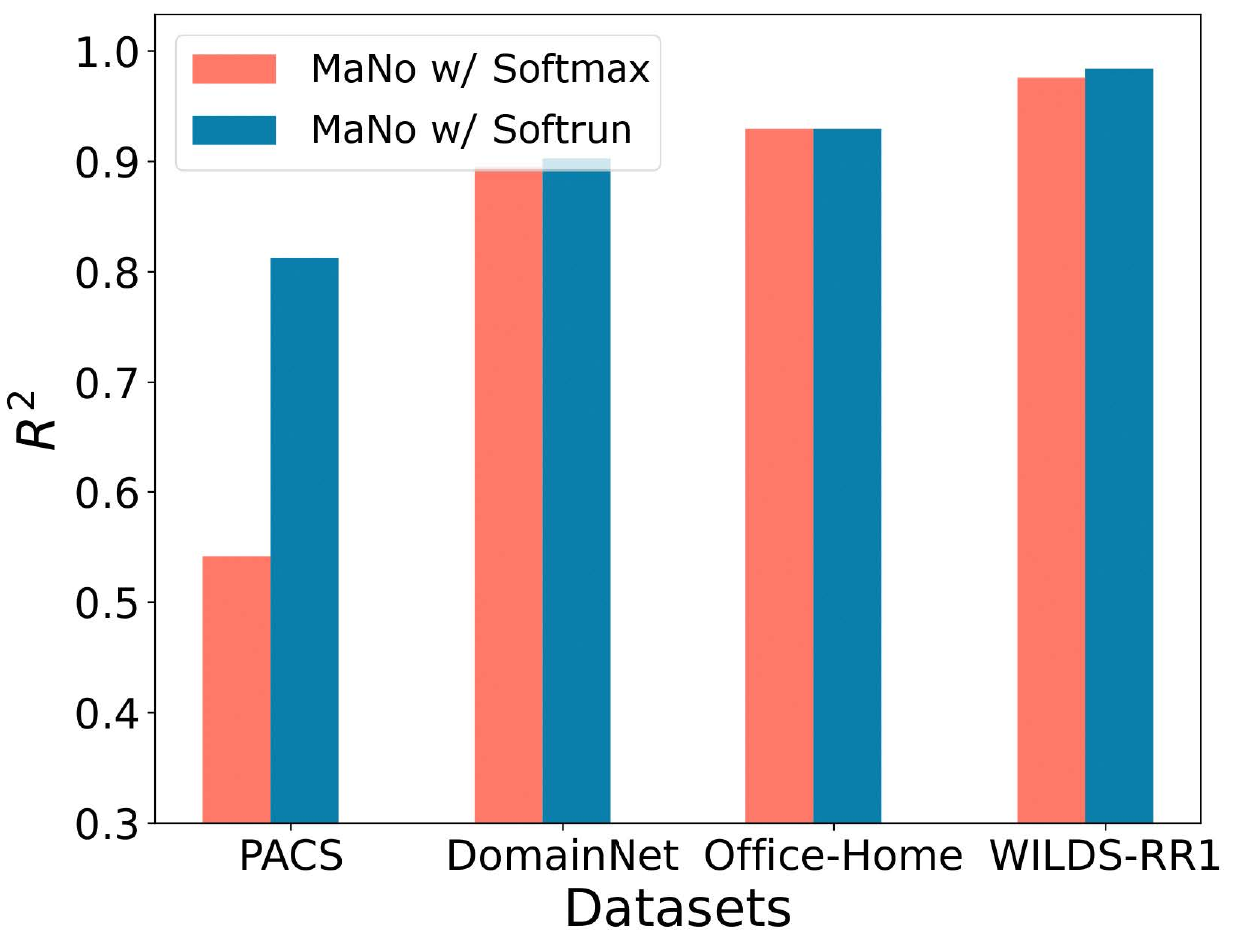}
        \label{fig:normalization}}
     \caption{\textbf{Sensitivity analysis with Resnet18.} \textbf{(a)} Effect of the $L_p$ norm types. \textbf{(b)} Impact of the Taylor approximation order, \textit{i.e.}, the number of terms in Eq~\ref{eq:taylor}. For instance, an order of $3$ means that $3$ terms are taken, which corresponds to the default setting in Eq.~\eqref{eq:criterion} and is used in all our experiments. \textbf{(c)} Type of normalization function.}
     \vspace{-10pt}
\end{figure}

To reflect the impact of different $L_p$ norms on estimation performance, we conduct a sensitivity study on $5$ datasets with ResNet18, whose results are shown in Figure~\ref{fig:l_p}. The performance for $p=1$ is ignored as in this case, $\lVert \mbf{Q} \rVert_1=1$ because $\mbf{Q}$ is right-stochastic. We can see that when we choose $p \in [2, 5]$, the results fluctuate within a satisfying range. This can be explained by the fact that within this range, we emphasize adequately the large positive feature-to-boundary distances without ignoring the other comparatively small distances. 

\subsection{Choice of Taylor Approximation Order} 
In Figure~\ref{fig:taylor}, we verify the impact of Taylor formula approximation on final accuracy estimation performance. It should be noted that for orders higher than in the default setting in Eq.~\eqref{eq:criterion}, the positivity is lost. To alleviate this issue, we consistently remove for all orders the minimum value of the obtained vector to each of its entries to ensure having an output in $\RR^K_+$. This extends~\citet{brebisson2016taylor} to orders higher than $2$. From this figure, we can see that when we reserve the first three terms in the Taylor formula, the average estimation performance is optimal. For well-calibrated datasets such as Office-Home and WILDS, there exists an increased trend of estimation performance when we reserve more Taylor formula terms. As for suboptimal-calibrated datasets such as PACS and Office-31, their performance rises when fewer terms are reserved. It empirically certifies that the normalization technique is a trade-off tool between the ground-truth logits' information and error accumulation. In addition, the optimal choice is to keep $3$ terms in Eq.~\eqref{eq:taylor} which motivates our default setting in Eq~\eqref{eq:criterion}.

\subsection{Choice of Calibration Threshold $\eta$}
\label{app:eta}
In Table~\ref{tab:eta}, we display the performance comparison for varying values of threshold $\eta$ on three datasets with ResNet18. It should be noted that taking $\eta=0$ corresponds to the case where the \texttt{softmax} is always taken, i.e., the common choice in the literature. This matches our theoretical insights in Appendix~\ref{app:thm_eta} and confirms that taking $\eta = 5$ is a robust and effective choice for $\normalization{}$. 
\begin{table}[!h]
\centering
\caption{Performance comparison for varying $\eta \in \{0, 1, 3, 5, 7, 9\}$ on CIFAR-10, Office-Home, and PACS with ResNet18. The metric used is $R^2$ (the higher the better). The best results are in \textbf{bold}. The results motivate our choice of $\eta = 5$.}
\scalebox{1}{
\begin{tabular}{ccccccc}
\toprule
Dataset & $\eta = 0$ &$\eta = 1$  &$\eta = 3$ &$\eta = 5$&$\eta = 7$ &$\eta = 9$\\
\midrule
Cifar-10 & \textbf{0.995}&	\textbf{0.995}& \textbf{0.995}& \textbf{0.995}& \textbf{0.995}& \textbf{0.995}\\
 Office-Home &\textbf{0.926}&	\textbf{0.926}&	\textbf{0.926}&	\textbf{0.926}&	0.777&	0.777\\
PACS &	0.541&	0.541&	0.541&	\textbf{0.827}&	\textbf{0.827}&	\textbf{0.827}\\
\midrule
\textcolor[rgb]{0.0, 0.53, 0.74}{Average} &\textcolor[rgb]{0.0, 0.53, 0.74}{0.820}&	\textcolor[rgb]{0.0, 0.53, 0.74}{0.820}&	\textcolor[rgb]{0.0, 0.53, 0.74}{0.820}&	\textcolor[rgb]{0.0, 0.53, 0.74}{\textbf{0.916}}&	\textcolor[rgb]{0.0, 0.53, 0.74}{0.866}&	\textcolor[rgb]{0.0, 0.53, 0.74}{0.866}\\
\bottomrule
\end{tabular}}
\label{tab:eta}
\end{table}

\subsection{Superiority of \normalization{}} 
To verify the effectiveness of our proposed normalization technique, \normalization{}, we conduct an ablation study by replacing our normalization with the softmax function under the natural shift. In Figure~\ref{fig:normalization}, we observe that our proposed normalization significantly enhances the estimation performance of datasets from the natural shift. Especially, $R^2$ for poorly-calibrated datasets such as PACS is improved from 0.541 to 0.812.

\subsection{Generalization Capabilities of \method{} on ImageNet-$\Bar{C}$} 
\label{app:generalization_capability}
\begin{figure}[!h]
    \includegraphics[width=\linewidth]{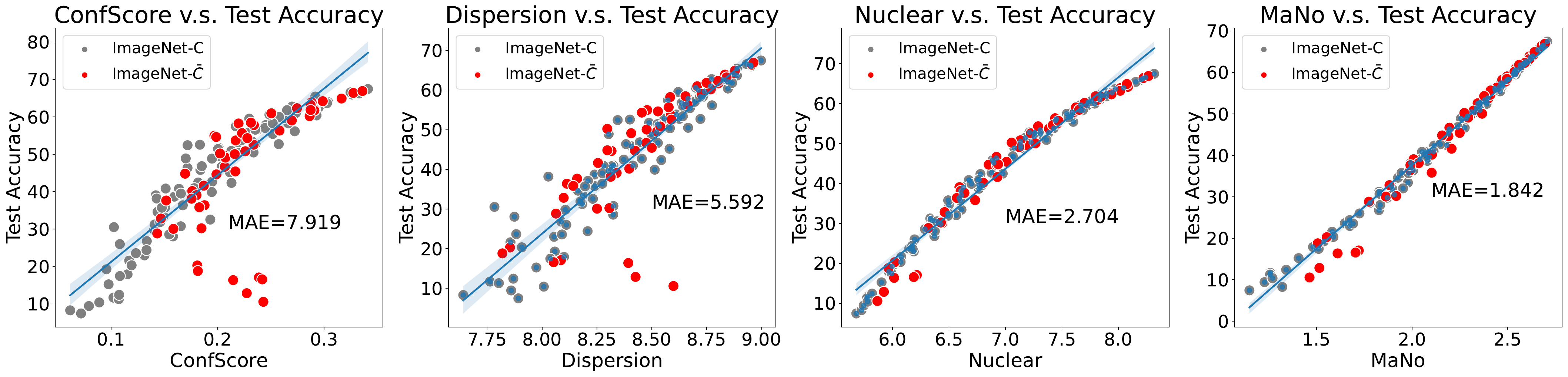}
    \vspace{-10pt}
    \caption{Comparison of generalization capability across four methods. Each subplot shows a linear regression model fitted on ImageNet-C to predict accuracy on ImageNet-$\Bar{C}$. Mean absolute error (MAE) is calculated on ImageNet-$\Bar{C}$~\citep{mintun2021interaction}. All experiments use ResNet18.}
    \label{fig:imagenet_c_bar}
    \vspace{-10pt}
\end{figure}

To further demonstrate the generalization capability of \method{}, we provide a similar experiment with that in Section~\ref{sec:discussion} on ImageNet-C and ImageNet-$\Bar{C}$ \citep{mintun2021interaction} in Figure~\ref{fig:imagenet_c_bar}. In particular, we fit a linear regression function on ImageNet-C and use the linear function to predict the accuracy of ImageNet-$\Bar{C}$. This figure shows that \method{} has better estimation performance than the other baselines when meeting different corruption types.

\section{How to Use \method{} in Real-World Applications?}
\label{app:mano_practice}
This work demonstrates the strong correlation between ground-truth OOD accuracy and the designed score, which can be particularly useful for model deployment applications. In this section, we provide two examples.
\begin{itemize}
    \item \textbf{Find difficult (under-performed) test set.} In cases such as retraining on under-performed datasets or annotating hard datasets, we only need to know the rank of datasets by accuracy. Therefore, we can calculate the proposed score for each dataset directly and fulfill the task based on this score's ranking.
    \item \textbf{Deployement risk estimation.} When deploying the model into production, it is important to estimate its safety. If the cost of getting test labels is prohibitive, our method can help to estimate the model's accuracy on the product's test data. A practitioner can additionally look at the variability of the score on multiple test sets. When multiple datasets are not available, we can alternatively construct adequate synthetic datasets via various visual transformations. 
\end{itemize}

In Table~\ref{tab:mae}, we provide an example, using 90\% of datasets to train a linear regression model and estimating the test accuracy of the rest 10\% of datasets via the trained linear regression model. The results are measured by Mean Absolute Error (MAE). From this table, we observe the superiority of \method{} for application in the real world.

\begin{table*}[!t]
    \centering
    \caption{Method comparison on four benchmarks using ResNet18, ResNet50, and WRN-50-2 under \textbf{natural shift} with the MAE metric (the lower the better). The best results are highlighted in \textbf{bold}. \method{} provides the best accuracy estimation overall.}
    \scalebox{0.75}{
    \begin{tabular}{ccccccccccc}
        \toprule
        \multirow{1}{*}{Dataset} &\multirow{1}{*}{Network}  &\multicolumn{1}{c}{ConfScore} &\multicolumn{1}{c}{Entropy}  &\multicolumn{1}{c}{ATC} &\multicolumn{1}{c}{Fr\'{e}chet} &\multicolumn{1}{c}{Dispersion} &\multicolumn{1}{c}{MDE} &\multicolumn{1}{c}{COT} &\multicolumn{1}{c}{Nuclear} &\multicolumn{1}{c}{\method{}}\\
        \midrule
          \multirow{4}{*}{CIFAR-10} & ResNet18 &5.131 &5.265 &3.968 &1.964 &3.842 &1.763 &0.952 &1.357 &\textbf{0.394}\\
          & ResNet50 &1.945 &1.891 &1.706 &2.477 &3.842 &0.928 &0.670 &1.024 &\textbf{0.450}\\
          & WRN-50-2 &2.956 &3.160 &3.086 &3.547 &2.671 &0.846 &\textbf{0.355} &0.843 &0.406\\
          \cline{2-11}
          & \textcolor[rgb]{0.0, 0.53, 0.74}{Average} & \textcolor[rgb]{0.0, 0.53, 0.74}{3.344} & \textcolor[rgb]{0.0, 0.53, 0.74}{3.439}& \textcolor[rgb]{0.0, 0.53, 0.74}{2.920} & \textcolor[rgb]{0.0, 0.53, 0.74}{2.662} & \textcolor[rgb]{0.0, 0.53, 0.74}{2.539} & \textcolor[rgb]{0.0, 0.53, 0.74}{1.179} & \textcolor[rgb]{0.0, 0.53, 0.74}{0.659}& \textcolor[rgb]{0.0, 0.53, 0.74}{1.075}& \textcolor[rgb]{0.0, 0.53, 0.74}{\textbf{0.417}}\\
          \midrule
        \multirow{4}{*}{CIFAR-100} & ResNet18 &2.944 &5.265 &3.968&1.964 &3.846 &1.763 &0.952 &1.357 &\textbf{0.394}\\
          & ResNet50 &2.128 &1.891 &1.706 &2.477 &2.671 &0.928  &0.670 &1.024 &\textbf{0.450}\\
          & WRN-50-2 &1.323 &3.160 &3.086 &3.547 &1.102 &0.846 &\textbf{0.355} &0.843 &0.406\\
          \cline{2-11}
          & \textcolor[rgb]{0.0, 0.53, 0.74}{Average} & \textcolor[rgb]{0.0, 0.53, 0.74}{3.344} & \textcolor[rgb]{0.0, 0.53, 0.74}{3.439}& \textcolor[rgb]{0.0, 0.53, 0.74}{2.920} & \textcolor[rgb]{0.0, 0.53, 0.74}{2.662} & \textcolor[rgb]{0.0, 0.53, 0.74}{2.539} & \textcolor[rgb]{0.0, 0.53, 0.74}{1.179} & \textcolor[rgb]{0.0, 0.53, 0.74}{0.659} & \textcolor[rgb]{0.0, 0.53, 0.74}{1.075} & \textcolor[rgb]{0.0, 0.53, 0.74}{\textbf{0.417}}\\
          \midrule
   \multirow{4}{*}{TinyImageNet} & ResNet18 &3.822 &3.559 &3.752 &5.998 &2.822 &1.926 &1.297 &1.165 &\textbf{0.612}\\
          & ResNet50 &3.376 &3.696 &3.435 &5.616 &1.667 &2.687 &1.653 &1.226 &\textbf{1.005}\\
          & WRN-50-2 &2.712 &2.854 &5.011 &4.862 &1.054 &1.703 &1.286 &\textbf{1.053} &1.145\\
          \cline{2-11}
          & \textcolor[rgb]{0.0, 0.53, 0.74}{Average} & \textcolor[rgb]{0.0, 0.53, 0.74}{3.303} & \textcolor[rgb]{0.0, 0.53, 0.74}{3.370} & \textcolor[rgb]{0.0, 0.53, 0.74}{4.066} & \textcolor[rgb]{0.0, 0.53, 0.74}{5.492} & \textcolor[rgb]{0.0, 0.53, 0.74}{1.848} & \textcolor[rgb]{0.0, 0.53, 0.74}{2.106} & \textcolor[rgb]{0.0, 0.53, 0.74}{1.412} & \textcolor[rgb]{0.0, 0.53, 0.74}{1.148} & \textcolor[rgb]{0.0, 0.53, 0.74}{\textbf{0.921}}\\
          \midrule
          \multirow{4}{*}{ImageNet} & ResNet18 &3.616 &3.812 &2.503 &3.750 &2.602 &4.818 &0.705 &2.679 &\textbf{1.057}\\
          & ResNet50 &3.325 &3.357 &2.911 &3.242 &7.318 &6.346 &1.796 &3.203 &\textbf{1.388}\\
          & WRN-50-2 &2.990 &4.132 &3.388 &5.210 &10.050 &6.755 &2.564 &4.091 &\textbf{0.695}\\
          \cline{2-11}
          & \textcolor[rgb]{0.0, 0.53, 0.74}{Average} & \textcolor[rgb]{0.0, 0.53, 0.74}{3.310} & \textcolor[rgb]{0.0, 0.53, 0.74}{3.767} & \textcolor[rgb]{0.0, 0.53, 0.74}{2.934} & \textcolor[rgb]{0.0, 0.53, 0.74}{4.067} & \textcolor[rgb]{0.0, 0.53, 0.74}{6.656} & \textcolor[rgb]{0.0, 0.53, 0.74}{5.973} & \textcolor[rgb]{0.0, 0.53, 0.74}{1.688} & \textcolor[rgb]{0.0, 0.53, 0.74}{3.325} & \textcolor[rgb]{0.0, 0.53, 0.74}{\textbf{1.047}} \\
          \midrule
          \multirow{4}{*}{Office-Home} & ResNet18 &1.987 &2.878 &10.014 &1.404 &7.976 &8.228 &\textbf{0.602} &0.885 &0.880\\
          & ResNet50 &0.855 &1.874 &7.509 &1.254 &7.285 &9.374 &\textbf{0.730} &2.176 &3.995\\
          & WRN-50-2 &2.607 &3.933 &11.272 &0.924 &7.591 &7.974 &\textbf{2.382} &2.764 &4.330\\
          \cline{2-11}
          & \textcolor[rgb]{0.0, 0.53, 0.74}{Average} & \textcolor[rgb]{0.0, 0.53, 0.74}{1.816} & \textcolor[rgb]{0.0, 0.53, 0.74}{2.895} & \textcolor[rgb]{0.0, 0.53, 0.74}{9.598} & \textcolor[rgb]{0.0, 0.53, 0.74}{1.947} & \textcolor[rgb]{0.0, 0.53, 0.74}{7.617} & \textcolor[rgb]{0.0, 0.53, 0.74}{8.525} & \textcolor[rgb]{0.0, 0.53, 0.74}{\textbf{1.238}} & \textcolor[rgb]{0.0, 0.53, 0.74}{1.941} & \textcolor[rgb]{0.0, 0.53, 0.74}{3.068} \\
          \midrule
          \multirow{4}{*}{DomainNet} & ResNet18 &\textbf{3.375} &7.067 &10.875 &7.416 &8.200 &7.066 &5.954 &7.313 &3.830\\
          & ResNet50 &4.778 &7.742 &10.039 &3.533 &8.949 &9.917 & 5.005 &7.407 &\textbf{4.230}\\
          & WRN-50-2 &\textbf{5.513} &6.310 &3.513 &10.001 &8.695 &9.230 &4.605  &6.953 &5.827\\
          \cline{2-11}
          & \textcolor[rgb]{0.0, 0.53, 0.74}{Average} & \textcolor[rgb]{0.0, 0.53, 0.74}{\bfseries 4.555} & \textcolor[rgb]{0.0, 0.53, 0.74}{7.039} & \textcolor[rgb]{0.0, 0.53, 0.74}{8.142} & \textcolor[rgb]{0.0, 0.53, 0.74}{6.983} & \textcolor[rgb]{0.0, 0.53, 0.74}{8.615} & \textcolor[rgb]{0.0, 0.53, 0.74}{8.738} & \textcolor[rgb]{0.0, 0.53, 0.74}{5.188}  & \textcolor[rgb]{0.0, 0.53, 0.74}{7.224} & \textcolor[rgb]{0.0, 0.53, 0.74}{4.629} \\
          \midrule
          \multirow{4}{*}{Entity-13} & ResNet18 &7.448 &7.416 &7.391 &5.544 &2.343 &3.798 &2.205 &2.182 &\textbf{0.790}\\
          & ResNet50 &5.183 &5.367 &6.155 &2.895 &6.696 &4.140 &4.132 &2.272 &\textbf{0.969}\\
          & WRN-50-2 &2.748 &2.893 &3.128 &1.817 &5.431 &3.159 &2.276 &1.297 &\textbf{0.674}\\
          \cline{2-11}
          & \textcolor[rgb]{0.0, 0.53, 0.74}{Average} & \textcolor[rgb]{0.0, 0.53, 0.74}{5.126} & \textcolor[rgb]{0.0, 0.53, 0.74}{5.225} & \textcolor[rgb]{0.0, 0.53, 0.74}{5.558} & \textcolor[rgb]{0.0, 0.53, 0.74}{3.419} & \textcolor[rgb]{0.0, 0.53, 0.74}{4.824}& \textcolor[rgb]{0.0, 0.53, 0.74}{3.699} & \textcolor[rgb]{0.0, 0.53, 0.74}{2.871} & \textcolor[rgb]{0.0, 0.53, 0.74}{1.917} & \textcolor[rgb]{0.0, 0.53, 0.74}{\textbf{\textbf{0.811}}}\\
          \midrule
          \multirow{4}{*}{Entity-30} & ResNet18 &5.060 &5.544 &5.731 &4.098 &3.768 &3.665 &1.867 &2.269 &\textbf{0.830}\\
          & ResNet50 &4.415 &5.14 &4.630 &4.499 &7.185 &4.326 &4.767 &2.158 &\textbf{1.455}\\
          & WRN-50-2 &3.477 &4.200 &3.363 &2.490 &4.17 &4.080 &2.103 &1.797 &\textbf{0.918}\\
          \cline{2-11}
          & \textcolor[rgb]{0.0, 0.53, 0.74}{Average} & \textcolor[rgb]{0.0, 0.53, 0.74}{4.317} & \textcolor[rgb]{0.0, 0.53, 0.74}{4.961} & \textcolor[rgb]{0.0, 0.53, 0.74}{4.575} & \textcolor[rgb]{0.0, 0.53, 0.74}{3.695} & \textcolor[rgb]{0.0, 0.53, 0.74}{5.042} & \textcolor[rgb]{0.0, 0.53, 0.74}{4.024} & \textcolor[rgb]{0.0, 0.53, 0.74}{2.912} & \textcolor[rgb]{0.0, 0.53, 0.74}{2.075} & \textcolor[rgb]{0.0, 0.53, 0.74}{\textbf{1.068}}\\
          \midrule
          \multirow{4}{*}{living-17} & ResNet18 &4.095 &4.098 &3.699 &3.767 &4.262 &3.737 &1.373 &2.569 &\textbf{1.975}\\
          & ResNet50 &2.802 &2.757 &1.574 &9.687 &4.361 &3.535 &3.260 &1.860 &\textbf{1.573}\\
          & WRN-50-2 &4.059 &4.250 &2.535 &3.889 &4.925 &3.833 &\textbf{2.922} &2.974 &3.281\\
          \cline{2-11}
          & \textcolor[rgb]{0.0, 0.53, 0.74}{Average} & \textcolor[rgb]{0.0, 0.53, 0.74}{3.652} & \textcolor[rgb]{0.0, 0.53, 0.74}{3.701} & \textcolor[rgb]{0.0, 0.53, 0.74}{2.603} & \textcolor[rgb]{0.0, 0.53, 0.74}{5.781} & \textcolor[rgb]{0.0, 0.53, 0.74}{4.516} & \textcolor[rgb]{0.0, 0.53, 0.74}{3.641} & \textcolor[rgb]{0.0, 0.53, 0.74}{2.519} & \textcolor[rgb]{0.0, 0.53, 0.74}{2.467} & \textcolor[rgb]{0.0, 0.53, 0.74}{\textbf{2.277}}\\
          \midrule
          \multirow{4}{*}{Nonliving-26} & ResNet18 &2.907 &3.767 &3.386 &1.891 &2.773 &3.453 &1.881 &2.168 &\textbf{1.010}\\
          & ResNet50 &4.004 &4.663 &4.685 &3.547 &6.461 &4.236 &3.955 &\textbf{2.087} &2.189\\
          & WRN-50-2 &1.903 &2.444 &2.593 &2.259 &4.220 &3.781 &2.785 &2.094 &\textbf{1.403}\\
          \cline{2-11}
          & \textcolor[rgb]{0.0, 0.53, 0.74}{Average} & \textcolor[rgb]{0.0, 0.53, 0.74}{2.938} & \textcolor[rgb]{0.0, 0.53, 0.74}{3.625} & \textcolor[rgb]{0.0, 0.53, 0.74}{3.555} & \textcolor[rgb]{0.0, 0.53, 0.74}{2.565} & \textcolor[rgb]{0.0, 0.53, 0.74}{4.485} & \textcolor[rgb]{0.0, 0.53, 0.74}{3.823} & \textcolor[rgb]{0.0, 0.53, 0.74}{2.874} & \textcolor[rgb]{0.0, 0.53, 0.74}{2.116} & \textcolor[rgb]{0.0, 0.53, 0.74}{\textbf{1.534}}\\
         \bottomrule
    \end{tabular}}
    \label{tab:mae}
\end{table*}

\end{document}